\documentclass{article}
\usepackage[dvipsnames]{xcolor}
\usepackage{iclr2027_conference,times}

\usepackage[utf8]{inputenc}
\usepackage[T1]{fontenc}
\usepackage{hyperref}
\usepackage{url}
\usepackage{booktabs}
\usepackage{amsfonts}
\usepackage{nicefrac}
\usepackage{microtype}
\usepackage{adjustbox}
\usepackage{colortbl}
\definecolor{bestcell}{rgb}{0.82, 0.93, 0.82}
\definecolor{secondcell}{rgb}{0.80, 0.87, 0.97}
\newcommand{\best}[1]{\cellcolor{bestcell}\textbf{#1}}
\newcommand{\scnd}[1]{\cellcolor{secondcell}#1}

\title{Learned End-to-End Guidance Schedules \\for Diffusion Models}

\author{%
  Aneesh Barthakur\thanks{Corresponding author: \texttt{aneesh.barthakur@simtech.uni-stuttgart.de}} \\
  University of Stuttgart
  \And
  Mathias Niepert \\
  University of Stuttgart
  \And
  Luiz F.\ O.\ Chamon \\
  \'Ecole polytechnique, Institut Polytechnique de Paris
}

\usepackage{amsthm}
\usepackage{amssymb}
\usepackage{amsmath}
\usepackage{mathtools}
\usepackage{amssymb}
\usepackage{scalerel}
\usepackage{dsfont}
\usepackage{upgreek}

\usepackage{caption}
\usepackage{subcaption}
\usepackage{graphicx}

\usepackage{mathrsfs}
\usepackage{thm-restate}
\usepackage{algorithm}
\usepackage{algpseudocode}
\usepackage{placeins}
\usepackage{float}
\usepackage{multirow}

\makeatletter
\newcommand{\globalcolor}[1]{%
  \color{#1}\global\let\default@color\current@color
}
\makeatother

\usepackage[normalem]{ulem}

\usepackage[shortlabels]{enumitem}

\usepackage[framemethod=TikZ]{mdframed}

\renewcommand{\sc}{\textsc}
\renewcommand{\sl}{\textsl}

\newcommand{\tx}{\text}

\renewcommand{\bar}{\overline}

\newcommand{\q}{\frac}

\newcommand{\ds}[1]{\frac{\del}{\del #1}}

\newcommand{\oc}{\left\lbrace}
\newcommand{\cc}{\right\rbrace}
\newcommand{\ob}{\left[}
\newcommand{\cb}{\right]}

\newcommand{\oV}{\left\|}
\newcommand{\cV}{\right\|}

\newcommand{\norm}[1]{\oV #1 \cV}
\newcommand{\set}[1]{\oc #1 \cc}

\newcommand{\R}{\mathbb{R}}

\renewcommand{\AA}{\mathcal{A}}

\newcommand{\CC}{\mathcal{C}}

\newcommand{\FF}{\mathcal{F}}

\newcommand{\LL}{\mathcal{L}}

\newcommand{\NN}{\mathcal{N}}

\renewcommand{\a}{\alpha}

\newcommand{\g}{\gamma}

\renewcommand{\th}{\theta}

\renewcommand{\l}{\lambda}

\newcommand{\Ex}[2]{\mathbb{E}_{#1}\ob #2 \cb}

\newcommand\numberthis{\addtocounter{equation}{1}\tag{\theequation}}

\newcommand{\nthis}{\numberthis}

\newcommand{\for}{\tx{FOR}}

\renewcommand{\tau}{\uptau}

\newcommand{\eps}{\epsilon}

\newcommand{\pushright}[1]{\ifmeasuring@#1\else\omit\hfill$\displaystyle#1$\fi\ignorespaces}
\newcommand{\pushleft}[1]{\ifmeasuring@#1\else\omit$\displaystyle#1$\hfill\fi\ignorespaces}

\newcommand{\ps}{P^\star}
\renewcommand{\ds}{D^\star}

\newcommand{\ls}{\l^\star}
\newcommand{\ms}{\mu^\star}

\newcommand{\psth}[1][]{\ps}
\newcommand{\dsth}[1][]{\ds}

\newcommand{\psnu}[1][]{\ps_{\phi \nu}}

\newcommand{\lsth}[1][]{\ls_{\theta}}
\newcommand{\msth}[1][]{\ms_{\theta}}

\newcommand{\tht}[1][]{\theta^{(t)}}
\newcommand{\thtm}[1][]{\theta^{(t-1)}}
\newcommand{\thtmt}[1][]{\theta^{(t-2)}}

\newcommand{\mlabel}[1]{\nthis \label{#1}}

\makeatletter
\def\nnil{\nil}
\newcounter{prob}

\makeatother

\newenvironment{prob*}{%
	\csname equation*\endcsname%
	\aligned%
}{%
	\endaligned%
	\csname endequation*\endcsname%
}

\newcommand{\abar}{\bar{\alpha}}

\global\mdfdefinestyle{exampledefault}{%
linecolor=black,linewidth=1.3pt,%
leftmargin=0cm,rightmargin=0cm,
innerleftmargin=2mm,
innerrightmargin=1.5mm,
innertopmargin=\topskip,
innerbottommargin=\topskip,
roundcorner=3pt,
backgroundcolor=black!10,
nobreak=true
}

\usepackage{titletoc}

\definecolor{indigo}{rgb}{0.0, 0.25, 0.42}

\newcommand{\sect}{Section~}

\graphicspath{{figures/}}
\iclrpreprint 
\begin{document}

\maketitle

\begin{abstract}
    Diffusion models are a powerful generative paradigm used across multimedia and scientific applications.
    Guided diffusion methods impose requirements on the generation by adding the gradient of a differentiable loss (the \emph{guidance function}) as a drift term during inference.
    The weight of this drift (the \emph{guidance scale}) is critical for the trade-off between data quality and requirement satisfaction. To achieve both of these goals, guided diffusion must resort to small guidance scales and lengthy sampling, incurring high computational costs.
    This work proposes \textit{learned end-to-end guidance schedules} (LEEGS) to achieve these objectives with \textit{fewer sampling steps}. LEEGS trains a time-dependent schedule by minimizing the guidance function over a small set of examples using stochastic gradient descent. Backpropagating through guided sampling is computationally expensive, so LEEGS uses an approximation of the gradient that cuts training time by a factor of 4.
    We evaluate LEEGS on diverse guidance tasks, including (a) image inpainting, (b) noisy image inverse problems, (c) face-ID-guided generation, and (d) forward and inverse PDE problems, outperforming baselines at equal budget (50 or 100 NFEs), or matching constant guidance with only 10\% of the steps.
\end{abstract}

\section{Introduction}
\label{sec:introduction}
Diffusion models have been successful in modeling and generating data across many domains, from multimedia tasks~[e.g., image synthesis \citep{ho2020denoisingdiffusionprobabilisticmodels,dhariwal2021diffusionmodelsbeatgans}, image restoration \citep{kawar2022ddrm, chung2023DPS}, text-to-image generation \citep{rombach2022LDM, nichol2022glide}] to scientific and engineering applications~[e.g., molecular generation \citep{hoogeboom2022equivariantdiffusionmoleculegeneration, xu2022geodiff}, protein generation \citep{watson2023rfdiff}, solutions to partial differential equations (PDEs) \citep{huang2024dpde,bastek2025physicsinformed}, design optimization \citep{maze2022topodiff}].
Guided diffusion [e.g., \citep{dhariwal2021diffusionmodelsbeatgans, 2023bansalunivguidance, wu2023practical, corso2023particle}] allows these models to produce data satisfying ancillary requirements, such as the physical viability of samples for scientific and engineering applications \citep{towards_chen_2024, projected_christopher_2024}, or solve inverse problems \citep{chung2023DPS,huang2024dpde}. While several training-free approaches induce a requirement in diffusion models, such as particle-based \citep{wu2023practical, corso2023particle} or projection-based methods \citep{christopher2024constrainedsynthesisprojecteddiffusion,projected_christopher_2024}, if its violation can be expressed as a differentiable loss $h(x)$ (the \emph{guidance function}), then modifying the diffusion with a drift $-\nabla_w h$
of strength $w > 0$ (the \emph{guidance scale})
 yields a powerful, general approach [see, e.g. \citep{dhariwal2021diffusionmodelsbeatgans, 2023bansalunivguidance}].
 
Yet, guidance exacerbates a well-known limitation of diffusion models: slow inference. All guidance techniques face a trade-off between data quality~(staying close to the model's learned distribution) and requirement satisfaction~(minimizing the violation~$h$). Small scales $(w \ll 1)$ do not sufficiently reduce the violation~$h$, yielding samples that violate the requirements; while large scales $(w \gg 1)$ cause substantial changes to the original diffusion trajectory, yielding unrealistic samples.
To generate high-quality data satisfying requirements, guided generation algorithms such as \citep{chung2023DPS, yu2023freedom} resort to \textit{small guidance scales} and many inference steps, resulting in high computational cost. Our aim is to reduce the number of steps in the sampling process to balance data quality and the satisfaction of requirements.

Different approaches have been proposed to tackle this problem. For instance, \citet{yang2024dsg} rescales the guidance gradient relative to the sampling noise at each step, enabling larger step sizes and thus shorter trajectories, but still requires choosing a guidance scale.
Methods that alternate between denoising and guidance optimization [e.g. \citep{zampini2025trainingfreeconstrainedgenerationstable,jayaram2025CDIM,huang2024constraineddiffusiontrustsampling}] control the deviation explicitly via a trust region or a proximal penalty. However, tuning the trust-region radius or penalty weight is as delicate as tuning the guidance scale itself. A different strategy for mitigating this trade-off has been explored in the classifier-free guidance (CFG) setting, where the guidance scale is made time-dependent~\citep{kynkaanniemi2024applyingguidancelimitedinterval,wang2024analysisclassifierfreeguidanceweight,sadat2024cadsunleashingdiversitydiffusion, malarz2025classifierfreeguidanceadaptivescaling} following the observation that constant guidance is suboptimal for CFG \citep{kynkaanniemi2024applyingguidancelimitedinterval}. \textit{Learning} a time-dependent scale has likewise been studied for CFG \citep{yehezkel2025annealing, galashov2025learnguidediffusionmodel,sun2025ousacoptimizedguidancescheduling}, but not for loss-based guidance, whose main advantage is that it conditions a pretrained model without retraining.

We bring \textit{learned} time-dependent guidance scales (\emph{guidance schedules}) to loss-guided diffusion, achieving better trade-offs between data quality and requirement satisfaction. LEEGS trains a schedule for a given model, task, and sampler by minimizing the value of the guidance function along the diffusion trajectory, averaged over a small set of examples. This requires backpropagating through the sampling trajectory, a naive implementation of which would incur untenable peak memory and per-batch compute. We use gradient checkpointing to bound the memory cost, similarly to when optimizing other diffusion parameters, such as sampling coefficients \citep{watson2022learningfastsamplersdiffusion}, discretizations \citep{tong2025learningdiscretizedenoisingdiffusion}, and noise \citep{tang2024inferencetimealignmentdiffusionmodels}. We further propose an approximation of the objective's gradient that reduces per-batch compute. Put together, these improvements allow LEEGS to effectively manage the guidance trade-off in a tractable manner.

The remainder of the paper is organized as follows. We first review guided diffusion sampling in \sect\ref{sec:preliminaries}, then present our problem formulation and method in \sect\ref{sec:method}. We introduce our main empirical results in \sect\ref{sec:experiments}, starting with a description of the applications we consider, namely various guided image generation tasks and PDE problems, followed by further experiments and ablation studies. Finally, we conclude in \sect\ref{sec:conclusion}.

\begin{figure}[t]
    \centering
    \begin{subfigure}[b]{\textwidth} 
        \includegraphics[width=\textwidth]{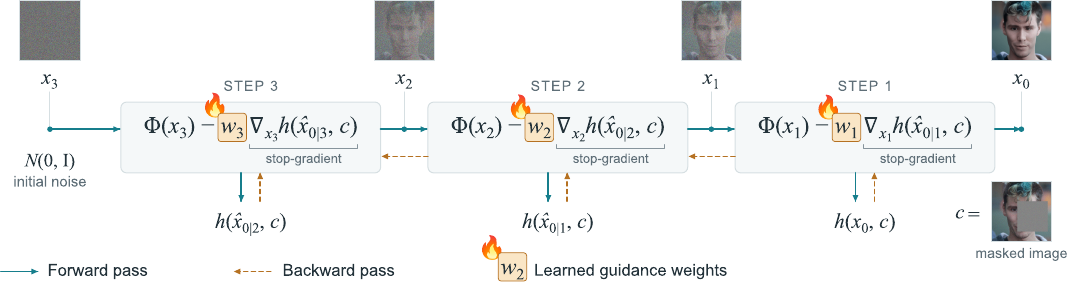}    
        \caption{\textbf{LEEGS training procedure.} The forward pass (computing $\LL(W)$) runs the guided diffusion sampling, while the backward pass (updating $W$) backpropagates the objective through the sampling trajectory.}
    \end{subfigure}

    \vspace{2mm}
    \begin{subfigure}[b]{\textwidth}
        \centering
        \includegraphics[width=\textwidth]{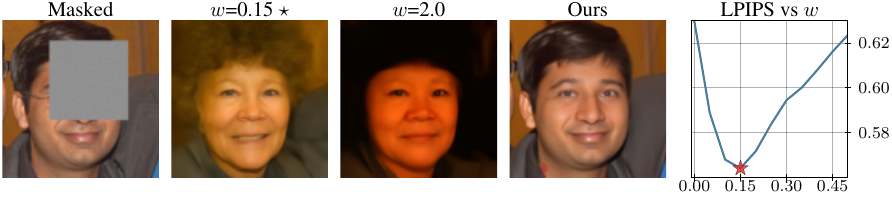}
        \caption{LEEGS achieves convincing inpainting results with \textit{only 10 inference steps} while constant guidance only reconstructs a plausible-looking but incorrect face even at $w=0.15$, which achieves the optimal LPIPS on a held-out set of 100 images. A larger scale ($w=2.0$) does not improve consistency with the unmasked regions, but produces unrealistic, saturated artifacts instead. See Section \ref{sec:applications} for details on the noisy box-inpainting task. \label{fig:compare_with_constant_10steps} }
    \end{subfigure}

\end{figure}
\section{Guided diffusion}
\label{sec:preliminaries}

Modern diffusion and score-based generative models \citep{ho2020denoisingdiffusionprobabilisticmodels,song2021scorebasedgenerativemodelingstochastic} use a learned noise-prediction network \(\eps_\theta(x,t)\) that defines a vector field whose reverse-time dynamics are integrated numerically to generate samples \(x_0 \in \mathcal{X}\) from a random starting point \(x_S \sim p(x_S)=\mathcal{N}(0,\mathbf{I})\). We denote by
\[
    x_{s-1} = \Phi(x_s, t_s; \eps_\theta), \qquad s = S,\ldots,1, \mlabel{eqn:unguided_sampling}
\]
a single step of this reverse diffusion, where \(\Phi\) represents a (possibly stochastic) numerical integrator~[e.g., DDIM \citep{song2022denoisingdiffusionimplicitmodels}, DDPM\citep{ho2020denoisingdiffusionprobabilisticmodels}, or higher-order solvers \citep{karras2022elucidatingdesignspacediffusionbased}] and $T=t_S > \ldots > t_1 > t_0=0$ is a discretization of the diffusion time. The resulting trajectory \(x_{0:S}\) is a Markov chain with joint distribution \(p(x_{0:S})\) and marginal distribution at time \(t_s\) denoted \(p(x_s)\), for $s=S,\ldots,0$. Since $\eps_\theta$ is fixed throughout, we suppress $\theta$ and write \(\hat{x}_{0|s} \coloneqq \hat{x}_0(x_s,t_s;\eps_\theta)\) for the one-step denoised estimate of $x_s$, recovered from $\eps_\theta(x_s,t_s)$ via Tweedie's formula (Appendix~\ref{app:pseudocode} \eqref{eqn:tweedie}). The diffusion model is trained so that $p(x_0)$ approximates the data distribution.

The reverse diffusion process can be modified to produce samples that satisfy additional properties or requirements using the mechanism of guidance. The guided generation process uses a non-negative guidance function $h : \mathcal{X} \times \mathcal{C} \to \mathbb{R}_+$ 
that measures the violation of a desired property or requirement, often parametrized by a condition $c \in \mathcal{C}$. The goal is to generate samples $x_0$ that minimize $h(x_0, c)$ [ideally, $h(x_0, c) = 0$]. Consider the example of image inpainting where the goal is to restore an image missing a block of pixels. Here, the condition $c$ is the degraded image itself and the guidance function is the normed difference between the generated sample and $c$ at the observed pixel locations  (see Sec.~\ref{sec:applications} for more examples). 

The guidance function induces a vector field
via its gradient $\nabla_{x_s} h\!\left(\hat{x}_{0|s}, c\right)$ evaluated at the one-step denoised estimate, which can be used to modify the diffusion reverse-time dynamics from \eqref{eqn:unguided_sampling}. Given a guidance schedule \(W \in \mathbb{R}^{S}_{\geq 0}\) that collects the time-dependent guidance scales $w_{s}$, the guided trajectory is obtained via
\begin{align*}
    x_{s-1}
    =
    \Phi\!\left(x_s, t_s; \eps_\theta\right) -  w_{s}\, \nabla_{x_s} h\!\left(\hat{x}_{0|s}, c\right), \qquad s = S,\ldots,1. \mlabel{eqn:guided_sampling}
\end{align*}
We follow \cite{chung2023DPS} and evaluate $h$ on $\hat{x}_{0|s}$ instead of $x_s$, since the former is a better proxy for the final sample. Note that computing $\nabla_{x_s} h\!\left(\hat{x}_{0|s}, c\right)$ requires differentiating through the diffusion model, since $\hat{x}_{0|s}$ depends on $\eps_{\th}(x_s, t_s)$. We denote by \(p_W(x_{0:S} \mid c)\) the joint distribution of the resulting trajectory and by \(p_W(x_s \mid c)\), $s=S,\ldots,0$, the marginal distributions of the samples at time \(t_s\). As the guided marginals $p_W(x_s \mid c)$ ($s>0$) get farther from the unguided $p(x_s)$, sampling error accumulates, pushing $p_W(x_0 \mid c)$ away from the data distribution and producing artifacts. Prior methods control this deviation by constraining each guidance step [a trust region or a proximal step \citep{jayaram2025CDIM,huang2024constraineddiffusiontrustsampling}], but the constraint radius or penalty weight is itself as delicate a hyperparameter as the guidance scale. Our approach instead \emph{learns} time-dependent scales that directly optimize $h$.

\section{Learned end-to-end guidance schedules}
\label{sec:method}

\paragraph{Objective.} We seek a guidance schedule \(W\) that yields samples that both satisfy the requirement encoded by \(h(x_0, c)\) and stay close to the unguided distribution \(p(x_0)\). We impose the requirements on average over conditions $c$ drawn from a fixed distribution $p_{\mathcal{C}}$,  with the pre-trained model \(\eps_\theta\) held fixed. The natural objective is to minimize the guidance function at the terminal sample alone, i.e. \(\Ex{c,\, x_0 \sim p_W(\cdot \mid c)}{h(x_0, c)}\); however, this terminal-only objective is hard to optimize directly, since the gradient of $h(x_0, c)$ with respect to $w_s$ decays for steps far from the end of the trajectory (large $s$), leaving long trajectories with little training signal (Appendix~\ref{app:trajectory_objective}). Including all the terms $h(\hat{x}_{0|s}, c)$ ($0 \leq s < S$) along the trajectory (with the convention that $\hat{x}_{0|0} = x_0$) restores a usable gradient at every step, giving the objective we minimize :
\begin{align*}
    \LL(W) =   \Ex{c \sim p_{\CC},\; x_{0:S} \sim p_W(\cdot\mid c)}{\sum_{s=0}^{S-1} h(\hat{x}_{0|s}, c)}. \mlabel{eqn:part_obj}
\end{align*}
Computing the objective for a single draw $(x_S, c)$ requires simulating the full guided trajectory \eqref{eqn:guided_sampling}, a cost that scales linearly with the number of sampling steps $S$.

\paragraph{Gradient computation.} Assume that the sampler $\Phi$ is deterministic, so that the entire trajectory $x_{0:S}$ (and hence every $\hat{x}_{0|s}$) is deterministic given $c$, $x_S$, and $W$. Writing~$\ell(W; x_S, c) \coloneqq \sum_{s=0}^{S-1} h(\hat{x}_{0|s}, c)$ for this now-deterministic function of $x_S$, the objective of \eqref{eqn:part_obj} becomes~$\LL(W)=\Ex{c \sim p_{\CC},\, x_S \sim p(x_S)}{\ell(W; x_S, c)}$. Unlike $p_W(\cdot \mid c)$, $p_{\CC}$ and $p(x_S)$ are fixed and do not depend on $W$. Removing $W$ from the sampling distribution lets us differentiate under the expectation, so that $\nabla_W\LL(W)=\Ex{c,\, x_S}{\nabla_W \ell(W; x_S, c)}$ can be estimated by sampling $c \sim p_{\CC}$, $x_S \sim p(x_S)$ and computing the average of $\nabla_W \ell(W; x_S, c)$ over these samples. For stochastic $\Phi$, a similar argument holds by reparameterizing the sampler noise \citep{kingma2013autoencoding}. Define $g_s = \nabla_{x_s} h(\hat{x}_{0|s}, c)$, the guidance gradient at step $s$. Expanding the derivative of $\ell(W; x_S, c)$ with respect to $w_k$, $k \in \set{1,\ldots,S}$, we obtain
\begin{align*}
    \q{\partial \ell(W; x_S,c)}{\partial w_k} &=\sum_{s=0}^{S-1} \q{\partial}{\partial x_s} h\!\left(\hat{x}_{0|s}, c\right) \cdot \q{\partial x_s}{\partial w_k}\\
    &=g_{k-1} \cdot \q{\partial x_{k-1}}{\partial w_k} + \sum_{s=0}^{k-2} g_s \cdot \q{\partial x_s}{\partial x_{s+1}} \q{\partial x_{s+1}}{\partial x_{s+2}} \cdots \q{\partial x_{k-2}}{\partial x_{k-1}} \q{\partial x_{k-1}}{\partial w_k}. \mlabel{eqn:grad_decomp}
\end{align*}
To obtain the second equality, observe that $x_s$ does not depend on $w_k$ for $s \geq k$. Differentiating the guided update \eqref{eqn:guided_sampling} with respect to $x_s$ gives
\begin{align*}
 \q{\partial x_{s-1}}{\partial x_s} = \q{\partial \Phi\!\left(x_s, t_s; \eps_\theta\right)}{\partial x_s} - w_s \cdot \q{\partial g_s}{\partial x_s} .\mlabel{eqn:grad_x_s}
\end{align*}
Likewise, computing the partial derivative $\q{\partial x_{k-1}}{\partial w_k}$ by differentiating the guided update \eqref{eqn:guided_sampling} with respect to $w_k$, we obtain $\q{\partial x_{k-1}}{\partial w_k}=-g_k$. This follows since $x_k$ does not itself depend on $w_k$, and $w_k$ enters the update only through the linear term $-w_k g_k$. While the first term in \eqref{eqn:grad_decomp} captures the effect of $w_k$ on the guidance function at the immediate next step, the second term captures its effect on all later steps. However, the latter term involves repeated computations of $\q{\partial g_s}{\partial x_s}$ in \eqref{eqn:grad_x_s}, which involves second derivatives of the network. 

Training with the true gradient is therefore prohibitive at large $S$, so we approximate \eqref{eqn:grad_x_s} by ignoring the term $w_s\,\q{\partial g_s}{\partial x_s}$. This is equivalent to applying a stop-gradient to the guidance gradient $g_s$ during sampling, so that $w_k$ still affects later steps through the sampler Jacobian $\q{\partial \Phi}{\partial x_s}$, but not through $g_s$. This is roughly $4\times$ cheaper to compute than the true gradient~(Appendix~\ref{app:method_comparison}). We optimize $W$ by projected stochastic gradient descent, clipping negative entries to zero after each step; once trained, $W$ is applied at sampling time with no additional cost. We refer to the resulting method as LEEGS.

Naive backpropagation stores the activations of all $S$ network evaluations along the trajectory, so peak memory grows linearly in $S$ and quickly exceeds GPU capacity. We avoid this with gradient checkpointing (rematerialization) \citep{kumar2019remat,tong2025learningdiscretizedenoisingdiffusion}: rather than store a step's activations, we recompute them during the backward pass, making peak memory independent of $S$ in exchange for extra computation.

\paragraph{Regularization.} Attempting to drive the guidance function $h$ too low may occasionally push the guided marginals~$p_W(x_s \mid c)$ too far from the unguided~$p(x_s)$, yielding unrealistic samples, which can be seen as a form of reward hacking \citep{tang2024inferencetimealignmentdiffusionmodels, zhai2025mira}. This manifests in  Figure~\ref{fig:wdecay_train_lpips} as an increase in the LPIPS metric after a certain point in the optimization. This phenomenon can be mitigated either by early stopping, when a held-out reward is available to detect the decline in performance, or by penalizing the magnitude of the guidance schedule $W$ in the training objective as in, e.g. minimizing  $\LL(W) + \g\,\norm{W}_2^2$ with $\g > 0$. Appendix~\ref{app:regularization} explores which tasks are prone to reward hacking and how the penalty alleviates it. Typically, even on tasks which \textit{are} prone to reward hacking, it appears only at high $S$ (e.g., $S=100$), not at lower $S$ (e.g., $S=25$).

\paragraph{Parametrization.} LEEGS optimizes the schedule freely over $W \in \R^{S}_{+}$. While $W$ can be constrained to a parametric family, $w_s = \rho_\phi(s)$ with $\phi$ small, trading flexibility for fewer parameters and a smoother schedule, this requires the functional form to be fixed in advance. Since LEEGS's cost is dominated by the guided sampling, we find this trade-off unnecessary in practice.

\section{Experiments}
\label{sec:experiments}

We evaluate LEEGS on guided image generation and PDE problems, comparing it with baseline methods across budgets and analyzing its schedules, sample efficiency, and sensitivity to sampler choice and inference steps.

\subsection{Benchmark tasks}
\label{sec:applications}

\paragraph{Image inpainting.} Inpainting seeks to recover a complete image from a partial observation. With condition $c = (y, M)$, where $y$ is the observed image and $M$ a binary mask marking the missing pixels, we consider the guidance function $h(x, c) = \norm{M \odot x - y}_2$, where $\odot$ denotes elementwise multiplication. Unlike projection-renoising methods, such as RePaint \citep{lugmayr2022repaintinpaintingusingdenoising}, which inject the observed pixels directly and can leave visible seams at low step counts, guided sampling enforces the observed pixels softly. We generate synthetic inpainting problems on ImageNet and CelebA-HQ using the five mask types from \cite{lugmayr2022repaintinpaintingusingdenoising}: wide and narrow (large or thin contiguous regions are not observed), super-resolution $2\times$ (one pixel kept per $2\times2$ block), alternating-lines (every other row removed), and half (right half removed). We compare against RePaint at a matched 200-NFE budget and include the original results using 2400 NFEs from \cite{lugmayr2022repaintinpaintingusingdenoising} as a reference. We report LPIPS \citep{zhang2018perceptual}, the perceptual distance to the ground truth image (lower is better).

\paragraph{Noisy inverse problems.} We use image restoration to illustrate an inverse problem \citep{kawar2022ddrm, chung2023DPS, wang2022zero} with a known degradation operator $\AA$. The noisy measurement of the image $x_{\tx{GT}}$ under condition  $c = (y, \AA)$ is $y = \AA(x_{\tx{GT}}) + \sigma n$, with an unknown noise level $\sigma > 0$ and $n \sim \NN(0, \mathbf{I})$. We choose the guidance function as $h(x, c) = \norm{\AA(x) - y}_2$. Following \cite{chung2023DPS}, we synthesize problems from held-out FFHQ and ImageNet images and evaluate five tasks at $\sigma = 0.05$: $4\times$ super-resolution, box inpainting (a large centred square removed), random inpainting (each pixel dropped independently with high probability), and Gaussian and motion deblurring (details of the operators are in App.~\ref{app:exp_details}). Unlike the noiseless inpainting benchmark above, the observations here are corrupted, handicapping methods that inject observed pixels directly, such as RePaint. We compare against recent methods for inverse problems (CDIM \citep{jayaram2025CDIM}, DSG \citep{yang2024dsg}, and DAPS \citep{zhang2025improvingdiffusioninverseproblem}) at 50 and 100 NFEs, and include the 1000-NFE DPS \citep{chung2023DPS} numbers reported in \cite{chung2023DPS} as a reference. We also compare against constant guidance and heuristic increasing and decreasing linear schedules, each with a single parameter tuned by grid search on a held-out set (App.~\ref{app:exp_details}). Alongside LPIPS, we report FID \citep{heusel2017gans}, the distributional distance between generated samples and a reference set (lower is better).

\paragraph{Face-ID guidance.} Following \cite{yu2023freedom,yang2024dsg}, we generate images of a target identity given a single reference photo $x_{\tx{ref}}$ (the condition $c$), using the guidance function $h(x, x_{\tx{ref}}) = \norm{F(x) - F(x_{\tx{ref}})}_2$, where $F$ is the feature embedding of a pretrained face-recognition network [we use the ArcFace model \citep{deng2019arcface}]. Such a network embeds faces so that images of the same person lie close together, so minimizing $h$ steers the generated image toward the reference identity. We compare against constant guidance and DSG \citep{yang2024dsg} on CelebA-HQ at 50 and 100 NFEs. We report FID and the guidance value itself, the \emph{FaceID score} (identity distance to the reference, lower is better).

\paragraph{PDE problems.} Diffusion models can solve partial differential equations (PDEs) from sparse observations, complementing classical \citep{trefethen1996finite, langtangen2019numericalbook} and learned \citep{raissi2019pinns, li2020fourier} solvers. Following DiffusionPDE \citep{huang2024dpde}, we consider boundary-value problems specified by a coefficient field $a$ and its solution field $u$. The diffusion model is pretrained on a dataset of solved instances and represents the joint distribution over coefficient--solution pairs $(a, u)$ rendered as images. Given observations $\{a_{\tx{obs},i}\}_{i=1}^{N_{\tx{obs}}}$ of the coefficient at locations $\{z_{\tx{obs},i}\}_{i=1}^{N_{\tx{obs}}}$, the inverse-problem guidance function is $h_{\tx{inv}}(x, c) = \sqrt{\q{1}{N_{\tx{obs}}}\sum_{i=1}^{N_{\tx{obs}}} \left(a(z_{\tx{obs},i}) - a_{\tx{obs},i}\right)^2}$ with condition $c = (a_{\tx{obs}}, z_{\tx{obs}})$; the forward guidance function is defined analogously on the solution field, $h_{\tx{fwd}}(x, c) = \sqrt{\q{1}{N_{\tx{obs}}}\sum_{i=1}^{N_{\tx{obs}}} \left(u(z_{\tx{obs},i}) - u_{\tx{obs},i}\right)^2}$ with condition $c = (u_{\tx{obs}}, z_{\tx{obs}})$. We evaluate forward (recover $u$ from $a$) and inverse (recover $a$ from $u$) problems for the Helmholtz, Poisson, and Darcy equations, with $\approx 3\%$ of pixels observed. We compare against constant guidance and heuristic increasing and decreasing linear schedules at 50 and 100 steps, and include the DiffusionPDE \citep{huang2024dpde} results at 2000 steps as a reference. While DiffusionPDE additionally applies a PDE-residual guidance term over a small part of the trajectory, we found little benefit from it and omit it for both constant guidance and LEEGS. Following DiffusionPDE, constant guidance uses a second-order sampler; training LEEGS with a second-order sampler did not improve on results obtained with a first-order one. We report the relative error to the ground-truth solution, or the error rate for the binary-valued Darcy coefficient.

\begin{table}[ht]
\centering
\begin{adjustbox}{max width=\linewidth}
\begin{tabular}{l l cc cc cc cc cc}
\hline
\vspace{1mm}
\multirow{2}{*}{} & \multirow{2}{*}{NFEs} & \multicolumn{2}{c}{SR (x4)} & \multicolumn{2}{c}{Inpaint (box)} & \multicolumn{2}{c}{Inpaint (random)} & \multicolumn{2}{c}{Deblur (gauss)} & \multicolumn{2}{c}{Deblur (motion)} \\ \vspace{1mm} 
 &  &LPIPS &FID &LPIPS &FID &LPIPS &FID &LPIPS &FID &LPIPS &FID \\ \hline
Const. & 50 & 0.409 & 58.53 & 0.465 & 59.06 & 0.400 & 47.36 & 0.430 & 54.30 & 0.466 & 52.74 \\
DAPS & 50 & 0.316 & 44.60 & 0.237 & 38.44 & 0.497 & 136.55 & 0.314 & 43.76 & 0.272 & 45.29 \\
DSG & 50 & 0.274 & 45.67 & 0.361 & 75.21 & 0.310 & 47.10 & 0.313 & 50.36 & 0.640 & 143.59 \\
CDIM & $\approx$ 50 & 0.250 & 39.39 & \scnd{0.178} & \best{31.51} & 0.245 & 35.81 & 0.273 & \best{32.83} & 0.283 & \best{30.12} \\
Linear increase & 50 & \scnd{0.224} & \best{35.08} & 0.189 & 36.12 & \scnd{0.199} & \scnd{34.03} & \scnd{0.256} & \scnd{42.52} & \scnd{0.219} & 40.35 \\
Linear decrease & 50 & 0.521 & 63.17 & 0.540 & 60.65 & 0.509 & 62.47 & 0.525 & 59.88 & 0.522 & 61.50 \\
\textbf{LEEGS (ours)} & 50 & \best{0.222} & \scnd{35.52} & \best{0.175} & \scnd{33.11} & \best{0.197} & \best{33.63} & \best{0.243} & 42.55 & \best{0.211} & \scnd{39.24} \\
\hline
Const. & 100 & 0.235 & \scnd{32.75} & 0.341 & 47.69 & 0.259 & 31.83 & 0.298 & 37.57 & 0.396 & 36.62 \\
DAPS & 100 & 0.287 & 39.19 & 0.230 & 38.15 & 0.406 & 84.38 & 0.291 & 39.55 & 0.241 & 39.27 \\
DSG & 100 & 0.260 & 38.16 & 0.308 & 73.86 & 0.250 & 36.92 & 0.283 & 40.14 & 0.476 & 62.95 \\
CDIM & $\approx$ 100 & 0.242 & 37.43 & 0.179 & 31.61 & 0.243 & 33.41 & 0.258 & \best{31.17} & 0.265 & \best{28.19} \\
Linear increase & 100 & \scnd{0.218} & 33.13 & \scnd{0.146} & \scnd{26.72} & \best{0.174} & \best{26.06} & \scnd{0.241} & 36.27 & \scnd{0.195} & 31.43 \\
Linear decrease & 100 & 0.400 & 47.74 & 0.463 & 50.40 & 0.405 & 48.18 & 0.412 & 44.91 & 0.408 & 46.45 \\
\textbf{LEEGS (ours)} & 100 & \best{0.217} & \best{32.40} & \best{0.141} & \best{25.39} & \scnd{0.176} & \scnd{26.44} & \best{0.228} & \scnd{34.05} & \best{0.181} & \scnd{30.29} \\
\hline
DPS & 1000 & 0.214 & 39.35 & 0.168 & 33.12 & 0.212 & 21.19 & 0.257 & 44.05  & 0.242 & 39.92 \\ \hline
\end{tabular}
\end{adjustbox}
\vspace{2mm}
\caption{Comparison for noisy image inverse tasks (FFHQ). \textbf{Green} (bold): best; \textbf{blue}: second best (non-reference rows).}
\label{tab:results_ffhq_noisy}
\end{table}

\begin{table}[ht]
\centering
\begin{adjustbox}{max width=\linewidth}
\begin{tabular}{l l cc cc cc cc cc}
\hline
\vspace{1mm}
\multirow{2}{*}{} & \multirow{2}{*}{NFEs} & \multicolumn{2}{c}{SR (x4)} & \multicolumn{2}{c}{Inpaint (box)} & \multicolumn{2}{c}{Inpaint (random)} & \multicolumn{2}{c}{Deblur (gauss)} & \multicolumn{2}{c}{Deblur (motion)} \\ \vspace{1mm} 
 &  &LPIPS &FID &LPIPS &FID &LPIPS &FID &LPIPS &FID &LPIPS &FID \\ \hline
Const. & 50 & 0.666 & 93.92 & 0.439 & 113.71 & 0.651 & 104.47 & 0.649 & 94.45 & 0.654 & 94.34 \\
DAPS & 50 & 0.402 & \scnd{53.43} & 0.296 & 70.73 & 0.597 & 206.29 & \scnd{0.405} & \best{54.89} & \best{0.297} & \best{35.40} \\
DSG & 50 & 0.428 & 71.28 & 0.483 & 107.38 & 0.461 & 88.15 & 0.471 & 78.82 & 0.660 & 149.75 \\
CDIM & $\approx$ 50 & 0.379 & 56.01 & \best{0.245} & \best{51.44} & \scnd{0.336} & \scnd{46.78} & \scnd{0.405} & 61.50 & 0.404 & 67.77 \\
Linear increase & 50 & \scnd{0.373} & 55.66 & 0.354 & 87.84 & 0.344 & 47.74 & 0.468 & 77.56 & 0.425 & 72.27 \\
Linear decrease & 50 & 0.689 & 101.31 & 0.707 & 107.11 & 0.702 & 104.71 & 0.689 & 94.88 & 0.695 & 99.46 \\
\textbf{LEEGS (ours)} & 50 & \best{0.333} & \best{46.86} & \scnd{0.286} & \scnd{55.07} & \best{0.272} & \best{31.54} & \best{0.395} & \scnd{58.87} & \scnd{0.338} & \scnd{50.76} \\
\hline
Const. & 100 & 0.488 & 76.41 & 0.299 & 82.75 & 0.416 & 73.85 & 0.536 & 77.64 & 0.595 & 85.46 \\
DAPS & 100 & 0.378 & 48.92 & 0.287 & 64.77 & 0.531 & 155.06 & \scnd{0.384} & \scnd{52.00} & \best{0.276} & \best{32.81} \\
DSG & 100 & \scnd{0.343} & \scnd{45.34} & 0.453 & 126.49 & 0.347 & 47.72 & 0.419 & 63.25 & 0.672 & 193.45 \\
CDIM & $\approx$ 100 & 0.363 & 49.37 & \scnd{0.241} & \scnd{47.95} & 0.330 & 43.45 & 0.391 & 57.24 & 0.381 & 58.27 \\
Linear increase & 100 & 0.355 & 50.97 & 0.282 & 73.71 & \scnd{0.272} & \scnd{31.37} & 0.437 & 65.23 & 0.380 & 58.26 \\
Linear decrease & 100 & 0.609 & 88.48 & 0.647 & 110.93 & 0.619 & 94.36 & 0.609 & 83.92 & 0.609 & 86.52 \\
\textbf{LEEGS (ours)} & 100 & \best{0.319} & \best{39.62} & \best{0.234} & \best{47.73} & \best{0.235} & \best{23.69} & \best{0.368} & \best{50.38} & \scnd{0.310} & \scnd{42.33} \\
\hline
DPS & 1000 & 0.337 & 50.66 & 0.262 & 38.82 & 0.303 & 35.87 & 0.444 & 62.72 & 0.389 & 56.08\\ \hline
\end{tabular}
\end{adjustbox}
\vspace{2mm}
\caption{Comparison for noisy image inverse tasks (Imagenet). \textbf{Green} (bold): best; \textbf{blue}: second best (non-reference rows).}
\label{tab:results_imagenet_noisy}
\end{table}

\begin{table}[ht]
    \small
\centering
\begin{tabular}{l | l cc cc cc cc cc}
\hline
\vspace{1mm}
Dataset & Method & NFEs & Wide & Narrow & S. res. (2x) & Alternate Lines & Half \\ \hline
\multirow[c]{3}{*}{Imagenet} & RePaint & 200 & 0.186 & 0.18 & 0.622 & 0.394 & 0.368\\
& \textbf{LEEGS (ours)} & 200 & \best{0.182} & \best{0.151} & \best{0.147} & \best{0.111} & \best{0.346} \\
\cline{2-8}
& RePaint & $\approx$2400 & 0.134 & 0.064 & 0.183 & 0.089 & 0.304 \\ \hline
\multirow[c]{3}{*}{CelebA-HQ} & RePaint & 200 & 0.128 & 0.128 & 0.205 & 0.049 & 0.283 \\
& \textbf{LEEGS (ours)} & 200 & \best{0.102} & \best{0.080} & \best{0.065} & \best{0.042} & \best{0.229} \\
\cline{2-8}
& RePaint & $\approx$2400 & 0.059 & 0.028 & 0.029 & 0.009 & 0.165 \\ \hline
\end{tabular}
\vspace{2mm}
\caption{Comparison for image inpainting tasks for CelebA-HQ and Imagenet. \textbf{Green} (bold): best (non-reference rows).}
\label{tab:results_inpainting_celeba}
\end{table}

\begin{figure}[ht]
    \centering
    \begin{minipage}[c]{0.48\textwidth}
        \small
        \centering
        \begin{tabular}{l l cc}
        \hline
        \vspace{1mm}
        Method & NFEs & FaceID Score & FID \\ \hline
        Const & 50 & \scnd{0.456} & \scnd{77.16} \\
        DSG & 50 & 0.466 & 93.38 \\
        \textbf{LEEGS (ours)} & 50 & \best{0.401} & \best{76.90} \\
        \hline
        Const & 100 & \scnd{0.376} & \scnd{73.60} \\
        DSG & 100 & 0.395 & 80.24 \\
        \textbf{LEEGS (ours)} & 100 & \best{0.324} & \best{71.33} \\ \hline
        \end{tabular}
        \vspace{2mm}
        \captionof{table}{Comparison for Face ID tasks for CelebA-HQ. \textbf{Green} (bold): best; \textbf{blue}: second best.}
        \label{tab:results_faceid_celeba}
    \end{minipage}
    \hfill
    \begin{minipage}[c]{0.48\textwidth}
        \centering
        \includegraphics[width=\textwidth]{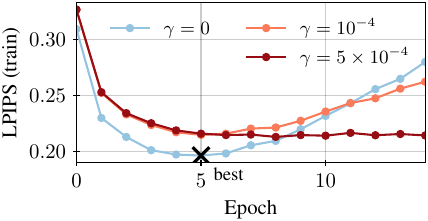}
        \captionof{figure}{\small Reward hacking at $S=100$ for FFHQ superresolution, subsequently solved by applying weight decay $\gamma \norm{W}_2^2$.}
        \label{fig:wdecay_train_lpips}
    \end{minipage}
\end{figure}

\begin{table}[ht]
    \small
\centering
\begin{tabular}{l l cc cc cc cc cc}
\hline
\vspace{1mm}
\multirow{2}{*}{Method} & \multirow{2}{*}{NFEs} & \multicolumn{2}{c}{Poisson} & \multicolumn{2}{c}{Helmholtz} & \multicolumn{2}{c}{Darcy} \\ \vspace{1mm}
 &  & Forward & Inverse & Forward & Inverse & Forward & Inverse \\ \hline
Const. & 50x2 & 19.34\% & 33.11\% & 27.56\% & 32.15\% & 21.54\% & \scnd{15.58\%} \\
Linear increase & 50x1 & 27.59\% & 46.43\% & 39.72\% & 45.42\% & 24.70\% & 20.98\% \\
Linear decrease & 50x1 & \scnd{10.25\%} & \scnd{29.81\%} & \scnd{17.07\%} & \scnd{29.84\%} & \best{11.03\%} & \best{14.05\%} \\
\textbf{LEEGS (ours)} & 50x1 & \best{5.80\%} & \best{29.36\%} & \best{10.03\%} & \best{29.71\%} & \scnd{11.33\%} & 17.84\% \\
\hline
Const. & 100x2 & 15.62\% & 30.10\% & 22.96\% & 29.54\% & 18.17\% & 16.71\% \\
Linear increase & 100x1 & 22.23\% & 28.46\% & 37.68\% & 28.93\% & 23.26\% & 18.36\% \\
Linear decrease & 100x1 & \scnd{7.61\%} & \scnd{25.39\%} & \scnd{12.25\%} & \scnd{26.01\%} & \scnd{8.53\%} & \best{14.79\%} \\
\textbf{LEEGS (ours)} & 100x1 & \best{5.84\%} & \best{25.11\%} & \best{9.57\%} & \best{25.84\%} & \best{8.20\%} & \scnd{16.58\%} \\
\hline
DPDE & 2000x2 & 4.5\% &  20.0\% & 8.8\% & 22.6\% & 2.5\% & 3.2\% &\\ \hline
\end{tabular}
\vspace{2mm}
\caption{Comparison for forward and inverse PDE problems with sparse observations. \textbf{Green} (bold): best; \textbf{blue}: second best (non-reference rows).}
\label{tab:results_dpde}
\end{table}

\begin{figure}[ht]
    \centering
    \begin{subfigure}[b]{0.48\textwidth}
        \includegraphics[width=\textwidth]{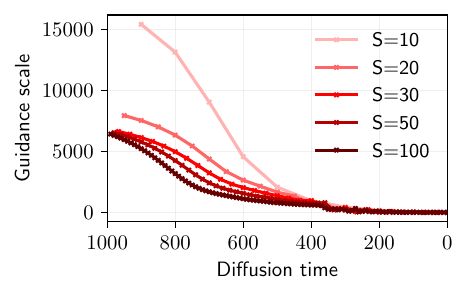}
    \end{subfigure}
    \hfill    
    \begin{subfigure}[b]{0.48\textwidth}
        \includegraphics[width=\textwidth]{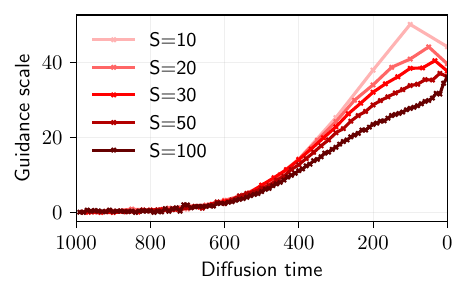}
    \end{subfigure}
    \caption{Learned guidance schedules for the forward Helmholtz problem (left) and the noisy box inpainting problem (right). The schedules are similar across different step budgets for a particular problem, but differ between problems, highlighting the need for learning methods.}
    \label{fig:guidance_schedule_panel}
\end{figure}

\subsection{Main results}

For each task and step budget, LEEGS trains a guidance schedule on 30 examples drawn from a training pool disjoint from the corresponding test set, using Adam with batch size 1 and training separately for every subtask and NFE budget. This pool has 90 images for every task except FFHQ, where we instead use a cross-validation-style split. All evaluations (except reference rows taken from original works) used test sets of size: 950 for FFHQ (Table~\ref{tab:results_ffhq_noisy}); 1000 for ImageNet noisy inverse problems (Table~\ref{tab:results_imagenet_noisy}); 100 for the inpainting and Face-ID tasks (Tables~\ref{tab:results_inpainting_celeba} and~\ref{tab:results_faceid_celeba}); and 1000 for Helmholtz and Darcy, 924 for Poisson, for the PDE problems (Table~\ref{tab:results_dpde}). Exact training and dataset details (e.g., learning rates, epochs, pretrained models) are given in App.~\ref{app:exp_details}. All results for LEEGS are the mean of three independent runs (with randomly drawn training sets). 

\paragraph{Noisy inverse problems.} Tables~\ref{tab:results_ffhq_noisy} and~\ref{tab:results_imagenet_noisy} report LPIPS and FID on FFHQ and ImageNet. On FFHQ, LEEGS is the best or second-best method in nearly every column at both 50 and 100 NFE, with the best LPIPS on four of the five subtasks. On ImageNet, LEEGS is best or second-best in every column at both 50 and 100 NFE, and the best in 4 out of 5 subtasks for ImageNet at 100 NFE with respect to both metrics. The tuned linear-\emph{increasing} schedule is itself a strong heuristic here, while linear-\emph{decreasing} is consistently poor, illustrating that effective schedule shapes are task-dependent (Fig.~\ref{fig:guidance_schedule_panel}) and LEEGS recovers a good one without that prior. LEEGS at 100 NFE also beats the 1000-NFE DPS reference on both LPIPS and FID for most subtasks, a $10\times$ reduction in steps.

\paragraph{Image inpainting.} At a matched 200-NFE budget, LEEGS improves over RePaint on every mask type and both datasets (Table~\ref{tab:results_inpainting_celeba}); RePaint with $12\times$ the budget stays ahead on most masks.

\paragraph{Face-ID guidance.} Table~\ref{tab:results_faceid_celeba} reports FaceID similarity and FID for guided face generation. At 50 NFE, all three rows are similar, except DSG's FID which is significantly worse. However, at 100 NFE, LEEGS performs best on both metrics, improving over DSG and constant guidance.

\paragraph{PDE problems.} LEEGS attains the lowest relative error on 4 of the 6 forward/inverse problems at 50 sampling steps, and on 5 of 6 at 100 steps (Table~\ref{tab:results_dpde}), in both cases at half the NFE of the constant-guidance baseline (which runs a second-order sampler). However, DiffusionPDE's results, which use $20\times$ more steps, retain an advantage. The linear-\emph{decreasing} schedule remains a strong fixed baseline here: it is in fact the best method for the inverse Darcy problem at both step budgets, and for the Darcy forward problem at 50 steps, though the same heuristic is among the worst on the noisy-inverse tasks, where linear-\emph{increasing} dominates instead. Which fixed shape works is task- and problem-specific and not knowable in advance; LEEGS matches or outperforms both heuristics on most problems.

\subsection{Analysis}

\paragraph{Learned schedules.} Figure~\ref{fig:guidance_schedule_panel} shows example schedules learned by LEEGS. For a given problem they have a similar shape across step budgets, but they differ substantially between problems: increasing schedules help the noisy inverse problems, whereas decreasing schedules help the PDE tasks, mirroring the performance of the linear-schedule baselines in Tables~\ref{tab:results_ffhq_noisy}, \ref{tab:results_imagenet_noisy}, and~\ref{tab:results_dpde}. Schedules differ even between closely related tasks:  among the five FFHQ inverse problems, App.~\ref{app:cross_task} shows that the schedule learned for a task is best on that task in nearly every case, motivating the need for a learned, task-specific schedule. 

\paragraph{Beyond DPS-style sampling.} LEEGS requires only a differentiable sampling trajectory and makes no assumption on the form of the sampler. Beyond stochastic (DDPM; Tables~\ref{tab:results_ffhq_noisy},~\ref{tab:results_imagenet_noisy},~\ref{tab:results_inpainting_celeba}) and deterministic (Euler; Table~\ref{tab:results_dpde}) samplers, we apply it without modification to a sampler that alternates guided reverse diffusion with \emph{renoising}, a common ingredient of inverse-problem solvers such as RePaint \citep{lugmayr2022repaintinpaintingusingdenoising}, DDNM \citep{wang2022zero}, and DAPS \citep{zhang2025improvingdiffusioninverseproblem}.
App.~\ref{app:time_travel} shows that this sampling scheme when combined with LEEGS can perform competitively with the vanilla version of LEEGS on the FFHQ inverse problems (with better results for box and random inpainting), highlighting the flexibility of LEEGS. Moreover, the learned schedules for this scheme consist of a sequence of hill-shaped segments (Fig.~\ref{fig:time_travel_schedules}). Such a schedule would be hard to recover even with a priori knowledge, while the cost of learning them with LEEGS is unchanged.

\paragraph{Effect of inference steps.} LEEGS generally improves with more inference steps, with diminishing returns (Tables~\ref{tab:results_ffhq_noisy}--\ref{tab:results_dpde}; App.~\ref{app:cross_nfe}). As the budget grows, the mean guidance weight decreases monotonically (Fig.~\ref{fig:guidance_schedule_panel}): more steps need less aggressive per-step guidance. Since the schedule's shape is largely shared across budgets and its magnitude follows a predictable trend, a schedule trained at small $S$ can be interpolated and rescaled to larger $S$ at inference time, offering a way to reduce the training cost at large $S$ (App.~\ref{app:cross_nfe}).

\paragraph{Sample efficiency.} LEEGS is trained on only $30$ examples per subtask. Both the learned schedule and the resulting test performance are largely insensitive to the size of the training set, changing little beyond $30$ samples (App.~\ref{app:variance}).

\section{Related work}
\label{sec:related_work}

\paragraph{Guidance scheduling (for CFG).}

CFG \citep{ho2022classifierfreediffusionguidance} steers a conditional diffusion model toward a condition $y$ (e.g., a class label or text embedding) by combining the conditional and unconditional noise predictions $\eps_\theta(x_s, y)$ and $\eps_\theta(x_s, \emptyset)$ ($\emptyset$ the null condition) with a guidance scale $w$. As in our setting, $w$ must be chosen, and can be varied over the course of generation.
Several works vary the CFG scale over the trajectory. It was shown by \citet{kynkaanniemi2024applyingguidancelimitedinterval} that a nontrivial scale ($w \neq 1$) is only needed within a limited interval, with the conditional noise prediction $\eps_\theta(x_s, y)$ used alone elsewhere. Fixed schedules were compared and proposed by \citet{wang2024analysisclassifierfreeguidanceweight, sadat2024cadsunleashingdiversitydiffusion, malarz2025classifierfreeguidanceadaptivescaling}, and adaptive per-step scales have been derived from the two noise predictions \citep{zhu2026mambogmagnitudeawaremitigationboosted, pavasovic2025nonlinearguidance}, from estimated log-likelihoods \citep{koulischer2025feedbackguidancediffusionmodels}, or by a stepwise greedy search \citep{papalampidi2025dynamicclassifierfreediffusionguidance}.

Methods that \textit{learn} the guidance scale [e.g., \citep{yehezkel2025annealing, galashov2025learnguidediffusionmodel, sun2025ousacoptimizedguidancescheduling, azangulov2026adaptivediffusionguidancestochastic}] are closest to our work. \citet{azangulov2026adaptivediffusionguidancestochastic} is a primarily theoretical work in which the guidance scale is learned as a function of the sample $x_s$ by optimizing an optimal-control objective. \citet{yehezkel2025annealing, galashov2025learnguidediffusionmodel} learn guidance networks that predict the CFG guidance scale as a function of the intermediate sample $x_s$ and time $t_s$. The training objectives of these methods do not use rewards and instead rely on labeled data. \citet{sun2025ousacoptimizedguidancescheduling} uses an evolutionary search to learn a sparse guidance schedule supervised by a constant-guidance reference process; this is expensive, requiring many 1000-step reference rollouts. To our knowledge, LEEGS is the first to \textit{learn} guidance \textit{schedules} for loss-guided diffusion.

\paragraph{Constrained diffusion.}
Constrained diffusion methods interpret the guidance function $h$ as a constraint function and seek to minimize it, as in our setting. These include projection \citep{christopher2024constrainedsynthesisprojecteddiffusion, chen2026physicallyvalidbiomolecularinteraction, projected_christopher_2024}, trust-sampling \citep{huang2024constraineddiffusiontrustsampling}, proximal \citep{zampini2025trainingfreeconstrainedgenerationstable, jayaram2025CDIM}, and consensus \citep{christopher2026constraineddiffusionproteindesign, blanke2026strictlyconstrainedgenerativemodeling} methods. Projection-based methods ensure constraint satisfaction, but closed-form projections are typically only available for convex constraints, so several methods approximate them with an inner optimization loop \citep{neurosymbolic_christopher_2025}. Moreover, projections take arbitrarily large guidance steps, which can deviate substantially from the distribution induced by the diffusion model; this is typically mitigated by renoising the projected sample, which lengthens the sampling process. Proximal and trust-region sampling methods explicitly address the guidance trade-off by optimizing the constraint only up to a threshold, or within a fixed distance of the denoised iterate. However, the threshold or radius is itself a hyperparameter as delicate to set as the guidance scale.

Consensus-based methods take another approach, splitting inference into two trajectories (one applying projections, one following the unguided reverse process) and annealing them to coincide by the end of sampling. A guidance term implicitly pulls the trajectories together, and \citet{blanke2026strictlyconstrainedgenerativemodeling} sets its scale by dual ascent. Consensus methods are closely related to guidance and could be combined with LEEGS. Unlike the projection- and threshold-based methods above, though, LEEGS neither projects nor renoises: it learns a single schedule offline and applies it during ordinary guided sampling at no extra cost.

\section{Conclusion}
\label{sec:conclusion}

LEEGS is a simple, general method for learning time-dependent guidance schedules for loss-guided diffusion. It trains a schedule on as few as 30 examples by backpropagating the guidance loss through the sampler. Across image inpainting, noisy inverse problems, face-ID guidance, and PDE problems (\sect\ref{sec:experiments}), it is competitive with strong task-specific baselines and matches the 1000-NFE DPS reference using only 100 steps. Two findings stand out: a \emph{learned} schedule beats constant guidance at a matched budget, and the useful schedule shape is task-dependent, so it should be learned, not fixed by heuristic. LEEGS is also robust to sampler choice, working with stochastic (DDPM) and deterministic (Euler) samplers alike.

\paragraph{Limitations.} Training cost for LEEGS grows with the number of steps $S$ (the objective spans the full trajectory), making very large $S$ costly. While our gradient approximation (\sect\ref{sec:method}) is efficient, it may be less effective when the guidance field is highly sensitive to $x_s$.  Finally, while we learn a \emph{task-specific} schedule; adapting it to the condition or initial noise remains open.

\subsection*{Acknowledgments}
The work of Aneesh Barthakur is funded by Deutsche Forschungsgemeinschaft (DFG, German Research Foundation) under Germany's Excellence Strategy -- EXC 2075 -- 390740016. The work of L.F.O.\ Chamon is supported by the Agence Nationale de la Recherche (ANR) project ANR-25-CE23-3477-01 as well as a chair from Hi!PARIS, funded in part by the ANR AI Cluster 2030 and ANR-22-CMAS-0002. The authors thank the Stuttgart Center for Simulation Science (SimTech) and the International Max Planck Research School for Intelligent Systems (IMPRS-IS) for supporting Aneesh Barthakur. We acknowledge the computing time provided by Artificial Intelligence Software Academy (AISA), funded by the Ministry of Science, Research and the Arts Baden-W\"urttemberg.

\newpage

\section*{AI use statement}

In this work, we used LLM tools to implement and modify our core codebase, specifically parts of the LEEGS training script, samplers, guidance functions, heuristic baselines such as linear increasing,  experiment orchestration scripts, implementation of LEEGS with renoising etc. LLM tools were also used to write notebooks for plotting and analysis that were used to produce the paper's figures and tables, and they were also used to assist in the interpretation of results.
LLM tools were used to assist in writing and editing the main text of the paper, and were used to generate initial drafts of several appendix sections. AI tools were also used to search for relevant papers in the early stages of the project, in conjunction with conventional tools for literature search. 
We did not use generative AI tools to generate synthetic datasets; to develop theoretical models, conceptual frameworks, or for cleaning and reformating datasets.
The remaining required disclosure tasks--- formulating mathematical claims; providing critical ingredients for proving mathematical claims, assisting in the writing of proofs, assisting with translation are not applicable to this work. 
We have reviewed all AI-assisted work, including code, code outputs, and text. Critical parts of the code implementing the method or experiments were manually reviewed. All LLM generated text was manually reviewed and edited by the authors.  
We take responsibility for the final content of this work, including text, claims, and artifacts produced with the aid of generative AI.

\bibliography{E04_references.bib}

\begin{thebibliography}{61}
\providecommand{\natexlab}[1]{#1}
\providecommand{\url}[1]{\texttt{#1}}
\expandafter\ifx\csname urlstyle\endcsname\relax
  \providecommand{\doi}[1]{doi: #1}\else
  \providecommand{\doi}{doi: \begingroup \urlstyle{rm}\Url}\fi

\bibitem[Azangulov et~al.(2026)Azangulov, Potaptchik, Li, Aamari,
  Deligiannidis, and
  Rousseau]{azangulov2026adaptivediffusionguidancestochastic}
Iskander Azangulov, Peter Potaptchik, Qinyu Li, Eddie Aamari, George
  Deligiannidis, and Judith Rousseau.
\newblock Adaptive diffusion guidance via stochastic optimal control, 2026.

\bibitem[Bansal et~al.(2023)Bansal, Chu, Schwarzschild, Sengupta, Goldblum,
  Geiping, and Goldstein]{2023bansalunivguidance}
Arpit Bansal, Hong-Min Chu, Avi Schwarzschild, Soumyadip Sengupta, Micah
  Goldblum, Jonas Geiping, and Tom Goldstein.
\newblock Universal guidance for diffusion models.
\newblock In \emph{CVPR Workshops}, pp.\  843--852, 2023.

\bibitem[Bastek et~al.(2025)Bastek, Sun, and
  Kochmann]{bastek2025physicsinformed}
Jan-Hendrik Bastek, WaiChing Sun, and Dennis Kochmann.
\newblock Physics-informed diffusion models.
\newblock In \emph{The Thirteenth International Conference on Learning
  Representations}, 2025.
\newblock URL \url{https://openreview.net/forum?id=tpYeermigp}.

\bibitem[Blanke et~al.(2026)Blanke, Qu, Shamekh, and
  Gentine]{blanke2026strictlyconstrainedgenerativemodeling}
Matthieu Blanke, Yongquan Qu, Sara Shamekh, and Pierre Gentine.
\newblock Strictly constrained generative modeling via split augmented langevin
  sampling, 2026.

\bibitem[Chen et~al.(2024)Chen, Zhang, Zhou, and Chen]{towards_chen_2024}
Jian Chen, Ruiyi Zhang, Yufan Zhou, and Changyou Chen.
\newblock Towards aligned layout generation via diffusion model with aesthetic
  constraints.
\newblock \emph{arXiv.org}, 2024.
\newblock \doi{10.48550/ARXIV.2402.04754}.

\bibitem[Chen et~al.(2026)Chen, Guo, Wang, Chen, Zhang, Chai, Yang, Matusik,
  and Chen]{chen2026physicallyvalidbiomolecularinteraction}
Siyuan Chen, Minghao Guo, Caoliwen Wang, Anka~He Chen, Yikun Zhang, Jingjing
  Chai, Yin Yang, Wojciech Matusik, and Peter~Yichen Chen.
\newblock Physically valid biomolecular interaction modeling with gauss-seidel
  projection, 2026.

\bibitem[Christopher et~al.(2024{\natexlab{a}})Christopher, Baek, and
  Fioretto]{projected_christopher_2024}
Jacob Christopher, Stephen Baek, and Ferdinando Fioretto.
\newblock Projected generative diffusion models for constraint satisfaction.
\newblock \emph{arXiv.org}, 2024{\natexlab{a}}.
\newblock \doi{10.48550/ARXIV.2402.03559}.

\bibitem[Christopher et~al.(2025)Christopher, Cardei, Liang, and
  Fioretto]{neurosymbolic_christopher_2025}
Jacob Christopher, Michael Cardei, Jinhao Liang, and Ferdinando Fioretto.
\newblock Neuro-symbolic generative diffusion models for physically grounded,
  robust, and safe generation.
\newblock 2025.

\bibitem[Christopher et~al.(2024{\natexlab{b}})Christopher, Baek, and
  Fioretto]{christopher2024constrainedsynthesisprojecteddiffusion}
Jacob~K Christopher, Stephen Baek, and Ferdinando Fioretto.
\newblock Constrained synthesis with projected diffusion models,
  2024{\natexlab{b}}.

\bibitem[Christopher et~al.(2026)Christopher, Seamann, Cui, Khare, and
  Fioretto]{christopher2026constraineddiffusionproteindesign}
Jacob~K. Christopher, Austin Seamann, Jingyi Cui, Sagar Khare, and Ferdinando
  Fioretto.
\newblock Constrained diffusion for protein design with hard structural
  constraints, 2026.

\bibitem[Chung et~al.(2023)Chung, Kim, McCann, Klasky, and Ye]{chung2023DPS}
Hyungjin Chung, Jeongsol Kim, Michael~Thompson McCann, Marc~Louis Klasky, and
  Jong~Chul Ye.
\newblock Diffusion {P}osterior {S}ampling for {G}eneral {N}oisy {I}nverse
  {P}roblems.
\newblock In \emph{ICLR}. OpenReview.net, 2023.

\bibitem[Corso et~al.(2023)Corso, Xu, de~Bortoli, Barzilay, and
  Jaakkola]{corso2023particle}
Gabriele Corso, Yilun Xu, Valentin de~Bortoli, Regina Barzilay, and Tommi
  Jaakkola.
\newblock Particle guidance: non-i.i.d. diverse sampling with diffusion models.
\newblock \emph{arXiv}, 2023.

\bibitem[Deng et~al.(2019)Deng, Guo, Xue, and Zafeiriou]{deng2019arcface}
Jiankang Deng, Jia Guo, Niannan Xue, and Stefanos Zafeiriou.
\newblock Arcface: Additive angular margin loss for deep face recognition.
\newblock In \emph{CVPR}, pp.\  4690--4699. Computer Vision Foundation / IEEE,
  2019.

\bibitem[Dhariwal \& Nichol(2021)Dhariwal and
  Nichol]{dhariwal2021diffusionmodelsbeatgans}
Prafulla Dhariwal and Alex Nichol.
\newblock Diffusion models beat gans on image synthesis, 2021.

\bibitem[Efron(2011)]{efron2011tweedie}
Bradley Efron.
\newblock Tweedie's formula and selection bias.
\newblock \emph{Journal of the American Statistical Association}, 106\penalty0
  (496):\penalty0 1602--1614, 2011.

\bibitem[Galashov et~al.(2025)Galashov, Pokle, Doucet, Gretton, Delbracio, and
  Bortoli]{galashov2025learnguidediffusionmodel}
Alexandre Galashov, Ashwini Pokle, Arnaud Doucet, Arthur Gretton, Mauricio
  Delbracio, and Valentin~De Bortoli.
\newblock Learn to guide your diffusion model, 2025.

\bibitem[Heusel et~al.(2017)Heusel, Ramsauer, Unterthiner, Nessler, and
  Hochreiter]{heusel2017gans}
Martin Heusel, Hubert Ramsauer, Thomas Unterthiner, Bernhard Nessler, and Sepp
  Hochreiter.
\newblock {GANs} trained by a two time-scale update rule converge to a local
  nash equilibrium.
\newblock In \emph{Advances in Neural Information Processing Systems
  (NeurIPS)}, 2017.

\bibitem[Ho \& Salimans(2022)Ho and
  Salimans]{ho2022classifierfreediffusionguidance}
Jonathan Ho and Tim Salimans.
\newblock Classifier-free diffusion guidance, 2022.

\bibitem[Ho et~al.(2020)Ho, Jain, and
  Abbeel]{ho2020denoisingdiffusionprobabilisticmodels}
Jonathan Ho, Ajay Jain, and Pieter Abbeel.
\newblock Denoising diffusion probabilistic models, 2020.

\bibitem[Hoogeboom et~al.(2022)Hoogeboom, Satorras, Vignac, and
  Welling]{hoogeboom2022equivariantdiffusionmoleculegeneration}
Emiel Hoogeboom, Victor~Garcia Satorras, Clément Vignac, and Max Welling.
\newblock Equivariant diffusion for molecule generation in 3d, 2022.

\bibitem[Huang et~al.(2024{\natexlab{a}})Huang, Yang, Wang, and
  Park]{huang2024dpde}
Jiahe Huang, Guandao Yang, Zichen Wang, and Jeong~Joon Park.
\newblock Diffusionpde: Generative pde-solving under partial observation,
  2024{\natexlab{a}}.

\bibitem[Huang et~al.(2024{\natexlab{b}})Huang, Jiang, Wouwe, and
  Liu]{huang2024constraineddiffusiontrustsampling}
William Huang, Yifeng Jiang, Tom~Van Wouwe, and C.~Karen Liu.
\newblock Constrained diffusion with trust sampling, 2024{\natexlab{b}}.

\bibitem[Jayaram et~al.(2025)Jayaram, Kemelmacher-Shlizerman, Seitz, and
  Thickstun]{jayaram2025CDIM}
Vivek Jayaram, Ira Kemelmacher-Shlizerman, Steven~M. Seitz, and John Thickstun.
\newblock Linearly constrained diffusion implicit models, 2025.

\bibitem[Karras et~al.(2022)Karras, Aittala, Aila, and
  Laine]{karras2022elucidatingdesignspacediffusionbased}
Tero Karras, Miika Aittala, Timo Aila, and Samuli Laine.
\newblock Elucidating the design space of diffusion-based generative models,
  2022.

\bibitem[Kawar et~al.(2022)Kawar, Elad, Ermon, and Song]{kawar2022ddrm}
Bahjat Kawar, Michael Elad, Stefano Ermon, and Jiaming Song.
\newblock Denoising diffusion restoration models.
\newblock In Sanmi Koyejo, S.~Mohamed, A.~Agarwal, Danielle Belgrave, K.~Cho,
  and A.~Oh (eds.), \emph{NeurIPS}, 2022.

\bibitem[Kingma \& Welling(2013)Kingma and Welling]{kingma2013autoencoding}
Diederik~P Kingma and Max Welling.
\newblock Auto-encoding variational bayes, 2013.

\bibitem[Koulischer et~al.(2025)Koulischer, Handke, Deleu, Demeester, and
  Ambrogioni]{koulischer2025feedbackguidancediffusionmodels}
Felix Koulischer, Florian Handke, Johannes Deleu, Thomas Demeester, and Luca
  Ambrogioni.
\newblock Feedback guidance of diffusion models, 2025.

\bibitem[Kumar et~al.(2019)Kumar, Purohit, Svitkina, Vee, and
  Wang]{kumar2019remat}
Ravi Kumar, Manish Purohit, Zoya Svitkina, Erik Vee, and Joshua~R. Wang.
\newblock Efficient rematerialization for deep networks.
\newblock In Hanna~M. Wallach, Hugo Larochelle, Alina Beygelzimer, Florence
  d'Alché Buc, Emily~B. Fox, and Roman Garnett (eds.), \emph{NeurIPS}, pp.\
  15146--15155, 2019.

\bibitem[Kynkäänniemi et~al.(2024)Kynkäänniemi, Aittala, Karras, Laine,
  Aila, and Lehtinen]{kynkaanniemi2024applyingguidancelimitedinterval}
Tuomas Kynkäänniemi, Miika Aittala, Tero Karras, Samuli Laine, Timo Aila, and
  Jaakko Lehtinen.
\newblock Applying guidance in a limited interval improves sample and
  distribution quality in diffusion models, 2024.

\bibitem[Langtangen \& Mardal(2019)Langtangen and
  Mardal]{langtangen2019numericalbook}
Hans~Petter Langtangen and Kent-Andre Mardal.
\newblock \emph{Introduction to Numerical Methods for Variational Problems}.
\newblock Springer International Publishing, 2019.
\newblock ISBN 9783030237882.
\newblock \doi{10.1007/978-3-030-23788-2}.
\newblock URL \url{http://dx.doi.org/10.1007/978-3-030-23788-2}.

\bibitem[Li et~al.(2020)Li, Kovachki, Azizzadenesheli, Liu, Bhattacharya,
  Stuart, and Anandkumar]{li2020fourier}
Zongyi Li, Nikola Kovachki, Kamyar Azizzadenesheli, Burigede Liu, Kaushik
  Bhattacharya, Andrew Stuart, and Anima Anandkumar.
\newblock Fourier neural operator for parametric partial differential
  equations, 2020.

\bibitem[Lugmayr et~al.(2022)Lugmayr, Danelljan, Romero, Yu, Timofte, and
  Gool]{lugmayr2022repaintinpaintingusingdenoising}
Andreas Lugmayr, Martin Danelljan, Andres Romero, Fisher Yu, Radu Timofte, and
  Luc~Van Gool.
\newblock Repaint: Inpainting using denoising diffusion probabilistic models,
  2022.

\bibitem[Malarz et~al.(2025)Malarz, Kasymov, Zieba, Tabor, and
  Spurek]{malarz2025classifierfreeguidanceadaptivescaling}
Dawid Malarz, Artur Kasymov, Maciej Zieba, Jacek Tabor, and Przemyslaw Spurek.
\newblock Classifier-free guidance with adaptive scaling, 2025.

\bibitem[Maze \& Ahmed(2023)Maze and Ahmed]{maze2022topodiff}
Francois Maze and Faez Ahmed.
\newblock Diffusion models beat gans on topology optimization.
\newblock In \emph{Proceedings of the AAAI Conference on Artificial
  Intelligence}, 2023.

\bibitem[Nichol et~al.(2022)Nichol, Dhariwal, Ramesh, Shyam, Mishkin, McGrew,
  Sutskever, and Chen]{nichol2022glide}
Alexander~Quinn Nichol, Prafulla Dhariwal, Aditya Ramesh, Pranav Shyam, Pamela
  Mishkin, Bob McGrew, Ilya Sutskever, and Mark Chen.
\newblock Glide: Towards photorealistic image generation and editing with
  text-guided diffusion models.
\newblock In Kamalika Chaudhuri, Stefanie Jegelka, Le~Song, Csaba Szepesvári,
  Gang Niu, and Sivan Sabato (eds.), \emph{ICML}, volume 162 of
  \emph{Proceedings of Machine Learning Research}, pp.\  16784--16804. PMLR,
  2022.

\bibitem[Papalampidi et~al.(2025)Papalampidi, Wiles, Ktena, Shtedritski,
  Bugliarello, Kajic, Albuquerque, and
  Nematzadeh]{papalampidi2025dynamicclassifierfreediffusionguidance}
Pinelopi Papalampidi, Olivia Wiles, Ira Ktena, Aleksandar Shtedritski, Emanuele
  Bugliarello, Ivana Kajic, Isabela Albuquerque, and Aida Nematzadeh.
\newblock Dynamic classifier-free diffusion guidance via online feedback, 2025.

\bibitem[Pavasovic et~al.(2025)Pavasovic, Verbeek, Biroli, and
  Mezard]{pavasovic2025nonlinearguidance}
Krunoslav~Lehman Pavasovic, Jakob Verbeek, Giulio Biroli, and Marc Mezard.
\newblock Classifier-free guidance: From high-dimensional analysis to
  generalized guidance forms, 2025.

\bibitem[Raissi et~al.(2019)Raissi, Perdikaris, and
  Karniadakis]{raissi2019pinns}
M.~Raissi, P.~Perdikaris, and G.E. Karniadakis.
\newblock Physics-informed neural networks: {A} deep learning framework for
  solving forward and inverse problems involving nonlinear partial differential
  equations.
\newblock \emph{Journal of Computational Physics}, 378:\penalty0 686--707,
  2019.

\bibitem[Rombach et~al.(2022)Rombach, Blattmann, Lorenz, Esser, and
  Ommer]{rombach2022LDM}
Robin Rombach, Andreas Blattmann, Dominik Lorenz, Patrick Esser, and Björn
  Ommer.
\newblock High-resolution image synthesis with latent diffusion models.
\newblock In \emph{CVPR}, pp.\  10674--10685. IEEE, 2022.

\bibitem[Sadat et~al.(2024)Sadat, Buhmann, Bradley, Hilliges, and
  Weber]{sadat2024cadsunleashingdiversitydiffusion}
Seyedmorteza Sadat, Jakob Buhmann, Derek Bradley, Otmar Hilliges, and Romann~M.
  Weber.
\newblock Cads: Unleashing the diversity of diffusion models through
  condition-annealed sampling, 2024.

\bibitem[Song et~al.(2022)Song, Meng, and
  Ermon]{song2022denoisingdiffusionimplicitmodels}
Jiaming Song, Chenlin Meng, and Stefano Ermon.
\newblock Denoising diffusion implicit models, 2022.

\bibitem[Song et~al.(2021)Song, Sohl-Dickstein, Kingma, Kumar, Ermon, and
  Poole]{song2021scorebasedgenerativemodelingstochastic}
Yang Song, Jascha Sohl-Dickstein, Diederik~P. Kingma, Abhishek Kumar, Stefano
  Ermon, and Ben Poole.
\newblock Score-based generative modeling through stochastic differential
  equations, 2021.

\bibitem[Sun et~al.(2025)Sun, Yang, Niu, and
  Sun]{sun2025ousacoptimizedguidancescheduling}
Ruitong Sun, Tianze Yang, Wei Niu, and Jin Sun.
\newblock Ousac: Optimized guidance scheduling with adaptive caching for dit
  acceleration, 2025.

\bibitem[Suvorov et~al.(2022)Suvorov, Logacheva, Mashikhin, Remizova, Ashukha,
  Silvestrov, Kong, Goka, Park, and Lempitsky]{suvorov2022lama}
Roman Suvorov, Elizaveta Logacheva, Anton Mashikhin, Anastasia Remizova,
  Arsenii Ashukha, Aleksei Silvestrov, Naejin Kong, Harshith Goka, Kiwoong
  Park, and Victor Lempitsky.
\newblock Resolution-robust large mask inpainting with fourier convolutions.
\newblock In \emph{WACV}, pp.\  3172--3182. IEEE, 2022.

\bibitem[Tang et~al.(2024)Tang, Peng, Tang, Hong, Wang, and
  Chang]{tang2024inferencetimealignmentdiffusionmodels}
Zhiwei Tang, Jiangweizhi Peng, Jiasheng Tang, Mingyi Hong, Fan Wang, and
  Tsung-Hui Chang.
\newblock Inference-time alignment of diffusion models with direct noise
  optimization, 2024.

\bibitem[Tong et~al.(2025)Tong, Trung-Dung, Liu, den Broeck, and
  Niepert]{tong2025learningdiscretizedenoisingdiffusion}
Vinh Tong, Hoang Trung-Dung, Anji Liu, Guy~Van den Broeck, and Mathias Niepert.
\newblock Learning to discretize denoising diffusion odes, 2025.

\bibitem[Trefethen(1996)]{trefethen1996finite}
Lloyd~N. Trefethen.
\newblock \emph{Finite Difference and Spectral Methods for Ordinary and Partial
  Differential Equations}.
\newblock 1996.
\newblock URL \url{http://people.maths.ox.ac.uk/trefethen/pdetext.html}.
\newblock Unpublished text.

\bibitem[Wang et~al.(2024)Wang, Dufour, Andreou, Cani, Abrevaya, Picard, and
  Kalogeiton]{wang2024analysisclassifierfreeguidanceweight}
Xi~Wang, Nicolas Dufour, Nefeli Andreou, Marie-Paule Cani, Victoria~Fernandez
  Abrevaya, David Picard, and Vicky Kalogeiton.
\newblock Analysis of classifier-free guidance weight schedulers, 2024.

\bibitem[Wang et~al.(2023)Wang, Yu, and Zhang]{wang2022zero}
Yinhuai Wang, Jiwen Yu, and Jian Zhang.
\newblock Zero-shot image restoration using denoising diffusion null-space
  model.
\newblock \emph{The Eleventh International Conference on Learning
  Representations}, 2023.

\bibitem[Watson et~al.(2022)Watson, Chan, Ho, and
  Norouzi]{watson2022learningfastsamplersdiffusion}
Daniel Watson, William Chan, Jonathan Ho, and Mohammad Norouzi.
\newblock Learning fast samplers for diffusion models by differentiating
  through sample quality, 2022.

\bibitem[Watson et~al.(2023)Watson, Juergens, Bennett, Trippe, Yim, Eisenach,
  Ahern, Borst, Ragotte, Milles, Wicky, Hanikel, Pellock, Courbet, Sheffler,
  Wang, Venkatesh, Sappington, Torres, Lauko, De~Bortoli, Mathieu, Ovchinnikov,
  Barzilay, Jaakkola, DiMaio, Baek, and Baker]{watson2023rfdiff}
Joseph~L. Watson, David Juergens, Nathaniel~R. Bennett, Brian~L. Trippe, Jason
  Yim, Helen~E. Eisenach, Woody Ahern, Andrew~J. Borst, Robert~J. Ragotte,
  Lukas~F. Milles, Basile I.~M. Wicky, Nikita Hanikel, Samuel~J. Pellock,
  Alexis Courbet, William Sheffler, Jue Wang, Preetham Venkatesh, Isaac
  Sappington, Susana~Vázquez Torres, Anna Lauko, Valentin De~Bortoli, Emile
  Mathieu, Sergey Ovchinnikov, Regina Barzilay, Tommi~S. Jaakkola, Frank
  DiMaio, Minkyung Baek, and David Baker.
\newblock De novo design of protein structure and function with rfdiffusion.
\newblock \emph{Nature}, 620\penalty0 (7976):\penalty0 1089--1100, 2023.

\bibitem[Wu et~al.(2023)Wu, Trippe, Naesseth, Blei, and
  Cunningham]{wu2023practical}
Luhuan Wu, Brian~L Trippe, Christian Naesseth, David Blei, and John~P
  Cunningham.
\newblock Practical and asymptotically exact conditional sampling in diffusion
  models.
\newblock \emph{Advances in Neural Information Processing Systems}, 2023.

\bibitem[Xu et~al.(2022)Xu, Yu, Song, Shi, Ermon, and Tang]{xu2022geodiff}
Minkai Xu, Lantao Yu, Yang Song, Chence Shi, Stefano Ermon, and Jian Tang.
\newblock Geodiff: A geometric diffusion model for molecular conformation
  generation.
\newblock In \emph{ICLR}, 2022.

\bibitem[Yang et~al.(2024)Yang, Ding, Cai, Yu, Wang, and Shi]{yang2024dsg}
Lingxiao Yang, Shutong Ding, Yifan Cai, Jingyi Yu, Jingya Wang, and Ye~Shi.
\newblock Guidance with {S}pherical {G}aussian {C}onstraint for {C}onditional
  {D}iffusion.
\newblock In \emph{ICML}, volume 235 of \emph{Proceedings of Machine Learning
  Research}, pp.\  56071--56095, 2024.

\bibitem[Yehezkel et~al.(2025)Yehezkel, Dahary, Voynov, and
  Cohen-Or]{yehezkel2025annealing}
Shai Yehezkel, Omer Dahary, Andrey Voynov, and Daniel Cohen-Or.
\newblock Navigating with annealing guidance scale in diffusion space.
\newblock In \emph{Proceedings of the SIGGRAPH Asia 2025 Conference Papers}, SA
  Conference Papers '25, pp.\  1--11. ACM, December 2025.
\newblock \doi{10.1145/3757377.3763830}.

\bibitem[Yu et~al.(2023)Yu, Wang, Zhao, Ghanem, and Zhang]{yu2023freedom}
Jiwen Yu, Yinhuai Wang, Chen Zhao, Bernard Ghanem, and Jian Zhang.
\newblock Freedom: Training-free energy-guided conditional diffusion model.
\newblock In \emph{ICCV}. IEEE, 2023.

\bibitem[Zampini et~al.(2025)Zampini, Christopher, Oneto, Anguita, and
  Fioretto]{zampini2025trainingfreeconstrainedgenerationstable}
Stefano Zampini, Jacob~K. Christopher, Luca Oneto, Davide Anguita, and
  Ferdinando Fioretto.
\newblock Training-free constrained generation with stable diffusion models,
  2025.

\bibitem[Zhai et~al.(2025)Zhai, Singh, Thatipelli, Chakraborty, Sahu, Huang,
  Bedi, and Shah]{zhai2025mira}
Kevin Zhai, Utsav Singh, Anirudh Thatipelli, Souradip Chakraborty, Anit~Kumar
  Sahu, Furong Huang, Amrit~Singh Bedi, and Mubarak Shah.
\newblock Mira: Towards mitigating reward hacking in inference-time alignment
  of t2i diffusion models.
\newblock \emph{CoRR}, October 2025.

\bibitem[Zhang et~al.(2025)Zhang, Chu, Berner, Meng, Anandkumar, and
  Song]{zhang2025improvingdiffusioninverseproblem}
Bingliang Zhang, Wenda Chu, Julius Berner, Chenlin Meng, Anima Anandkumar, and
  Yang Song.
\newblock Improving {{Diffusion Inverse Problem Solving}} with {{Decoupled
  Noise Annealing}}.
\newblock In \emph{Proceedings of the {{IEEE}}/{{CVF}} Conference on Computer
  Vision and Pattern Recognition ({{CVPR}})}, pp.\  20895--20905, 2025.

\bibitem[Zhang et~al.(2018)Zhang, Isola, Efros, Shechtman, and
  Wang]{zhang2018perceptual}
Richard Zhang, Phillip Isola, Alexei~A Efros, Eli Shechtman, and Oliver Wang.
\newblock The unreasonable effectiveness of deep features as a perceptual
  metric.
\newblock In \emph{CVPR}, 2018.

\bibitem[Zhu et~al.(2026)Zhu, Peng, Shu, Hu, Yang, Zhang, Pu, Zheng, Cui, Zhao,
  Feng, and Cheng]{zhu2026mambogmagnitudeawaremitigationboosted}
Shangwen Zhu, Qianyu Peng, Zhilei Shu, Yuting Hu, Zhantao Yang, Han Zhang, Zhao
  Pu, Andy Zheng, Xinyu Cui, Jian Zhao, Ruili Feng, and Fan Cheng.
\newblock Mambo-g: Magnitude-aware mitigation for boosted guidance, 2026.

\end{thebibliography}

\newpage

\begin{center}
    \hrule
    \color{indigo}
    \startcontents[sections]\vbox{\vspace{4mm}\LARGE \textsc{Learned End-to-End Guidance Schedules for Diffusion Models} \\
    \textsc{\small \textbf{Additional Material}}} \vspace{5mm} \hrule height .5pt
    \printcontents[sections]{l}{0}{\setcounter{tocdepth}{2}}
\end{center}

\newpage
\appendix
\section{Pseudocode for LEEGS training}
\label{app:pseudocode}

\begin{algorithm}[h]
    \captionof{algorithm}{LEEGS training algorithm}
    \label{alg:leegs}
    \begin{algorithmic}[1]
        \State \textbf{Input:} training conditions $\mathcal{D} = \{c^{(i)}\}_{i=1}^{N}$, steps $S$, epochs $E$, flag \texttt{compute\_leegs\_approx}
        \State \textbf{Output:} guidance schedule $W$
        \State $W \gets (w_1, \ldots, w_S)$ with $w_s \sim \mathcal{U}[0, 1]$ i.i.d. \Comment{schedule initialization}
        \For{$\text{epoch} = 1, \ldots, E$}
            \For{minibatch $\mathcal{B} \subseteq \mathcal{D}$}
                \State $\bar{g} \gets \tfrac{1}{|\mathcal{B}|} \sum_{c \in \mathcal{B}} \textsc{ComputeGradient}(W,\ c,\ \texttt{compute\_leegs\_approx})$
                \State $W \gets \textsc{AdamStep}(W,\ \bar{g})$
                \State $W \gets \max(W,\ 0)$ \Comment{project onto $\mathbb{R}^S_{\ge 0}$}
            \EndFor
        \EndFor
        \State \Return $W$
        \Statex
        \Function{ComputeGradient}{$W, c, \texttt{compute\_leegs\_approx}$}
            \State $x \sim \mathcal{N}(0, \mathbf{I})$
            \State $\ell \gets 0$
            \For{$s = S, \ldots, 1$}
                \State $t \gets t_s$ \Comment{diffusion time at step $s$}
                \State $(x, h_s) \gets \textsc{Checkpoint}\big(\textsc{GuidedStep},\ x,\ t,\ w_s,\ c,\ \texttt{compute\_leegs\_approx}\big)$
                \State $\ell \gets \ell + h_s$ \Comment{add intermediate term $h(\hat{x}_{0|s}, c)$}
            \EndFor
            \State $h_0$ $\gets \texttt{compute\_guidance\_function}(x, c)$ \Comment{terminal term $h(x_0, c)$}
            \State $\ell \gets \ell + h_0$
            \State \Return $\textsc{Backward}(\ell,\ W)$ \Comment{populate leaf nodes of autodiff graph with gradients}
        \EndFunction
        \Statex
        \Function{GuidedStep}{$x, t, w, c, \texttt{compute\_leegs\_approx}$}
            \State $(\texttt{denoised\_x},\ \texttt{next\_x}) \gets \texttt{sampler\_step}(x, t)$
            \State $h \gets \texttt{compute\_guidance\_function}(\texttt{denoised\_x}, c)$
            \State $g \gets \textsc{Grad}\big(h,\ x;\ \texttt{create\_graph} = \textbf{not}\ \texttt{compute\_leegs\_approx}\big)$
            \If{\texttt{compute\_leegs\_approx}} \label{line:sg}
                \State $g \gets g\texttt{.detach()}$ \Comment{drop $\partial g / \partial x$; keep $\partial \Phi / \partial x$}
            \EndIf
            \State $\texttt{next\_x} \gets \texttt{next\_x} - w\, g$ \Comment{guided update, \eqref{eqn:guided_sampling}}
            \State \Return $(\texttt{next\_x},\ h)$
        \EndFunction
    \end{algorithmic}
\end{algorithm}

Algorithm~\ref{alg:leegs} gives pseudocode for training a guidance schedule with LEEGS. The
outer loop performs Adam updates over epochs and minibatches of training conditions, projecting
$W$ onto $\mathbb{R}^S_{\ge 0}$ after every step. \textsc{ComputeGradient} evaluates the inner
objective over a freshly sampled trajectory and returns its gradient with respect to $W$, using
gradient checkpointing to keep the peak memory independent of $S$. \textsc{GuidedStep} applies one
guided reverse-diffusion step \eqref{eqn:guided_sampling} and returns the updated iterate together
with that step's guidance value $h_s$. We implement this in PyTorch using the built-in
checkpointing functionality (\texttt{torch.utils.checkpoint}).

\paragraph{Primitives.} We assume the following building blocks, each acting on a single sampling step:
\begin{itemize}
    \item $\texttt{sampler\_step}(x, t)$ -- returns the pair $(\texttt{denoised\_x},\ \texttt{next\_x})$,
    where $\texttt{denoised\_x} = \hat{x}_0(x, t; \eps_\theta)$ is the denoiser prediction ($\hat{x}_{0|s}$ when $x = x_s$ and $t = t_s$) and $\texttt{next\_x} = \Phi(x, t; \eps_\theta)$ is the unguided reverse-diffusion step of \eqref{eqn:unguided_sampling}. Recall that the (DDIM) forward noising process relates a clean sample to its noised iterate via $x_s = \sqrt{\abar_s}\, x_0 + \sqrt{1-\abar_s}\, \eps$ for $\eps \sim \mathcal{N}(0, \mathbf{I})$, where $\abar_s = \prod_{i=1}^{s} \a_i$ is the cumulative product of the (per-step) variance-schedule coefficients $\a_i$; accordingly $\sqrt{\abar_s}$ and $\sqrt{1-\abar_s}$ are the signal and noise coefficients at time $t_s$. When the network is parameterized as a noise predictor $\eps_\theta$, the denoiser prediction is recovered from it via Tweedie's formula \citep{efron2011tweedie},
    \begin{align*}
        \hat{x}_0(x_s, t_s; \eps_\theta) = \frac{x_s - \sqrt{1-\abar_s}\, \eps_\theta(x_s, t_s)}{\sqrt{\abar_s}}. \mlabel{eqn:tweedie}
    \end{align*}
    Many diffusion models instead predict the denoised sample $\hat{x}_0$ directly, in which case $\texttt{denoised\_x}$ is simply read off the network output and \eqref{eqn:tweedie} is unnecessary.
    \item $\texttt{compute\_guidance\_function}(x, c)$ -- evaluates the guidance function $h(x, c)$ on a (denoised) sample.
    \item Autodiff primitives, which map directly onto their PyTorch counterparts:
    \begin{itemize}
        \item $\textsc{Grad}(y, x;\ \texttt{create\_graph})$ -- the gradient $\nabla_x y$ of a scalar $y$
        w.r.t.\ a tensor $x$ (\texttt{torch.autograd.grad}). The flag $\texttt{create\_graph}$ retains
        the graph of the gradient itself, so that its Jacobian can be differentiated in a later
        backward pass.
        \item $\textsc{Backward}(y, x)$ -- the reverse-mode gradient $\nabla_x y$ accumulated through the retained graph (\texttt{y.backward()} followed by reading \texttt{x.grad}).
        \item $x\texttt{.detach()}$ -- returns $x$ detached from the autodiff graph
        (\texttt{torch.Tensor.detach}).
        \item $\textsc{Checkpoint}(f, \ldots)$ -- evaluates $f$ without storing its internal
        activations, recomputing them during the backward pass
        (\texttt{torch.utils.checkpoint.checkpoint}).
    \end{itemize}
\end{itemize}

In \textsc{GuidedStep}, the flag \texttt{compute\_leegs\_approx} controls whether the guidance
gradient $g_s$ retains its graph: with $\texttt{compute\_leegs\_approx} = \texttt{false}$
(\texttt{full}) nothing is detached and the guidance Hessian $\partial g_s / \partial x_s$ in
\eqref{eqn:grad_x_s} contributes to the gradient; with $\texttt{compute\_leegs\_approx} =
\texttt{true}$ (\texttt{LEEGS}) $g_s$ is treated as a constant of $x_s$, so only the sampler
Jacobian $\partial \Phi / \partial x_s$ is propagated across steps. The two settings differ only in
the $\texttt{.detach()}$ on line~\ref{line:sg}.

\clearpage
\newpage
\section{Additional experimental details}
\label{app:exp_details}

Unless noted otherwise, LEEGS is trained with Adam and batch size 1, separately for each subtask and NFE budget, on 30 examples randomly drawn from a training pool that is disjoint from the test set. The number of epochs is chosen from the convergence of the training objective. Our implementation builds on the codebases of \citet{chung2023DPS,huang2024dpde,yang2024dsg}.

\subsection{Noisy inverse problems}
\label{app:exp_noisy_inverse}

\begin{figure}[ht]
    \centering
    \includegraphics[width=\textwidth]{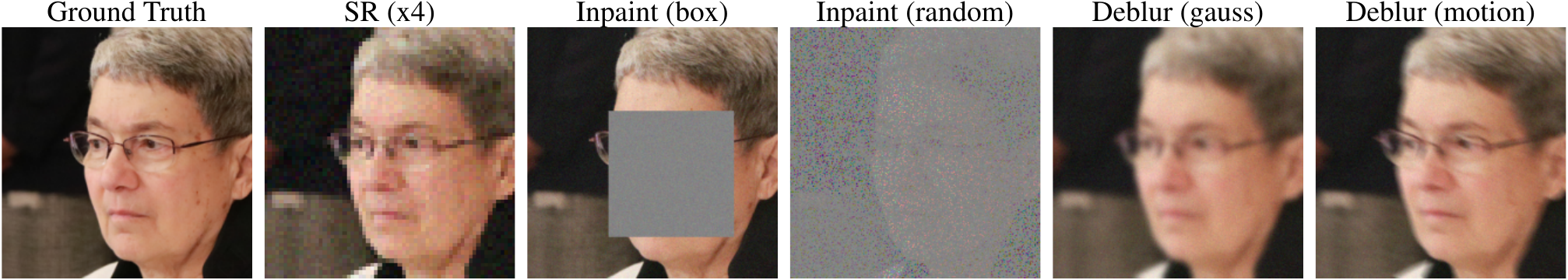}
    \caption{Ground truth and the 5 noisy measurement operators used for Tables \ref{tab:results_ffhq_noisy} and \ref{tab:results_imagenet_noisy} (FFHQ example).}
    \label{fig:task_illustration_noisy_inverse}
\end{figure}

\paragraph{Models.} Following \citet{chung2023DPS}, we use the pretrained FFHQ model of \citet{chung2023DPS} and the pretrained ImageNet model of \citet{dhariwal2021diffusionmodelsbeatgans}.

\paragraph{Dataset generation.} We follow the implementation and settings of \citet{chung2023DPS} for both FFHQ and ImageNet, which we recap for completeness; Figure~\ref{fig:task_illustration_noisy_inverse} illustrates the five measurement operators. For superresolution, the low-resolution images are generated by bicubic downsampling at $4\times$ scale. For Gaussian deblurring, the images are convolved with a $61\times61$ Gaussian kernel with standard deviation $3.0$. For motion deblurring, the images are convolved with a $61\times61$ motion-blur kernel of intensity $0.5$, generated with the implementation from \url{https://github.com/LeviBorodenko/motionblur}. For box and random inpainting, the masks are sampled randomly for each image, where for box inpainting the mask is an approximately centered $128\times128$ square, and for random inpainting the mask is a random binary mask with each pixel masked with a probability of $0.92$. 

For FFHQ, we use the test set of $1000$ images of \citet{chung2023DPS} for training and validation. For the other datasets, we set aside a disjoint pool of $100$ images ($90$ for training and $10$ for validation metrics during training), but for FFHQ we use a cross-validation-style split of the $1000$ images instead, to maintain a reasonable test set size : the first $50$ images are set aside, of which the first $30$ are used to train LEEGS and the remaining $20$ for validation, and the other $950$ images form the test set of Table~\ref{tab:results_ffhq_noisy}. For ImageNet, we build our own training and test sets. The test set of Table~\ref{tab:results_imagenet_noisy} consists of $1000$ images sampled from the ILSVRC 2012 validation set, one per class, and a disjoint pool of $100$ images is split into $90$ images for training, from which each LEEGS run draws $30$, and $10$ for validation.

\begin{figure}[ht]
    \centering
    \includegraphics[width=0.9\textwidth]{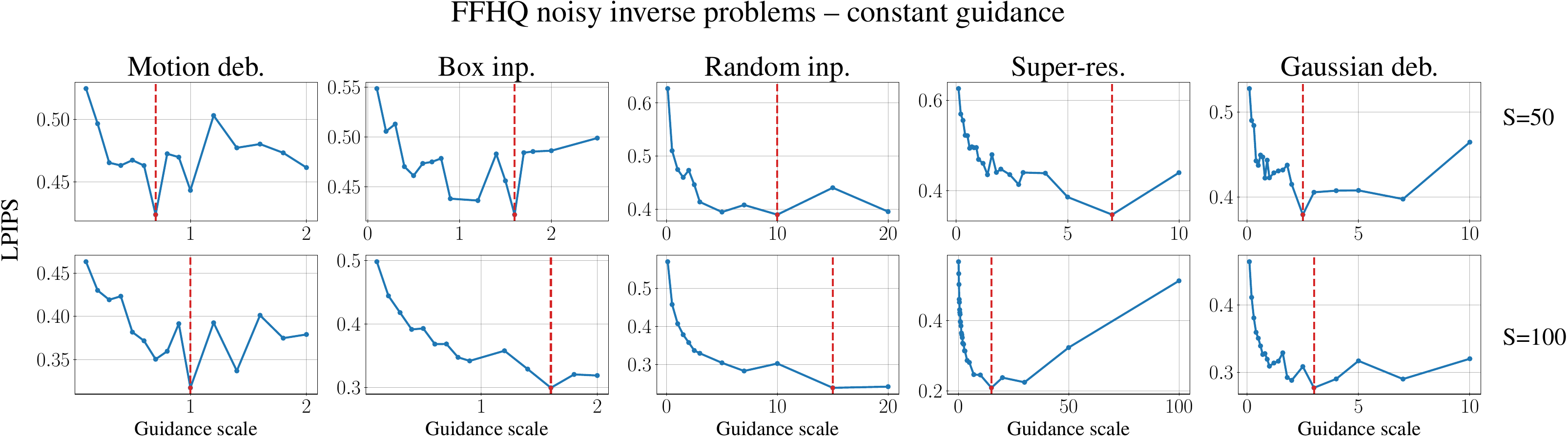}
    \caption{Constant guidance, FFHQ: LPIPS vs.\ guidance scale; best value in red.}
    \label{fig:grid_search_dps_ffhq}
\end{figure}

\begin{figure}[ht]
    \centering
    \includegraphics[width=0.9\textwidth]{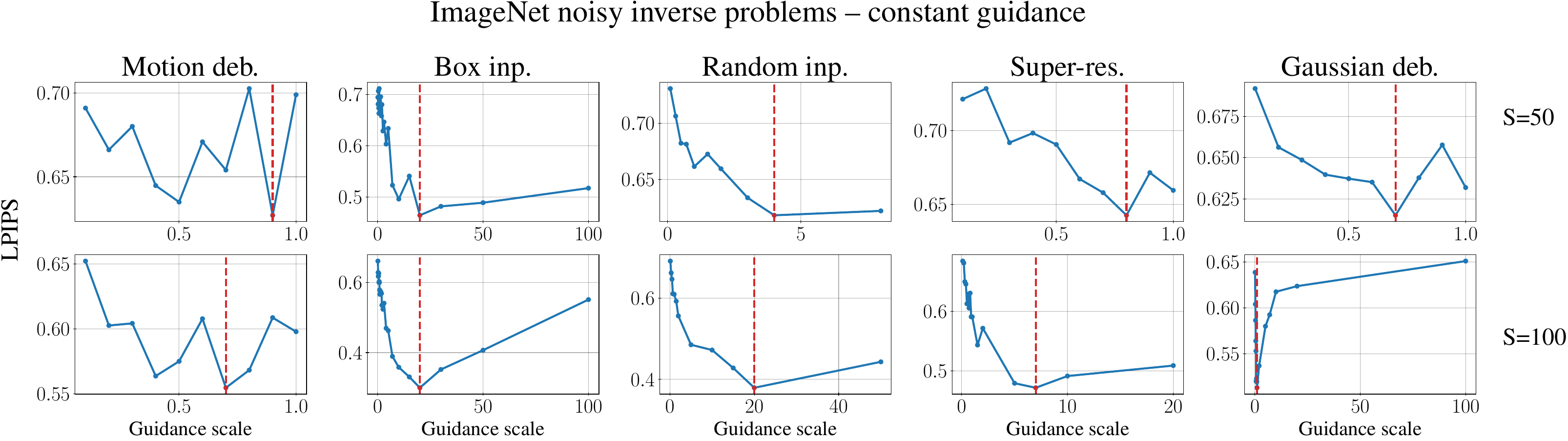}
    \caption{Constant guidance, ImageNet: LPIPS vs.\ guidance scale; best value in red.}
    \label{fig:grid_search_dps_imagenet}
\end{figure}

\begin{figure}[ht]
    \centering
    \includegraphics[width=0.9\textwidth]{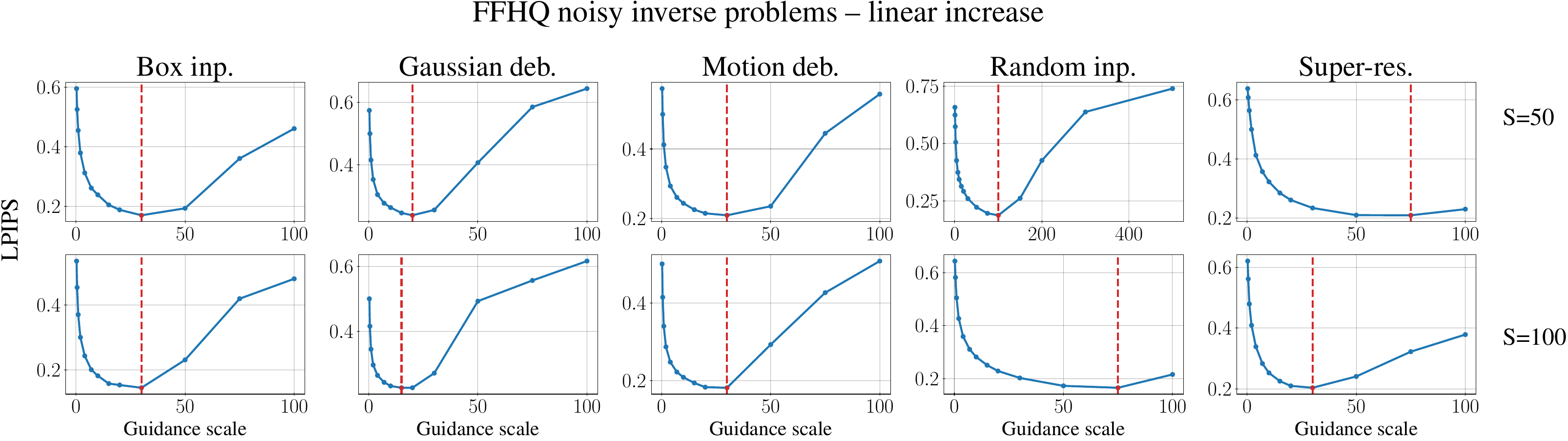}
    \caption{Linear increase, FFHQ: LPIPS vs.\ guidance scale; best value in red.}
    \label{fig:grid_search_linear_increase_ffhq}
\end{figure}

\begin{figure}[ht]
    \centering
    \includegraphics[width=0.9\textwidth]{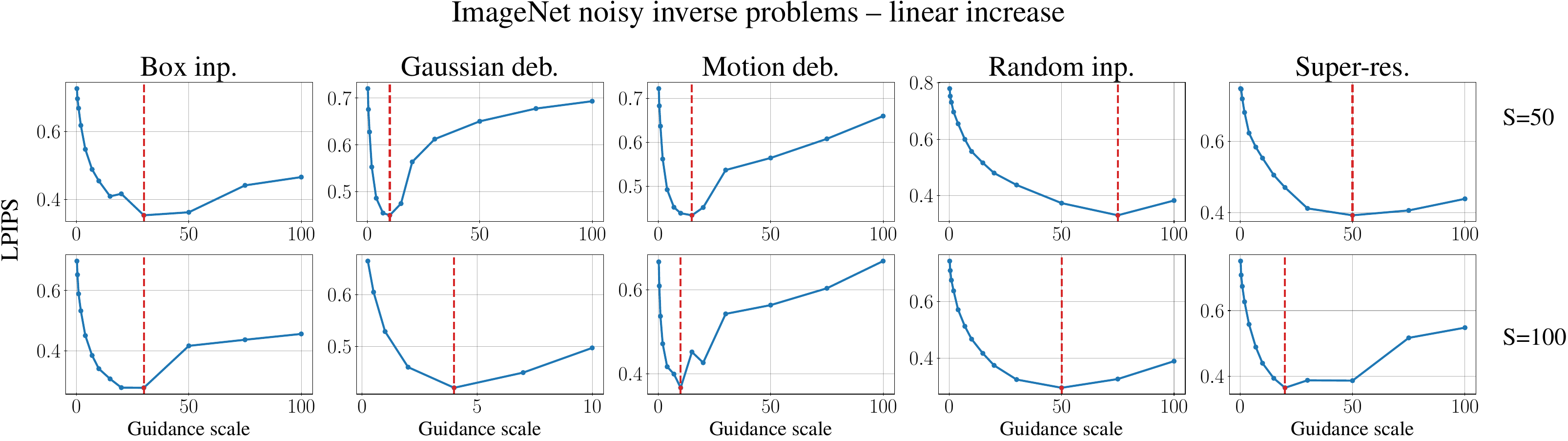}
    \caption{Linear increase, ImageNet: LPIPS vs.\ guidance scale; best value in red.}
    \label{fig:grid_search_linear_increase_imagenet}
\end{figure}

\begin{figure}[ht]
    \centering
    \includegraphics[width=0.9\textwidth]{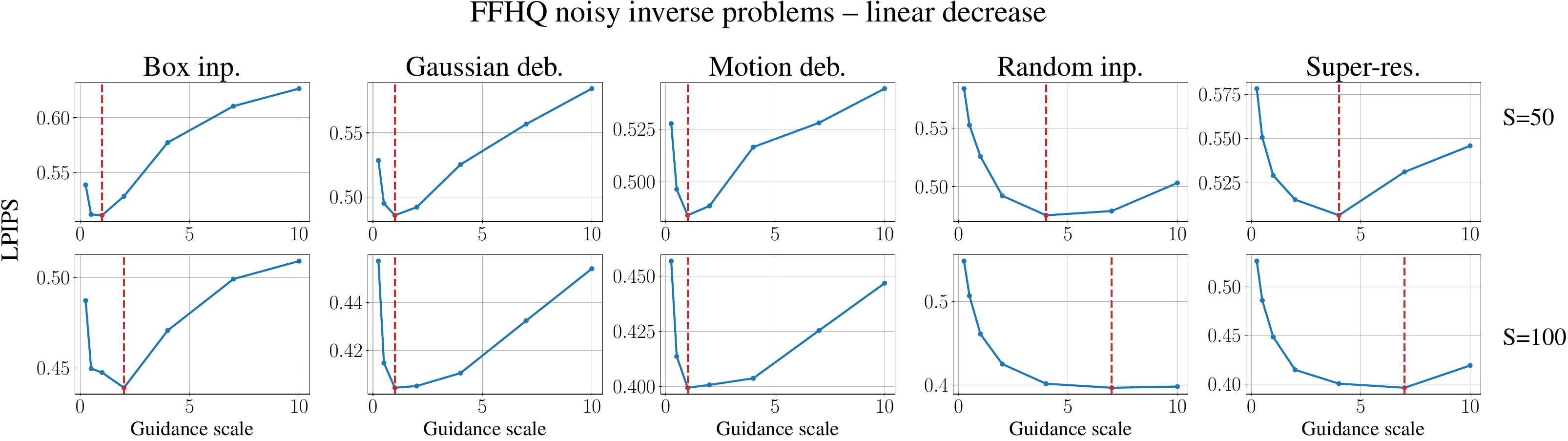}
    \caption{Linear decrease, FFHQ: LPIPS vs.\ guidance scale; best value in red.}
    \label{fig:grid_search_linear_decrease_ffhq}
\end{figure}

\begin{figure}[ht]
    \centering
    \includegraphics[width=0.9\textwidth]{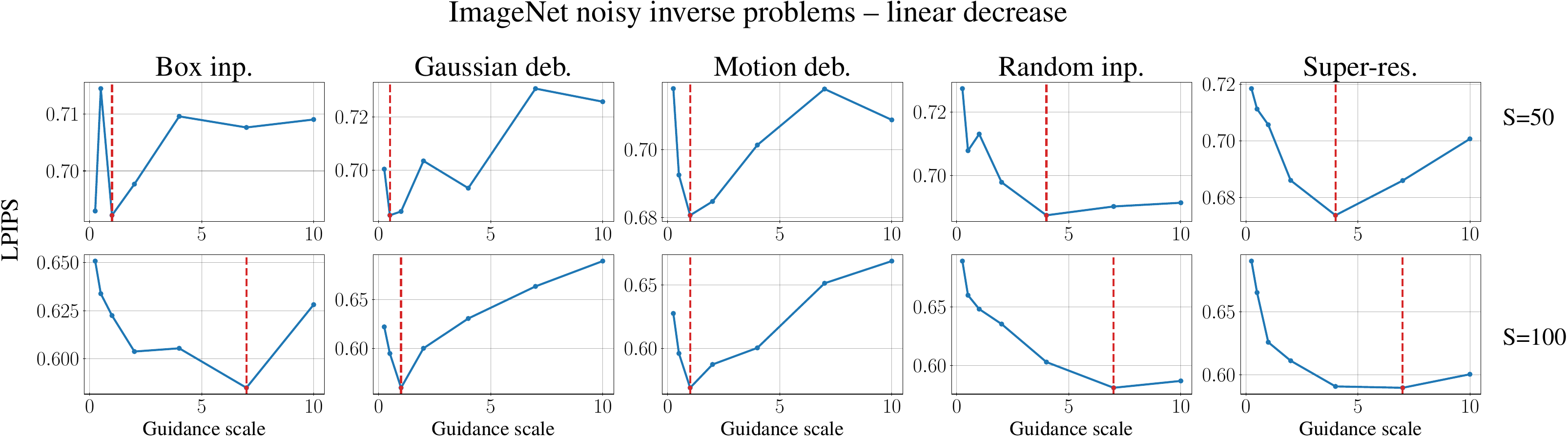}
    \caption{Linear decrease, ImageNet: LPIPS vs.\ guidance scale; best value in red.}
    \label{fig:grid_search_linear_decrease_imagenet}
\end{figure}

\begin{figure}[ht]
    \centering
    \includegraphics[width=0.9\textwidth]{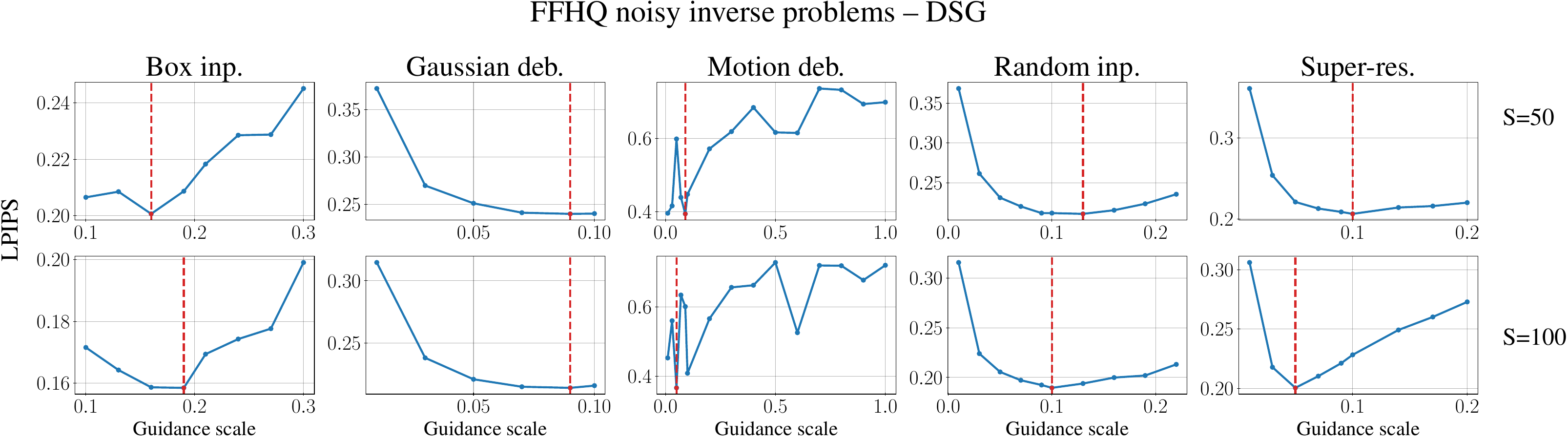}
    \caption{DSG, FFHQ: LPIPS vs.\ guidance scale; best value in red.}
    \label{fig:grid_search_dsg_ffhq}
\end{figure}

\begin{figure}[ht]
    \centering
    \includegraphics[width=0.9\textwidth]{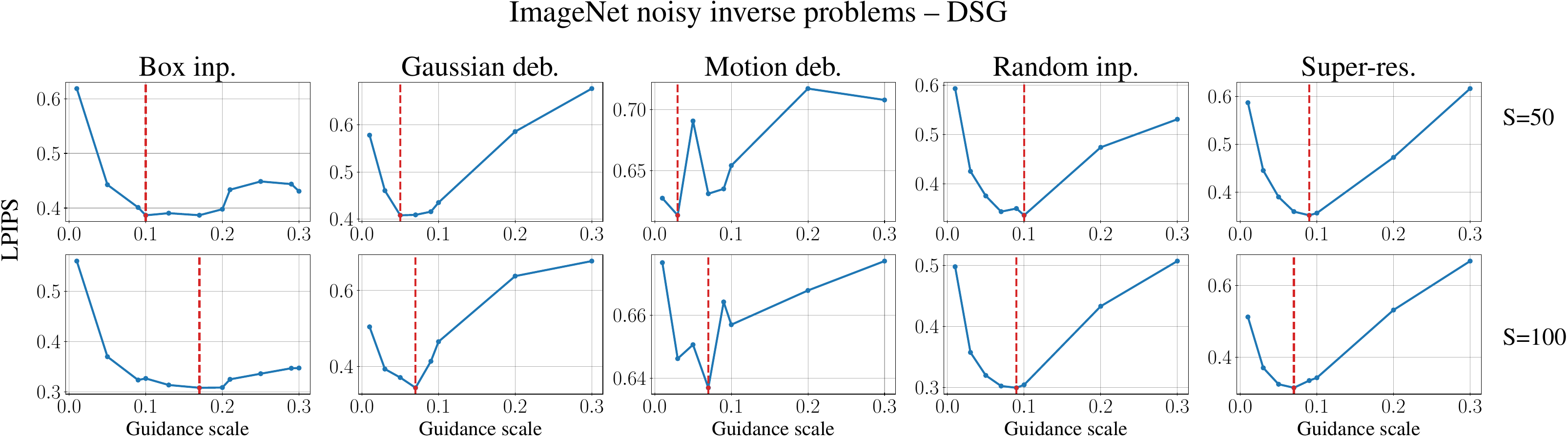}
    \caption{DSG, ImageNet: LPIPS vs.\ guidance scale; best value in red.}
    \label{fig:grid_search_dsg_imagenet}
\end{figure}

\begin{figure}[ht]
    \centering
    \includegraphics[width=0.9\textwidth]{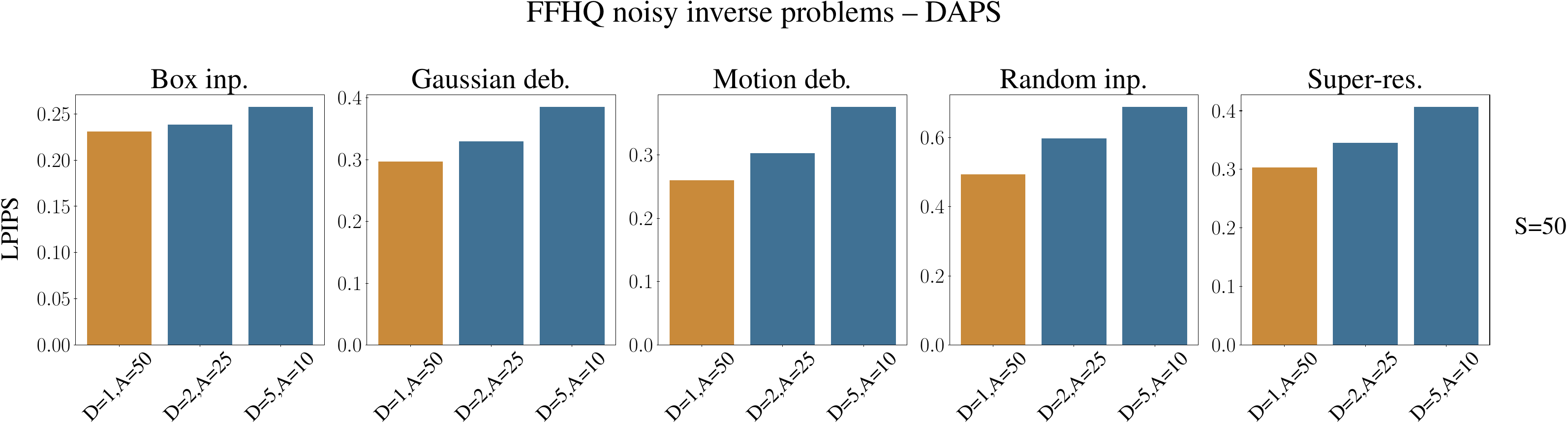}
    \caption{DAPS, FFHQ, $50$ NFEs: LPIPS per configuration $(D, A)$; best value in orange.}
    \label{fig:grid_search_daps_ffhq_s50}
\end{figure}

\begin{figure}[ht]
    \centering
    \includegraphics[width=0.9\textwidth]{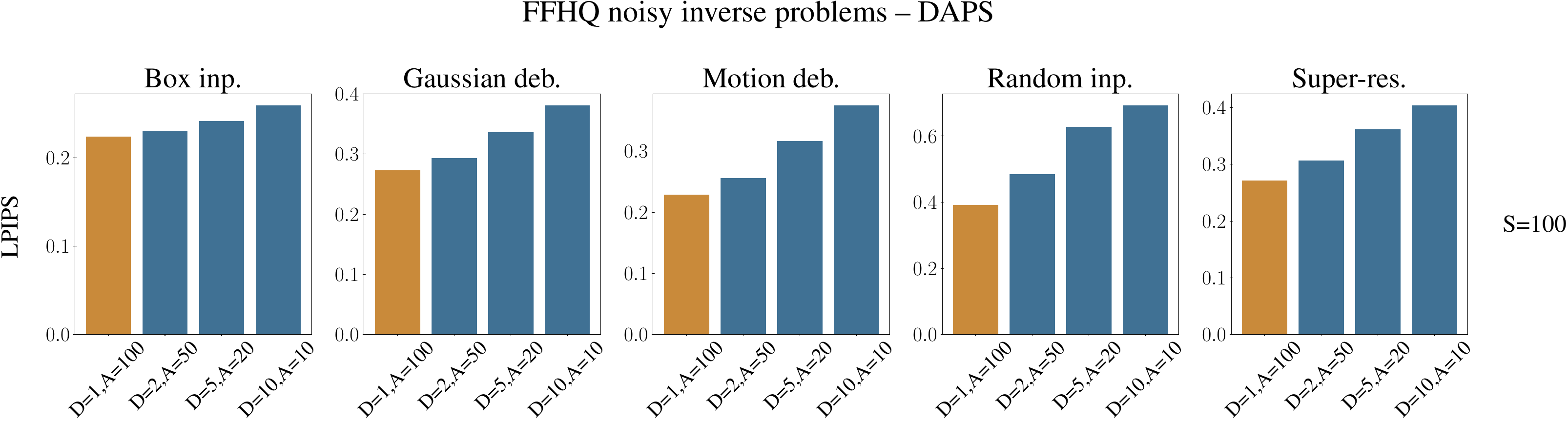}
    \caption{DAPS, FFHQ, $100$ NFEs: LPIPS per configuration $(D, A)$; best value in orange.}
    \label{fig:grid_search_daps_ffhq_s100}
\end{figure}

\begin{figure}[ht]
    \centering
    \includegraphics[width=0.9\textwidth]{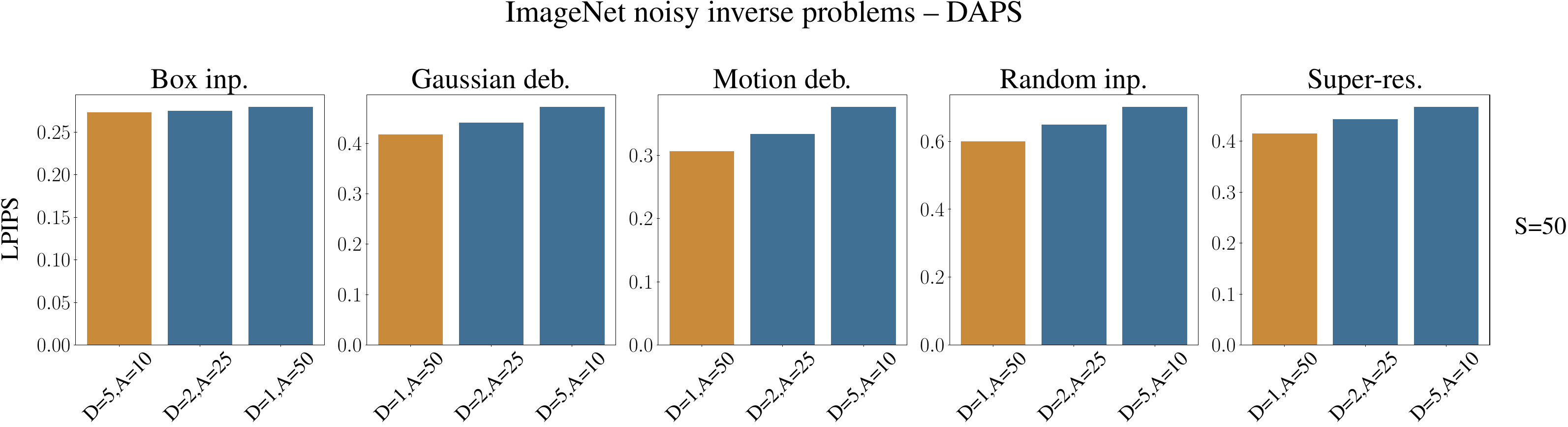}
    \caption{DAPS, ImageNet, $50$ NFEs: LPIPS per configuration $(D, A)$; best value in orange.}
    \label{fig:grid_search_daps_imagenet_s50}
\end{figure}

\begin{figure}[ht]
    \centering
    \includegraphics[width=0.9\textwidth]{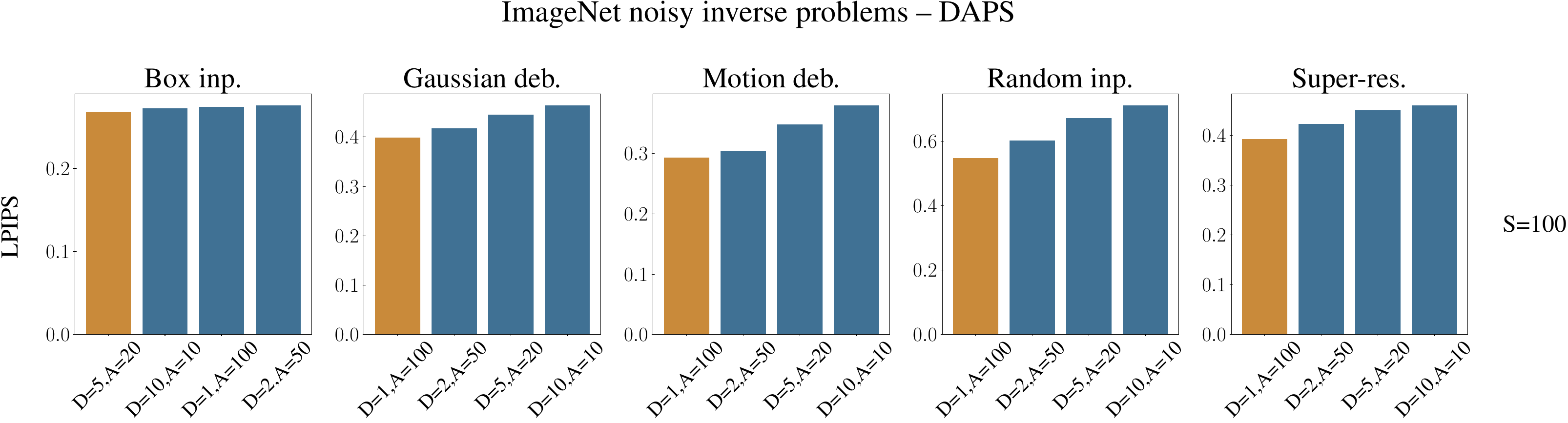}
    \caption{DAPS, ImageNet, $100$ NFEs: LPIPS per configuration $(D, A)$; best value in orange.}
    \label{fig:grid_search_daps_imagenet_s100}
\end{figure}

\paragraph{Baselines.} Unless stated otherwise, the hyperparameters of each baseline are selected by LPIPS on $20$ validation images, and the search results are given in Appendix~\ref{app:additional_hyperparameter_searches}. For the guidance-scale searches, the grid is expanded whenever the optimum lies at its boundary.
\begin{itemize}
    \item \textbf{Constant guidance} applies a fixed guidance scale $w$ at every step. We select $w$ by a grid search (Figures~\ref{fig:grid_search_dps_ffhq} and~\ref{fig:grid_search_dps_imagenet}).
    \item \textbf{Linear increase} ramps the guidance scale linearly from $0$ at the first step to $w$ at the final step, and \textbf{linear decrease} ramps it from $w$ down to $0$. We select $w$ by a grid search, as for constant guidance (Figures~\ref{fig:grid_search_linear_increase_ffhq}--\ref{fig:grid_search_linear_decrease_imagenet}).
    \item \textbf{DSG} \citep{yang2024dsg}: we use the authors' repository (\url{https://github.com/LingxiaoYang2023/DSG2024}), apply DSG guidance at every step (interval $1$), and select the guidance scale by a grid search, leaving all other hyperparameters at their defaults (Figures~\ref{fig:grid_search_dsg_ffhq} and~\ref{fig:grid_search_dsg_imagenet}).
    \item \textbf{CDIM} \citep{jayaram2025CDIM}: we use the authors' repository (\url{https://github.com/vivjay30/cdim}) with its default configurations at $T=25$ and $T=50$ for the budgets of $50$ and $100$ NFEs, respectively; these values of $T$ were chosen because they are the authors' own defaults, for which good hyperparameters (e.g.\ the step size $\lambda$) were already established. CDIM's number of NFEs is not fixed but varies per sample, and at these settings its average is consistently higher than $50$ and $100$, respectively, across every task and dataset (Table~\ref{tab:cdim_nfes}), except random inpainting at $T=50$.
    \item \textbf{DAPS} \citep{zhang2025improvingdiffusioninverseproblem} alternates denoising with renoising, and its number of NFEs is the product of the number of outer annealing steps $A$ and inner diffusion (ODE) steps $D$. Using the authors' repository (\url{https://github.com/zhangbingliang2019/DAPS}), we search over different splits of the budget: $(A, D) \in \{(10, 5), (25, 2), (50, 1)\}$ for $50$ NFEs, and $(A, D) \in \{(10, 10), (20, 5), (50, 2), (100, 1)\}$ for $100$ NFEs (Figures~\ref{fig:grid_search_daps_ffhq_s50}--\ref{fig:grid_search_daps_imagenet_s100}).
    \item \textbf{DPS} \citep{chung2023DPS}: the 1000-NFE results are taken from \citet{chung2023DPS}.
\end{itemize}

\begin{table}[ht]
    \small
\centering
\begin{tabular}{l cc}
\toprule
Subtask & Avg. NFEs ($T=25$) & Avg. NFEs ($T=50$) \\
\midrule
\multicolumn{3}{l}{\emph{FFHQ}} \\
Box inpainting & 55.7 & 102.8 \\
Random inpainting & 50.7 & 93.7 \\
Super-resolution ($4\times$) & 54.5 & 104.7 \\
Gaussian deblur & 56.0 & 104.4 \\
Motion deblur & 65.8 & 114.2 \\
\midrule
\multicolumn{3}{l}{\emph{ImageNet}} \\
Box inpainting & 73.4 & 123.0 \\
Random inpainting & 53.7 & 95.0 \\
Super-resolution ($4\times$) & 62.1 & 112.6 \\
Gaussian deblur & 66.1 & 113.7 \\
Motion deblur & 80.6 & 128.6 \\
\bottomrule
\end{tabular}
\vspace{2mm}
\caption{Average number of NFEs used by CDIM per subtask at its two settings, $T=25$ and $T=50$ (nominally $50$ and $100$ NFEs), computed over the full test set of Tables \ref{tab:results_ffhq_noisy} and \ref{tab:results_imagenet_noisy}.}
\label{tab:cdim_nfes}
\end{table}

\paragraph{Training configuration.} Table~\ref{tab:cfg_noisy} gives the learning rate, number of epochs and weight decay used to train LEEGS for each subtask.

\begin{table}[ht]
    \small
\centering
\begin{tabular}{l ccc}
\toprule
Subtask & Learning rate $\eta$ & Epochs & Weight decay \\
\midrule
\multicolumn{4}{l}{\emph{FFHQ}} \\
Super-resolution ($4\times$) & 0.3 & 25 ($S=50$) / 15 ($S=100$) & 1e-4 ($S=50$) / 5e-4 ($S=100$) \\
Box inpainting & 0.2 & 25 ($S=50$) / 25 ($S=100$) & 0.0 \\
Random inpainting & 0.8 & 25 ($S=50$) / 20 ($S=100$) & 0.0 \\
Gaussian deblur & 0.5 & 25 ($S=50$) / 25 ($S=100$) & 0.0 \\
Motion deblur & 0.3 & 30 ($S=50$) / 25 ($S=100$) & 0.0 \\
\midrule
\multicolumn{4}{l}{\emph{ImageNet}} \\
Super-resolution ($4\times$) & 0.3 & 20 ($S=50$) / 15 ($S=100$) & 0.0 ($S=50$) / 5e-4 ($S=100$) \\
Box inpainting & 0.3 & 20 ($S=50$) / 15 ($S=100$) & 0.0 \\
Random inpainting & 0.8 & 20 ($S=50$) / 15 ($S=100$) & 0.0 \\
Gaussian deblur & 0.2 & 20 ($S=50$) / 15 ($S=100$) & 0.0 \\
Motion deblur & 0.3 & 20 ($S=50$) / 15 ($S=100$) & 0.0 \\
\bottomrule
\end{tabular}
\vspace{2mm}
\caption{LEEGS training configuration for the noisy inverse problems (Tables \ref{tab:results_ffhq_noisy}, \ref{tab:results_imagenet_noisy}).}
\label{tab:cfg_noisy}
\end{table}

\subsection{Image inpainting}
\label{app:exp_inpainting}

\begin{figure}[ht]
    \centering
    \includegraphics[width=\textwidth]{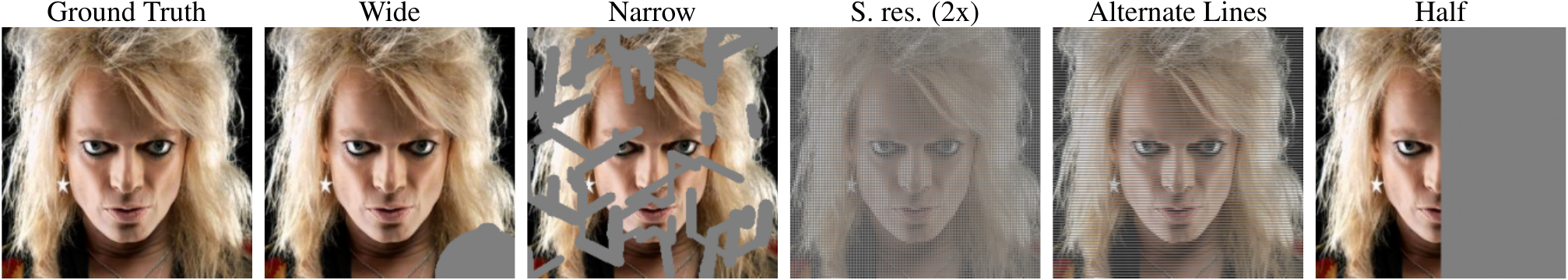}
    \caption{Ground truth and the 5 noiseless inpainting mask operators used for Table \ref{tab:results_inpainting_celeba} (CelebA-HQ example).}
    \label{fig:task_illustration_inpainting}
\end{figure}

\paragraph{Models.} Following \citet{lugmayr2022repaintinpaintingusingdenoising}, we use the pretrained CelebA-HQ model of \citet{lugmayr2022repaintinpaintingusingdenoising} and the pretrained ImageNet model of \citet{dhariwal2021diffusionmodelsbeatgans}.

\paragraph{Dataset generation.} For CelebA-HQ, we use the validation split (\texttt{val\_source\_256}) produced by the preprocessing pipeline of \citet{suvorov2022lama}, which is also used by \citet{lugmayr2022repaintinpaintingusingdenoising}. The test set of Table~\ref{tab:results_inpainting_celeba} consists of $100$ images from this split, and a disjoint pool of $100$ images from the same split is divided into $90$ images for training, from which each LEEGS run draws $30$, and $10$ for validation. For ImageNet, the test set consists of $100$ images subsampled from the test set of the noisy inverse problems (\sect\ref{app:exp_noisy_inverse}), and another disjoint pool of $100$ images is split in the same way into $90$ images for training and $10$ for validation. We use the same set of masks provided by \citet{lugmayr2022repaintinpaintingusingdenoising} for both training and testing (Figure~\ref{fig:task_illustration_inpainting}). All mask types except for the wide and narrow masks are fixed across images.

\paragraph{Baselines.} We run RePaint using the authors' repository (\url{https://github.com/andreas128/RePaint}). RePaint has three hyperparameters: the number of sampling steps $S$, the jump length for renoising $J$, and the number of times $N$ the jump is taken, resulting in approximately $S \cdot N$ NFEs. We try five configurations that yield approximately $200$ NFEs ($201$ in all cases): ($S=200, N=1, J=1$), ($S=101, N=2, J=5$), ($S=101, N=2, J=10$), ($S=51, N=4, J=10$) and ($S=51, N=4, J=5$), and select the best per task by LPIPS on $20$ validation images. The results of the hyperparameter search are shown in Figures~\ref{fig:grid_search_celeba_inpaint} and~\ref{fig:grid_search_imagenet_inpaint} in Appendix~\ref{app:additional_hyperparameter_searches}.

\paragraph{Training configuration.} Table~\ref{tab:cfg_inpaint} gives the learning rate, number of epochs and weight decay used to train LEEGS for each mask.

\begin{table}[ht]
    \small
\centering
\begin{tabular}{l ccc}
\toprule
Mask & Learning rate $\eta$ & Epochs & Weight decay \\
\midrule
\multicolumn{4}{l}{\emph{ImageNet}} \\
Wide & 0.2 & 5 & 0.0 \\
Narrow & 0.2 & 5 & 0.0 \\
Super-resolve ($2\times$) & 0.2 & 5 & 0.0 \\
Alternating lines & 0.2 & 5 & 0.0 \\
Half & 0.2 & 5 & 0.0 \\
\midrule
\multicolumn{4}{l}{\emph{CelebA-HQ}} \\
Wide & 0.2 & 5 & 0.0 \\
Narrow & 0.2 & 5 & 0.0 \\
Super-resolve ($2\times$) & 0.2 & 5 & 0.0 \\
Alternating lines & 0.2 & 5 & 0.0 \\
Half & 0.2 & 5 & 0.0 \\
\bottomrule
\end{tabular}
\vspace{2mm}
\caption{LEEGS training configuration for image inpainting (Table \ref{tab:results_inpainting_celeba}).}
\label{tab:cfg_inpaint}
\end{table}

\subsection{Face-ID guidance}
\label{app:exp_faceid}

\paragraph{Models.} We use the unconditional CelebA-HQ $256\times256$ model of \citet{lugmayr2022repaintinpaintingusingdenoising}, as for image inpainting. Identity is measured with an ArcFace (IR-SE-50) face-recognition network \citep{deng2019arcface}.

\paragraph{Dataset generation.} As for image inpainting (\sect\ref{app:exp_inpainting}), the images are drawn from the CelebA-HQ \texttt{val\_source\_256} split, with a test set of $100$ images and a disjoint training pool from which each LEEGS run draws $30$ images. The identity target for each image is the image itself. FID is computed against the $26{,}000$ images of the CelebA-HQ \texttt{train\_256} split.

\paragraph{Baselines.} For constant guidance, we select the guidance scale for each NFE budget by a grid search on $20$ validation images, minimizing the mean FaceID score. For DSG \citep{yang2024dsg}, we tune the guidance scale in the same way, and use our own implementation for sampling. We set its interval to $1$, so that DSG guidance is applied at every step.

\paragraph{Training configuration.} Table~\ref{tab:cfg_faceid} gives the learning rate, number of epochs and weight decay used to train LEEGS for each NFE budget.

\begin{table}[ht]
    \small
\centering
\begin{tabular}{l ccc}
\toprule
NFEs & Learning rate $\eta$ & Epochs & Weight decay \\
\midrule
50 & 0.1 & 15 & 0.0 \\
100 & 0.1 & 10 & 0.0 \\
\bottomrule
\end{tabular}
\vspace{2mm}
\caption{LEEGS training configuration for Face-ID guidance (Table \ref{tab:results_faceid_celeba}).}
\label{tab:cfg_faceid}
\end{table}

\subsection{PDE problems}
\label{app:exp_pde}

\paragraph{Background.} We consider static boundary-value problems (BVPs)
\begin{align*}
    \FF(z; a, u) &= 0, \quad z \in \Omega \subseteq \R^d, \mlabel{eqn:bvp_defn} \\
    u(z) &= g(z), \quad z \in \partial \Omega,
\end{align*}
where $\FF$ is a (possibly non-linear) differential operator, $\Omega$ is a bounded spatial domain with boundary $\partial \Omega$, $g$ specifies the boundary condition, $a(\cdot)$ is a spatially varying coefficient, and $u(\cdot)$ is the solution. The \emph{forward problem} recovers $u$ from $a$, and the \emph{inverse problem} recovers $a$ from $u$. Following DiffusionPDE \citep{huang2024dpde}, a pretrained diffusion model represents the joint distribution over pairs $(a, u)$ rendered as $128 \times 128$ images, and guidance enforces consistency with sparse observations: for the forward problem, $a$ is observed at $500$ randomly chosen locations $z_{\tx{obs}}$ ($\approx 3\%$ of pixels), and analogously $u$ for the inverse problem. Following \citet{huang2024dpde}, we consider the three problems
\begin{align*}
    \text{Poisson:} \quad & -\nabla^2 u(z) = a(z), \\
    \text{Helmholtz:} \quad & \nabla^2 u(z) + k^2 u(z) = a(z), \\
    \text{Darcy flow:} \quad & -\nabla \cdot (a(z)\nabla u(z)) = 1,
\end{align*}
each on a bounded domain $\Omega$ with $u(z) = 0$ on $\partial \Omega$.

\paragraph{Models.} We use the pretrained models of \citet{huang2024dpde} for all three problems.

\paragraph{Dataset generation.} We use the test set provided by \citet{huang2024dpde} ($1024$ samples per problem). We set aside $100$ samples, split into $90$ for training, from which each LEEGS run draws $30$, and $10$ for validation. We form our test set from the remaining samples ($1000$ for Helmholtz and Darcy, and $924$ for Poisson). The $500$ observed locations are sampled randomly for each example.

\begin{figure}[ht]
    \centering
    \includegraphics[width=0.9\textwidth]{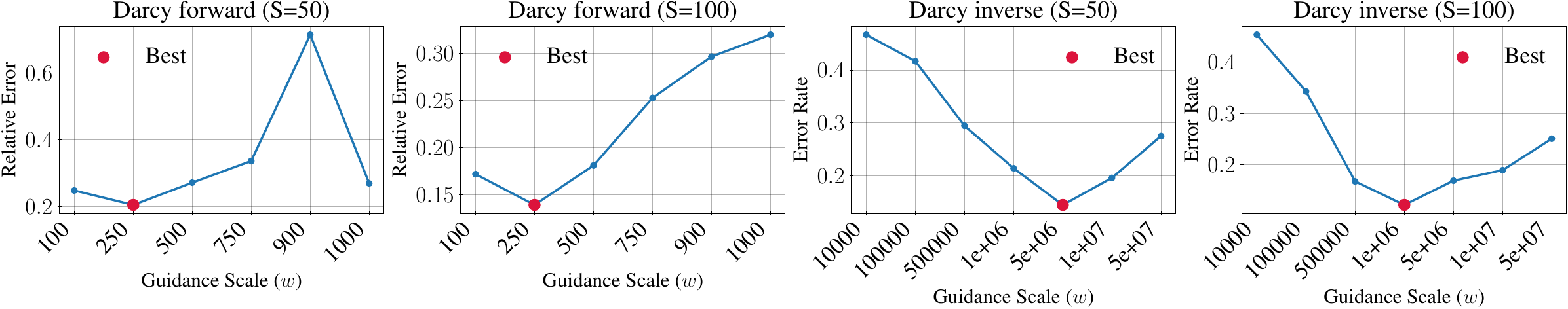}
    \caption{Constant guidance, Darcy: error vs.\ guidance scale; columns: forward/inverse $\times$ \{$50$, $100$ NFEs\}; best value in red.}
    \label{fig:grid_search_pde_darcy}
\end{figure}

\begin{figure}[ht]
    \centering
    \includegraphics[width=0.9\textwidth]{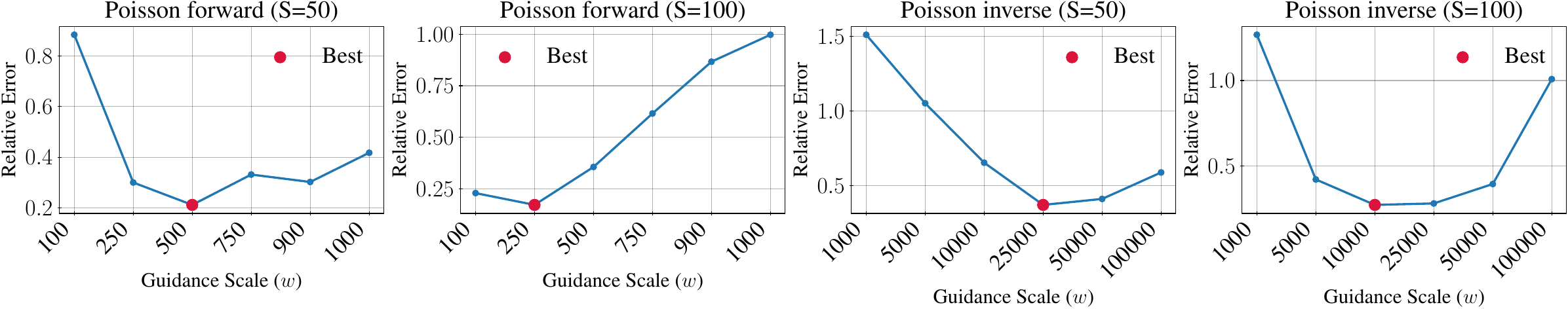}
    \caption{Constant guidance, Poisson: as in Figure~\ref{fig:grid_search_pde_darcy}.}
    \label{fig:grid_search_pde_poisson}
\end{figure}

\begin{figure}[ht]
    \centering
    \includegraphics[width=0.9\textwidth]{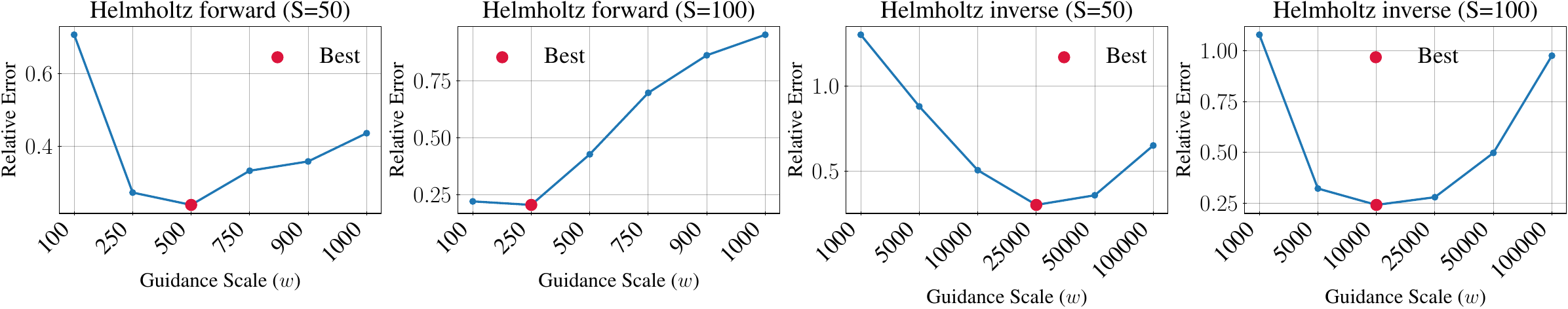}
    \caption{Constant guidance, Helmholtz: as in Figure~\ref{fig:grid_search_pde_darcy}.}
    \label{fig:grid_search_pde_helmholtz}
\end{figure}

\paragraph{Baselines.} As for the noisy inverse problems (\sect\ref{app:exp_noisy_inverse}), the constant-guidance and linear-schedule baselines are each defined by a single guidance scale, which we select by a grid search on $20$ validation samples, using the relative error or the error rate as appropriate. The resulting large guidance scales are consistent with those used by \citet{huang2024dpde}. The search curves are given in Figures~\ref{fig:grid_search_pde_darcy}--\ref{fig:grid_search_pde_helmholtz} and~\ref{fig:grid_search_dpde_linear_increase_darcy}--\ref{fig:grid_search_dpde_linear_decrease_helmholtz} in Appendix~\ref{app:additional_hyperparameter_searches}. Following \citet{huang2024dpde}, the constant-guidance baseline uses a second-order sampler, and the DiffusionPDE results are those reported by \citet{huang2024dpde}.

\paragraph{Training configuration.} Table~\ref{tab:cfg_pde} gives the learning rate, number of epochs and weight decay used to train LEEGS for each problem.

\begin{table}[ht]
    \small
\centering
\begin{tabular}{l ccc}
\toprule
Problem & Learning rate $\eta$ & Epochs & Weight decay \\
\midrule
Poisson, forward & 15.0 & 15 ($S=50$) / 15 ($S=100$) & 0.0 \\
Poisson, inverse & 500.0 & 15 ($S=50$) / 15 ($S=100$) & 0.0 \\
Helmholtz, forward & 15.0 ($S=50$) / 10.0 ($S=100$) & 15 ($S=50$) / 15 ($S=100$) & 0.0 \\
Helmholtz, inverse & 1500.0 & 15 ($S=50$) / 10 ($S=100$) & 0.0 \\
Darcy, forward & 30.0 & 15 ($S=50$) / 15 ($S=100$) & 0.0 \\
Darcy, inverse & 150000.0 ($S=50$) / 100000.0 ($S=100$) & 15 ($S=50$) / 15 ($S=100$) & 0.0 \\
\bottomrule
\end{tabular}
\vspace{2mm}
\caption{LEEGS training configuration for the PDE problems (Table \ref{tab:results_dpde}).}
\label{tab:cfg_pde}
\end{table}

\subsection{Computational resources}

Our experiments were run on an internal cluster with NVIDIA A40 and A100 GPUs and an internal workstation with an NVIDIA RTX 3090 GPU. Table~\ref{tab:timing_baselines} gives per-task training wall-clock times, measured on a single A100 with no validation logging (validation is used only for monitoring, not for early stopping or model selection, so it is not part of the reported training cost). Since different tasks within the same results table can use different epoch counts at a fixed NFE budget $S$ (Tables~\ref{tab:cfg_noisy}, \ref{tab:cfg_inpaint}, \ref{tab:cfg_pde}), we report the minimum and maximum times spent at each $S$ whenever they differ.

\begin{table}[ht]
    \small
\centering
\begin{tabular}{l l cc c}
\toprule
Results table & $S$ & Epochs & Wall-clock & Time/epoch \\
\midrule
\multirow{2}{*}{FFHQ (Table~\ref{tab:results_ffhq_noisy})} & 50 & 25 -- 30 & 1h26m -- 1h57m & 3.4 -- 3.9 min \\
 & 100 & 15 -- 25 & 1h52m -- 3h06m & 7.5 min \\
\multirow{2}{*}{ImageNet (Table~\ref{tab:results_imagenet_noisy})} & 50 & 20 & 3h11m & 9.6 min \\
 & 100 & 15 & 4h44m & 19.0 min \\
\multirow{2}{*}{Inpainting (Table~\ref{tab:results_inpainting_celeba})} & 200 (CelebA-HQ) & 5 & 3h13m & 38.7 min \\
 & 200 (ImageNet) & 5 & 3h14m & 38.7 min \\
\multirow{2}{*}{PDE problems (Table~\ref{tab:results_dpde})} & 50x1 & 15 & 1h08m & 4.5 min \\
 & 100x1 & 10 -- 15 & 1h37m -- 2h21m & 9.4 -- 9.7 min \\
\multirow{2}{*}{Face-ID (Table~\ref{tab:results_faceid_celeba})} & 50 & 15 & 2h34m & 10.3 min \\
 & 100 & 10 & 3h27m & 20.8 min \\
\bottomrule
\end{tabular}
\vspace{2mm}
\caption{Training wall-clock time by results table and NFE budget $S$, on a single A100 GPU with no validation logging. Where tasks at a fixed $S$ use different epoch counts, both the minimum and maximum are reported (as a range); a single value indicates every task at that $S$ trains for the same number of epochs.}
\label{tab:timing_baselines}
\end{table}

\newpage
\clearpage
\section{Further experiments and ablations}
\subsection{Ablation on training dataset size}
\label{app:variance}

This section analyzes the variance in the learned guidance schedule due to the choice of training set. We consider two tasks : the noisy motion deblur on FFHQ, and the Helmholtz forward PDE problem, both at $S=10$. We train the guidance schedule on datasets of varying size $n$, where each dataset is drawn randomly from a larger pool of $500$ samples. For each $n$ we draw $10$ datasets and train a separate guidance schedule on each, holding the total number of training iterations fixed as $n$ varies. For both tasks, we sweep $n \in \{1, 5, 30, 50, 100\}$.

\begin{figure}[ht]
    \centering
    \includegraphics[width=0.48\textwidth]{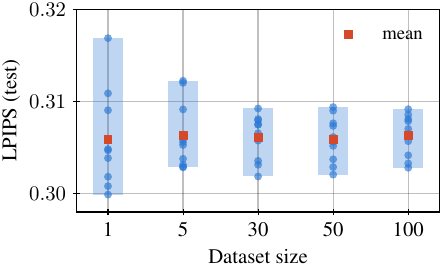}
    \hfill
    \includegraphics[width=0.48\textwidth]{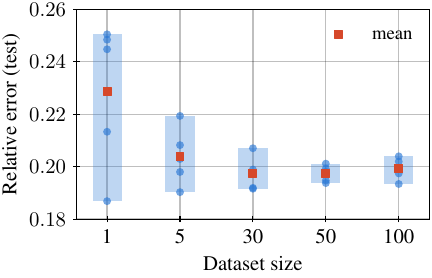}
    \caption{Held-out test-set reconstruction quality as a function of training dataset size
    $n$. Scatter points are the $10$ individual dataset draws, and the vertical bar spans min to max, with the mean indicated by a red square. \emph{Left:} FFHQ motion deblur. \emph{Right:} Helmholtz forward.}
    \label{fig:variance_dataset_size}
\end{figure}

\begin{figure}[ht]
    \centering
    \includegraphics[width=0.48\textwidth]{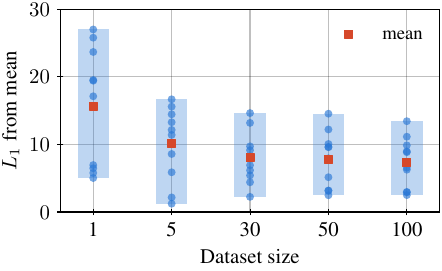}
    \hfill
    \includegraphics[width=0.48\textwidth]{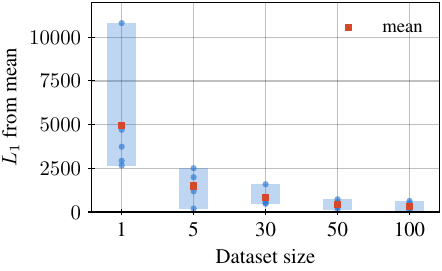}
    \caption{Per-draw $L_1$ distance of the learned guidance schedule from the draw-averaged
    schedule (all $S=10$ steps), as a function of training dataset size $n$. \emph{Left:} FFHQ
    motion deblur. \emph{Right:} Helmholtz forward.}
    \label{fig:variance_dataset_size_l1}
\end{figure}

\begin{figure}[ht]
    \centering
    \includegraphics[width=0.48\textwidth]{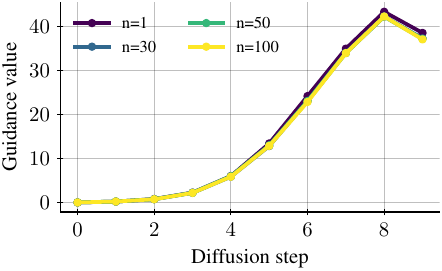}
    \hfill
    \includegraphics[width=0.48\textwidth]{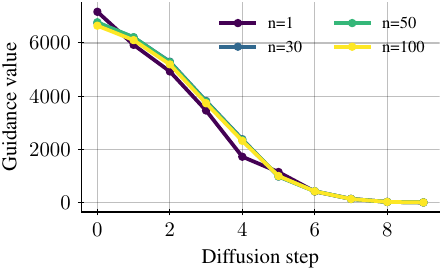}
    \caption{Draw-averaged guidance schedule (mean over $10$ dataset draws) as a function of diffusion step, compared across training dataset sizes $n$ \emph{Left:} FFHQ motion deblur. \emph{Right:} Helmholtz forward.}
    \label{fig:variance_dataset_size_mean_schedules}
\end{figure}

We evaluate each learned schedule on a held-out test set of $300$ samples, reporting the reconstruction quality [LPIPS for FFHQ, relative $L_2$ error with respect to $u$ (the solution) for Helmholtz forward; see Figure \ref{fig:variance_dataset_size}] and the per-draw distance of the learned schedule from the draw-averaged schedule (see Figure \ref{fig:variance_dataset_size_l1}). The latter metric directly measures the variability of the schedule, while the former measures its effect on overall performance. We also plot the mean of the learned schedules across the $10$ draws for different $n$ in Figure \ref{fig:variance_dataset_size_mean_schedules}. 

We see that the variance with respect to both metrics decreases with increasing $n$. While the evaluation metrics plateau at larger $n$ for both tasks, the variability of the learned schedules continues to decrease for the Helmholtz forward problem. This could be due to the use of a deterministic sampler for the PDE problem, which reduces the variance of the learned schedule. In contrast, the FFHQ motion deblur task uses a stochastic sampler, which introduces additional variability in the learned schedule. 

Figure \ref{fig:variance_dataset_size_mean_schedules} shows that the \textit{mean schedule} is relatively stable across $n$, especially for the FFHQ motion deblur task.  Note that this is not a measure of variance, but the trend in the average schedule learned across the $10$ draws for each $n$. This observation seems to imply that training with fewer samples has more variance and less bias, with respect to the mean schedule learned with larger datasets.

\subsection{Comparison of LEEGS vs true gradient}
\label{app:method_comparison}

\begin{figure}[ht]
    \centering
    \includegraphics[width=\textwidth]{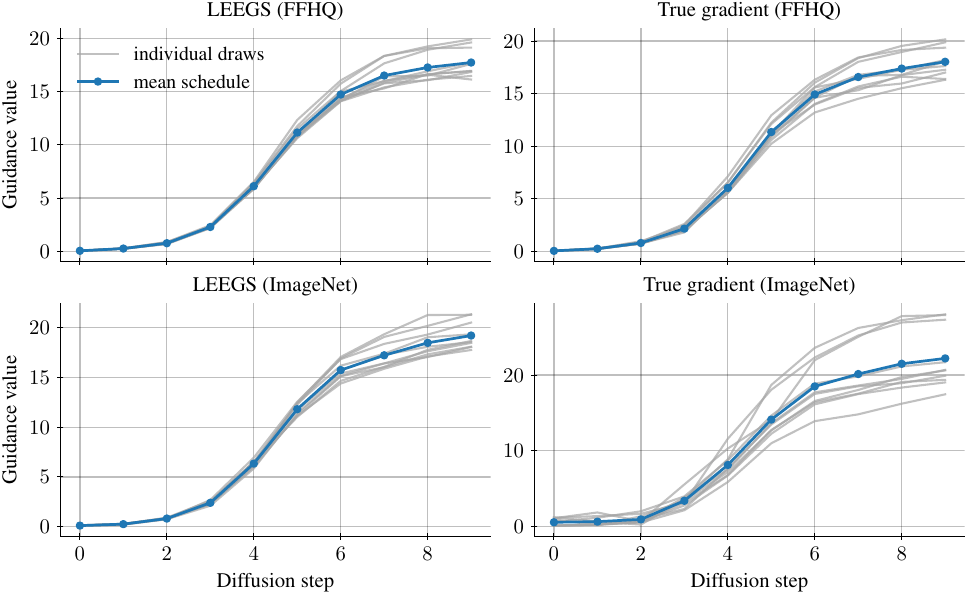}
    \caption{Learned guidance schedules for LEEGS and the true gradient (FFHQ and ImageNet
    motion deblur, $S=10$). Each panel shows the $10$ per-draw schedules (thin grey) and their
    mean (bold); the true gradient's higher spread on ImageNet is visible in the wider band
    around the mean, especially at later diffusion steps.}
    \label{fig:method_comparison_schedules_mean}
\end{figure}

\paragraph{Schedule comparison.} We compare the LEEGS gradient (guidance term detached before backpropagation) against the true gradient (nothing detached) on the noisy motion deblur task (observation noise $\sigma=0.05$), on FFHQ and ImageNet datasets at $S=10$ with the DDPM sampler. For each dataset (FFHQ/ImageNet) and method (true gradient/LEEGS) we train 10 runs while varying the dataset of training samples, where each run uses $50$ random images resampled from a $500$-image pool. Each run is trained for $8$ epochs with a learning rate of $0.08$. 

We evaluate each draw on $300$ held-out test images. We report LPIPS against the clean source image, and $L_1$/$L_2$ distance of the draw's schedule from the $10$-draw mean schedule in Table \ref{tab:method_comparison_lpips_spread} and plot the learned schedules in Figure \ref{fig:method_comparison_schedules_mean}. On FFHQ, the two methods are comparable, while on ImageNet the true gradient is both worse and far less stable than LEEGS (std $4$--$6\times$ higher on LPIPS, distances $\sim\!3\times$ higher). The true gradient's instability is visible in the wider band around the mean schedule in Figure \ref{fig:method_comparison_schedules_mean}, especially at later diffusion steps.

\begin{table}[ht]
    \small
\centering
\begin{tabular}{l l c c c}
\hline
Dataset & Mode & LPIPS ($\downarrow$) & $L_2$ dist.\ from mean & $L_1$ dist.\ from mean \\ \hline
\multirow{2}{*}{FFHQ} & LEEGS         & $0.341 \pm 0.004$ & $2.05$ & $4.50$ \\
                      & True gradient & $0.342 \pm 0.006$ & $2.34$ & $5.52$ \\ \hline
\multirow{2}{*}{ImageNet} & LEEGS         & $0.560 \pm 0.007$ & $2.28$ & $5.13$ \\
                          & True gradient & $0.594 \pm 0.041$ & $7.03$ & $17.89$ \\ \hline
\end{tabular}
\vspace{2mm}
\caption{LEEGS vs.\ the true gradient on motion deblur ($S=10$; mean $\pm$ std over $10$ draws).
$L_1$/$L_2$ dist.\ from mean is each draw's schedule distance to the $10$-draw mean. On ImageNet
the true gradient is both worse and far less stable than LEEGS (std $4$--$6\times$ higher on
LPIPS, distances $\sim\!3\times$ higher); on FFHQ the two are comparable.}
\label{tab:method_comparison_lpips_spread}
\end{table}

\paragraph{Timing comparison.} We measure the wall-clock cost of one training iteration (batch size $1$) as a function of the inference budget $S \in \{5, 10, 20, 40, 80, 100\}$. We compare LEEGS, the true gradient, and include plain sampling (no guidance-schedule optimization) as a reference. We report the mean over $10$ iterations, on FFHQ box inpainting (image domain) and Helmholtz forward (PDE domain), measured on a single NVIDIA RTX 5090 GPU. We see that the LEEGS approximation is $\sim\!4-5\times$ faster than the true gradient, while the previous section shows that the results are comparable on FFHQ, and the true gradient is less stable on Imagenet.

\begin{figure}[ht]
    \centering
    \includegraphics[width=0.48\textwidth]{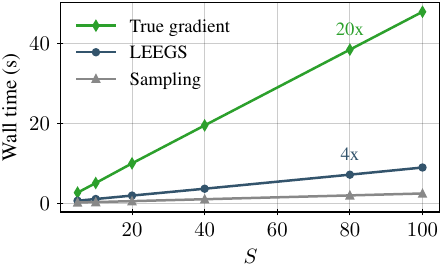}
    \hfill
    \includegraphics[width=0.48\textwidth]{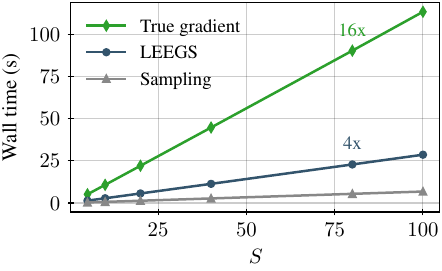}
    \caption{Wall-clock cost per training iteration vs.\ inference budget $S$, for LEEGS and the
    true gradient, relative to plain sampling. \emph{Left:} FFHQ box inpainting --
    true gradient $\approx 20\times$ sampling, LEEGS $\approx 4\times$. \emph{Right:} Helmholtz
    forward -- true gradient $\approx 16\times$, LEEGS $\approx 4\times$.}
    \label{fig:method_comparison_timing}
\end{figure}

\subsection{Effect of regularization}
\label{app:regularization}

This section studies the effect of weight decay on the end-to-end LEEGS guidance schedule, and its interaction with the number of diffusion steps $S$. We consider two FFHQ inverse tasks that behave differently under prolonged training: \textit{superresolution}, where we observe an overfitting elbow in which the terminal constraint keeps improving while held-out reconstruction quality worsens, and \textit{box inpainting}, where no such elbow appears. For
each task we train the guidance schedule with the Adam optimizer's weight decay
($\text{wdecay}$) applied to the guidance-schedule parameters, sweeping $\text{wdecay} \in \{0, 10^{-5}, 10^{-4}, 5\times10^{-4}\, 10^{-3}\}$ across $S \in \{10, 25, 50, 100\}$ diffusion steps, for a total of $20$ configurations per task. We train using a fixed set of $30$ samples for all runs. The learning rate and number of training epochs are fixed per task and $S$. For superresolution, we use a learning rate of $0.3$ and train for $30$ epochs at $S \in \{10, 25\}$, $25$ epochs at $S=50$, and $15$ epochs at $S=100$. For the box inpainting task we train for 20 epochs with a learning rate of $0.3$ for $S=10$, $25$ epochs with a learning rate of $0.3$ for $S=25$, and $25$ epochs with a learning rate of $0.2$ for $S \in \{50, 100\}$.

\begin{figure}[ht]
    \centering
    \includegraphics[width=0.48\textwidth]{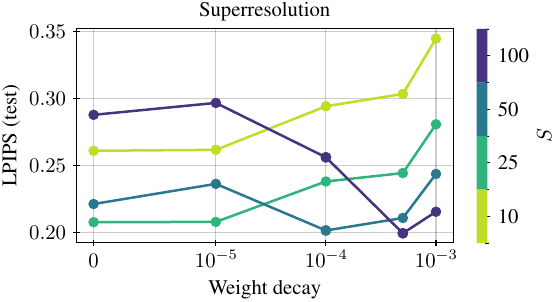}
    \hfill
    \includegraphics[width=0.48\textwidth]{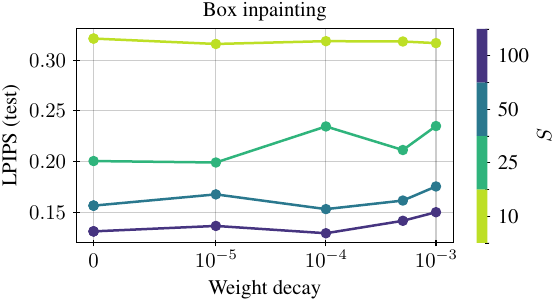}
    \caption{Held-out LPIPS as a function of weight decay $\gamma$, one curve per $S$
    (discrete colorbar). \emph{Left:} superresolution. \emph{Right:} box inpainting.}
    \label{fig:wdecay_eval_vs_wd}
\end{figure}

We evaluate each learned schedule on the held-out test set of $300$ samples, reporting LPIPS
between the reconstruction and the ground-truth image. Figure~\ref{fig:wdecay_train_lpips} in
the main text shows the corresponding \emph{training} LPIPS at $S=100$: without weight decay,
training exhibits a sharp reward-hacking elbow, past which the terminal constraint keeps
improving while the reconstruction degrades. Increasing weight decay progressively flattens
this elbow, and at $\gamma=5\times10^{-4}$ it disappears entirely. The resulting training LPIPS, however, never reaches the value attained just before
the unregularized elbow, so early stopping at that point is another fix for this
failure mode.

Figure~\ref{fig:wdecay_eval_vs_wd} (left) shows the analogous picture for superresolution with respect to the LPIPS on the held-out 300 sample test set. At $S=100$, held-out LPIPS improves
with weight decay up to $\gamma=5\times10^{-4}$, then turns back up at $\gamma=10^{-3}$, placing the optimum at $5\times10^{-4}$ rather than at the largest weight decay tried. The same effect appears at $S=50$, but weaker, with the optimum instead at $\gamma=10^{-4}$. At $S=10,25$, held-out LPIPS increases monotonically with weight decay: with fewer diffusion steps, the
schedule cannot overoptimize the constraint function to begin with, so regularization only costs reconstruction quality.

Box inpainting (Figure~\ref{fig:wdecay_eval_vs_wd}, right) is comparatively flat: held-out LPIPS is largely insensitive to weight decay across all $S$, except at $\gamma=10^{-3}$, where it clearly hurts, except at $S=10$ where it remains flat.

\subsection{Generalization across tasks}
\label{app:cross_task}

This section asks whether a guidance schedule learned for one task generalizes to other tasks, or equivalently, whether task-specific training is beneficial. We study this on the five FFHQ noisy inverse tasks in three parts: we compare the learned schedules, evaluate each schedule on every other task, and repeat this evaluation after matching the schedule magnitudes to isolate the effect of shape. All schedules are the LEEGS runs (first draw) used for Table~\ref{tab:results_ffhq_noisy} at $S \in \{50, 100\}$; training details are given in Appendix~\ref{app:exp_details} (Table~\ref{tab:cfg_noisy}). Learned schedules for ImageNet and the PDE tasks are given in Appendix~\ref{app:additional_schedules}.

\begin{figure}[ht]
    \centering
    \includegraphics[width=0.9\textwidth]{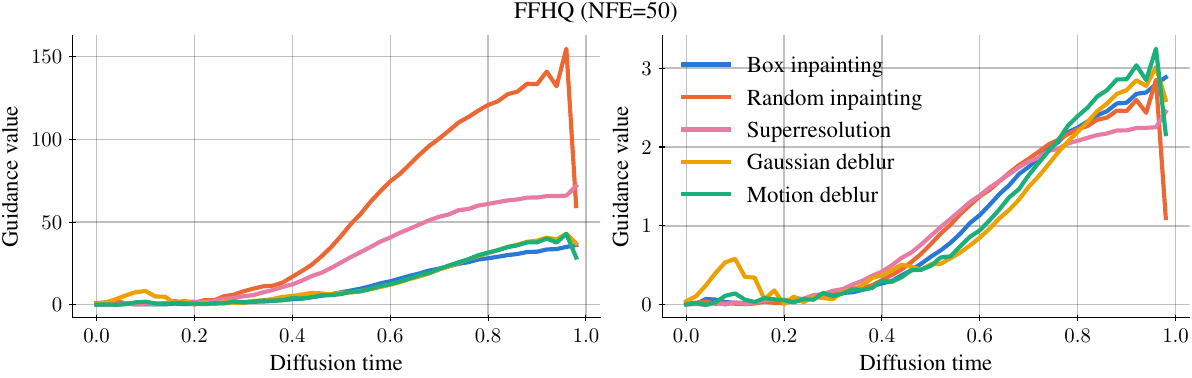}\\[1ex]
    \includegraphics[width=0.9\textwidth]{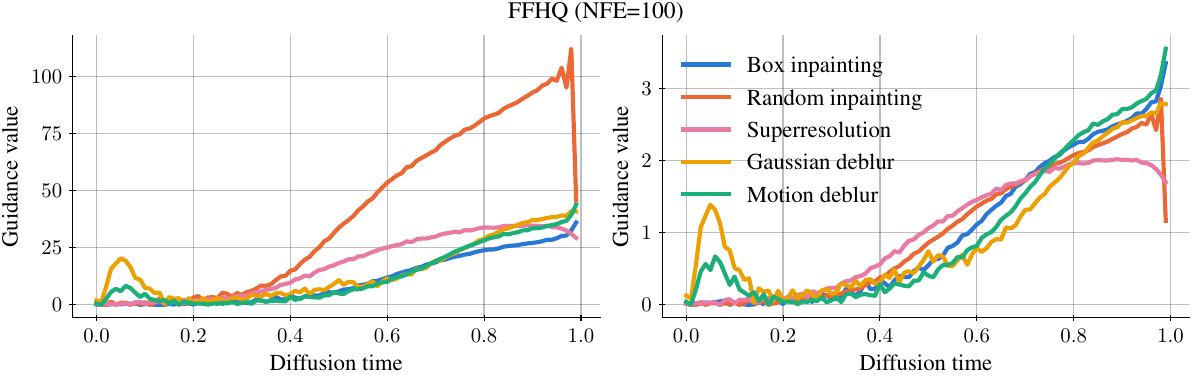}
    \caption{Learned guidance schedules for the five FFHQ noisy inverse tasks against normalized diffusion time ($0$: noisiest, $1$: clean). \emph{Left:} raw. \emph{Right:} each schedule divided by its mean. \emph{Top:} $S=50$. \emph{Bottom:} $S=100$. The schedules differ strongly in magnitude and, to a lesser extent, in shape.}
    \label{fig:crosstask_ffhq}
\end{figure}

\paragraph{Learned schedules.} Figure~\ref{fig:crosstask_ffhq} (left) shows that the learned schedules differ in magnitude: box inpainting and the two deblur tasks cluster together, superresolution is larger, and random inpainting has by far the largest guidance. To compare shapes independently of magnitude, Figure~\ref{fig:crosstask_ffhq} (right) divides each schedule by its mean. The shapes also differ, consistently across both $S$: gaussian deblur rises roughly linearly over the second half of the trajectory, while superresolution and random inpainting rise earliest and level off or drop towards the end, with a sharp drop in the final steps for random inpainting. Superresolution appears to differ the most with respect to shape from the other tasks, with a much more gradual slope as well as a hill shape. 

\begin{figure}[ht]
    \centering
    \includegraphics[width=0.48\textwidth]{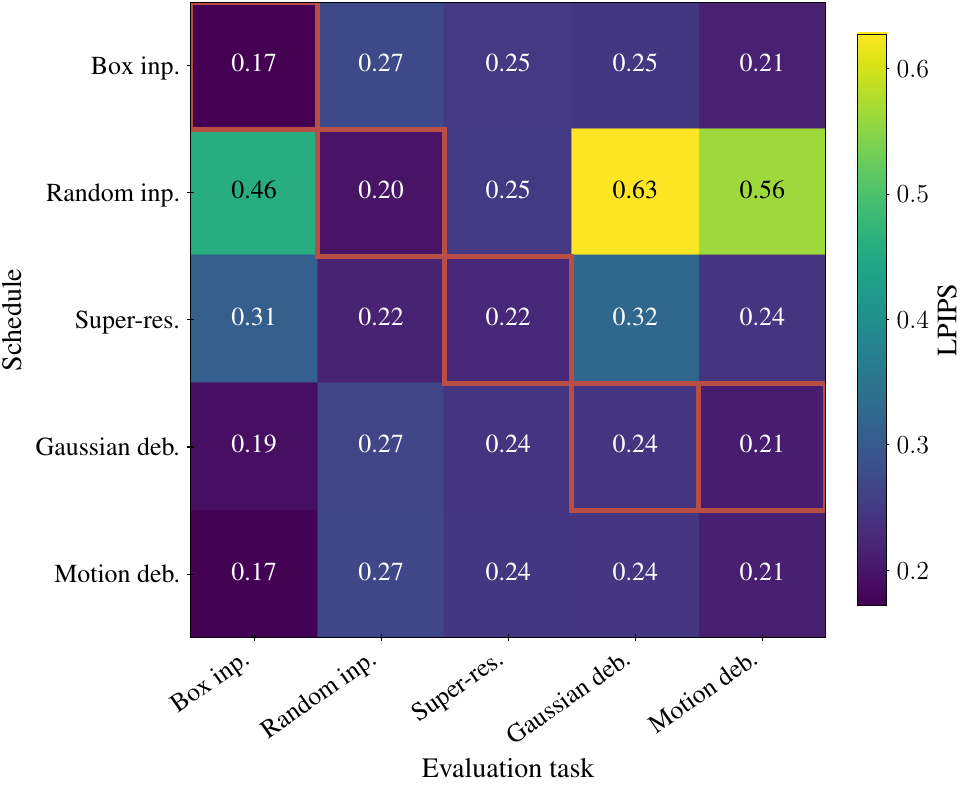}
    \hfill
    \includegraphics[width=0.48\textwidth]{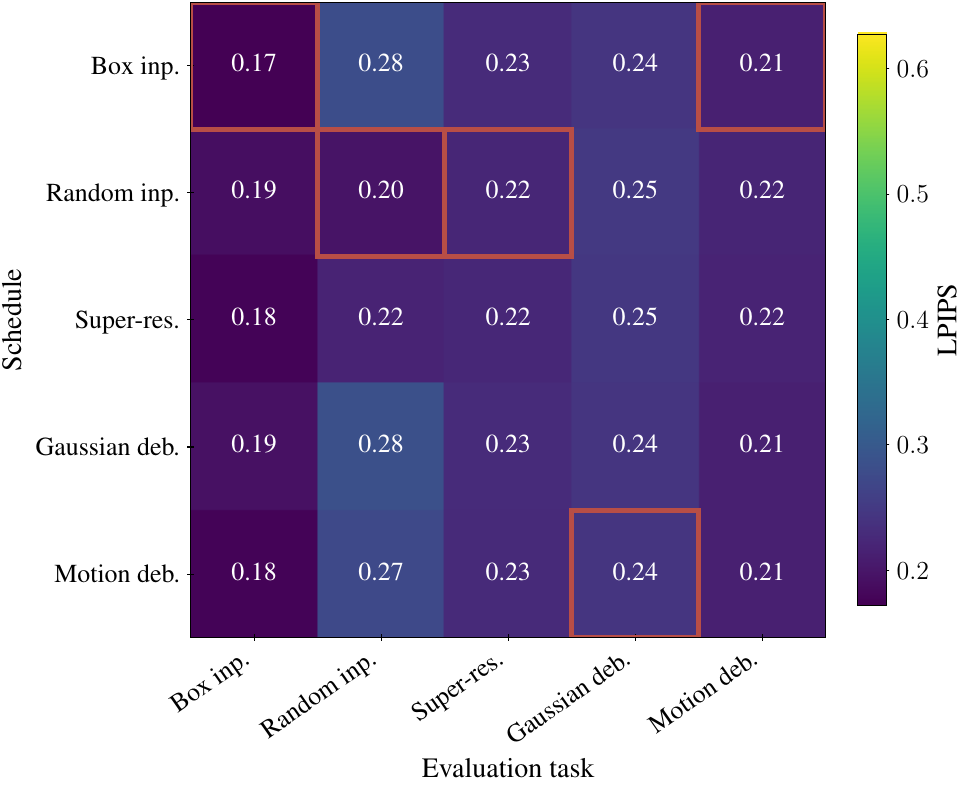}\\[1ex]
    \includegraphics[width=0.48\textwidth]{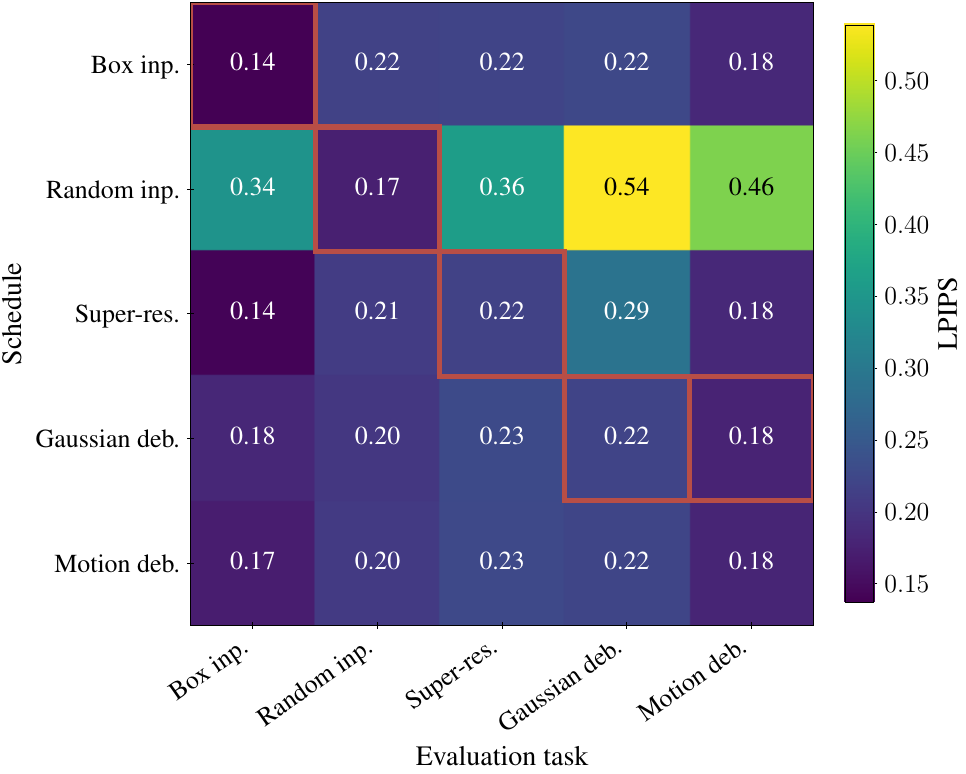}
    \hfill
    \includegraphics[width=0.48\textwidth]{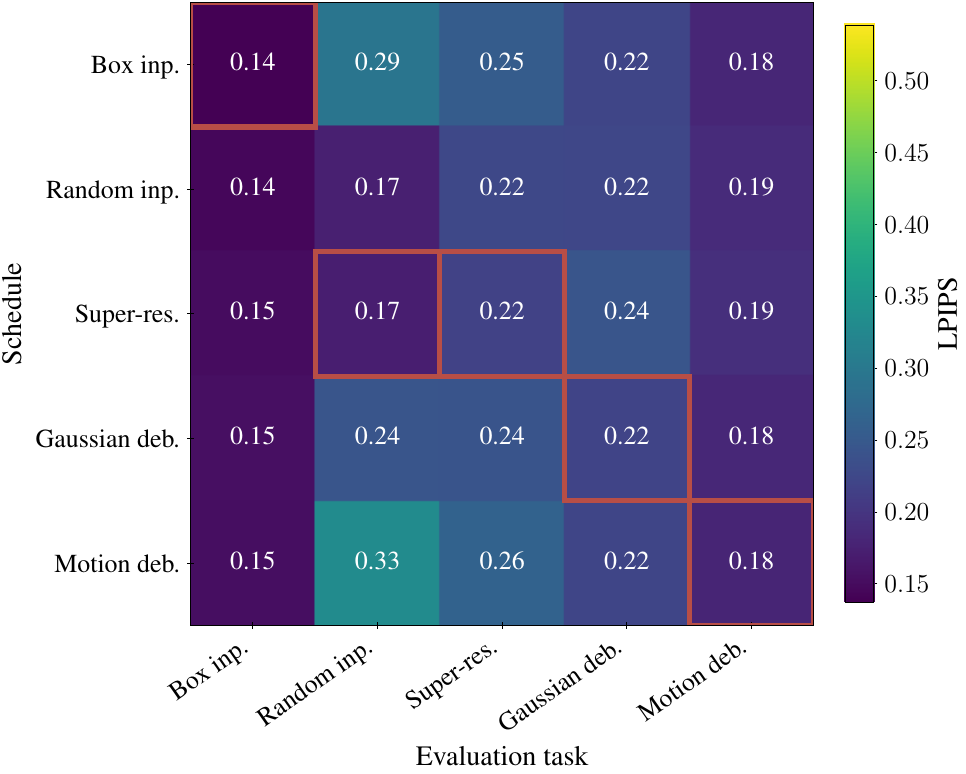}
    \caption{Held-out LPIPS ($300$ FFHQ images) when sampling with the schedule learned for the row task under the measurement operator of the column task. The best entry per column is in bold. \emph{Left:} original schedules. \emph{Right:} off-diagonal schedules rescaled to the mean guidance of the column task's schedule. \emph{Top:} $S=50$. \emph{Bottom:} $S=100$. The matched schedule is best in almost every column before rescaling; after rescaling,  some tasks continue to remain selective with respect to shape.}
    \label{fig:crossgen_matrix}
\end{figure}

\paragraph{Cross-task transfer.} To test how much these differences matter in practice, we sample with the schedule learned for each task (row) under the measurement operator of every task (column), giving a $5\times5$ matrix for each $S$. Each cell is evaluated on the same fixed subset of $300$ images of the test set of Table~\ref{tab:results_ffhq_noisy}; the diagonal is re-evaluated on this subset and therefore differs slightly from the Table~\ref{tab:results_ffhq_noisy}. Figure~\ref{fig:crossgen_matrix} (left) shows that the matched schedule is best in $4$ of the $5$ columns at both $S$. The exception is motion deblur, where the gaussian deblur schedule wins, but the two are nearly identical ($0.209$ vs.\ $0.212$ at $S=50$, $0.177$ vs.\ $0.179$ at $S=100$). This is expected, since both tasks blur the image with a convolution kernel under the same noise model, and their learned schedules are close in magnitude.

\paragraph{Magnitude-matched transfer.} Part of the transfer gap is a magnitude mismatch, most visibly for the random inpainting schedule applied to the deblur tasks. To isolate the effect of shape, we multiply each off-diagonal schedule by a scalar such that its mean guidance over the trajectory equals that of the target task's learned schedule, which leaves the shape unchanged. The diagonal has a factor of $1$ and is reused. Figure~\ref{fig:crossgen_matrix} (right) shows a mixed result, and the matched schedule no longer has a clear advantage. For most target tasks the columns flatten out (e.g.\ $0.242$--$0.247$ for gaussian deblur at $S=50$). Random inpainting is the exception: the random inpainting and superresolution schedules perform best ($0.196$, $0.217$ at $S=50$; $0.173$, $0.171$ at $S=100$), while the other three remain substantially worse ($\geq 0.242$). These are the two schedules that rise earliest and turn down towards the end (Figure~\ref{fig:crosstask_ffhq}, right). Box inpainting is also selective: its own schedule remains best on its task at both $S$ after rescaling.

In summary, task-specific training pays off: the learned schedule is best on its own task in nearly every case, and both of its components (magnitude and shape) matter. The magnitude, which differs by up to a factor of four across tasks, is learned jointly with the shape, and for some tasks (here, random and box inpainting) the task-specific shape gives a further advantage that remains even after matching magnitudes.

\subsection{Generalization across number of steps}
\label{app:cross_nfe}

In this section we investigate how similar the guidance schedules learned at different numbers of steps $S$ are, and whether a schedule learned at one $S$ generalizes to a larger $S$ without retraining. The latter goal may especially be helpful in practical applications where training is especially expensive, e.g. due to a large diffusion backbone. We consider two tasks: FFHQ box inpainting and the Helmholtz forward PDE problem.

\paragraph{Setup.} For each task we train a separate schedule at every $S \in \{10, 20, 30, 50, 100\}$, with all other hyperparameters fixed to the task's main-results configuration (Appendix~\ref{app:exp_details}). Each run is trained for $30$ epochs for box inpainting, and $25$ epochs for Helmholtz, which is sufficient for convergence of train LPIPS and relative error respectively. We evaluate on held-out samples disjoint from the training set: $300$ FFHQ images with LPIPS for box inpainting, and $300$ Helmholtz samples with the relative $L_2$ error on the solution $u$ for Helmholtz.

Given a schedule trained at a source budget $S_{\text{src}}$, we adapt it to a larger target budget $S_{\text{tgt}}$ at inference time using two strategies. \emph{Interp.}\ linearly interpolates the learned schedule of size $S_{\text{src}}$,  as a function of normalized diffusion time onto the larger grid of size $S_{\text{tgt}}$. While this preserves the shape of the schedule, the magnitude of the interpolated schedule may not be calibrated for the new budget. Therefore, we try the \emph{Interp.+rescale} strategy, that additionally multiplies the interpolated schedule by a scalar that corrects its magnitude. We fit a line $f(S)$ to the mean guidance value of the five trained schedules as a function of $S$, and use the multiplier $f(S_{\text{tgt}})/f(S_{\text{src}})$ so that the mean guidance value of the \emph{Interp.+rescale} schedule matches the value $f(S_{\text{tgt}})$ predicted by the fit at the target budget. Both interpolation strategies are compared against the schedule \emph{trained} directly at $S_{\text{tgt}}$.

\begin{figure}[ht]
    \centering
    \includegraphics[width=0.48\textwidth]{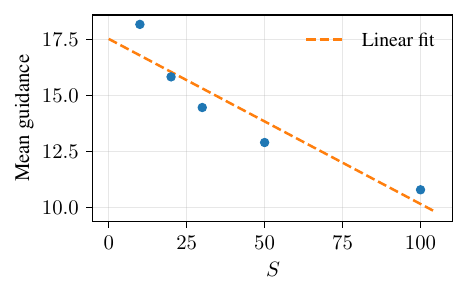}
    \hfill
    \includegraphics[width=0.48\textwidth]{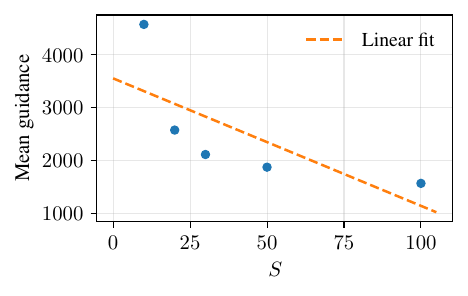}
    \caption{Mean guidance value of the trained schedules as a function of $S$, with the linear fit used for rescaling. \emph{Left:} FFHQ box inpainting. \emph{Right:} Helmholtz forward. The mean guidance decreases monotonically with $S$.}
    \label{fig:cross_nfe_mean_guidance}
\end{figure}

\paragraph{Learned schedules.} Figure~\ref{fig:guidance_schedule_panel} in the main text shows the schedules trained at each $S$. For both tasks the shapes are similar across $S$, while the magnitude decreases with $S$. This is also reflected in Figure~\ref{fig:cross_nfe_mean_guidance}, which shows that the mean guidance decreases monotonically with $S$. We fit a line to the data to give a coarse approximation of this trend to compute rescale factors for the \emph{Interp.+rescale} strategy. The decreasing trend is expected: with more diffusion steps, guidance can be more distributed across the trajectory, so the mean guidance can be smaller while still achieving the same overall effect.

\begin{table}[ht]
\centering
\begin{tabular}{l ccc}
\toprule
$S_{\text{src}} \to S_{\text{tgt}}$ & Trained & Interp. & Interp.+rescale \\
\midrule
$10 \to 20$  & 0.217 & \textbf{0.203} & 0.204 \\
$10 \to 30$  & 0.173 & \textbf{0.169} & 0.170 \\
$10 \to 50$  & 0.140 & 0.142 & \textbf{0.139} \\
$10 \to 100$ & 0.119 & 0.156 & \textbf{0.113} \\
$20 \to 30$  & 0.173 & \textbf{0.172} & 0.174 \\
$20 \to 50$  & 0.140 & \textbf{0.139} & 0.140 \\
$20 \to 100$ & 0.119 & 0.133 & \textbf{0.115} \\
$30 \to 50$  & 0.140 & \textbf{0.138} & 0.141 \\
$30 \to 100$ & 0.119 & 0.126 & \textbf{0.115} \\
$50 \to 100$ & 0.119 & 0.122 & \textbf{0.114} \\
\bottomrule
\end{tabular}
\vspace{2mm}
\caption{FFHQ box inpainting: LPIPS ($\downarrow$) on $300$ held-out samples. \textbf{Bold}: best of the three columns.}
\label{tab:cross_nfe_box_inpainting}
\end{table}

\begin{table}[ht]
\centering
\begin{tabular}{l ccc}
\toprule
$S_{\text{src}} \to S_{\text{tgt}}$ & Trained & Interp. & Interp.+rescale \\
\midrule
$10 \to 20$  & 11.70\% & \textbf{9.80\%} & 10.01\% \\
$10 \to 30$  & 10.34\% & 9.13\% & \textbf{9.01\%} \\
$10 \to 50$  & 9.33\% & 9.88\% & \textbf{8.73\%} \\
$10 \to 100$ & 9.07\% & 15.05\% & \textbf{8.77\%} \\
$20 \to 30$  & 10.34\% & \textbf{9.54\%} & 9.77\% \\
$20 \to 50$  & 9.33\% & 9.09\% & \textbf{9.04\%} \\
$20 \to 100$ & \textbf{9.07\%} & 11.73\% & 9.14\% \\
$30 \to 50$  & 9.33\% & \textbf{9.06\%} & 9.24\% \\
$30 \to 100$ & \textbf{9.07\%} & 11.11\% & 9.36\% \\
$50 \to 100$ & \textbf{9.07\%} & 10.13\% & 9.42\% \\
\bottomrule
\end{tabular}
\vspace{2mm}
\caption{Helmholtz forward: relative $L_2$ error on $u$ (\%, $\downarrow$) on $300$ held-out samples. \textbf{Bold}: best of the three columns.}
\label{tab:cross_nfe_helmholtz}
\end{table}

\paragraph{Interpolation and rescaling.} Tables~\ref{tab:cross_nfe_box_inpainting} and~\ref{tab:cross_nfe_helmholtz} show that plain interpolation performs well for small gaps between $S_{\text{src}}$ and $S_{\text{tgt}}$, often matching or even beating the trained schedule (the latter could indicate that the guidance magnitude for the target had not yet converged to the optimal value), but degrades as the gap grows. However interpolating from $S_{\tx{src}} \in \set{10, 20, 30, 50}$ to $S_{\text{tgt}} = 100$ performs consistently worse than the trained schedule across both tasks. Rescaling largely removes this degradation: Interp.+rescale performs very close to the trained schedule in every row of both tables. For the smallest gaps, where there is little magnitude mismatch to correct, plain interpolation is sometimes marginally better.

In conclusion, we see that the schedules learned for the same task at different $S$ have a strongly shared shape, and the trend in their mean guidance value is predictable. The latter fact can be leveraged to improve the performance of LEEGS by training at a small number of steps and extrapolating the schedule to a larger number of steps when the computational budget for training is limited.

\subsection{Ablation on trajectory objective vs terminal objective}
\label{app:trajectory_objective}

This section studies the effect of dropping the non-terminal (trajectory-level) constraint term from the training objective, keeping only the terminal constraint $h(x_0, c)$, where $x_0$ is the final denoised sample. In Section \ref{sec:method} we motivated the trajectory objective by noting that the terminal constraint's gradient with respect to the guidance scale $w_s$ decays for steps $s$ far from the end of the trajectory, so a terminal-only objective gives it progressively less usable training signal as the trajectory gets longer. We now verify this intuition empirically for the box inpainting task on FFHQ, using the same training setup as in the cross-step-budget experiment of Appendix~\ref{app:cross_nfe}. We reuse the  trained schedules from the cross-step-budget experiment of Appendix~\ref{app:cross_nfe} ($S \in \{10, 20, 30, 50, 100\}$ for box inpainting, which wer trained for $30$ epochs each with the full LEEGS objective $\LL(W)$ (which we call "Trajectory"). Using the same hyperparameters, we train a \textit{terminal-only} variant for each $S$, where the loss function is $\LL'(W)=\Ex{c,x_s}{h(x_0, c)}$. Both variants are evaluated on the same held-out set of $300$ samples used for the cross-step-budget study.

\begin{figure}[ht]
    \centering
    \includegraphics[width=0.32\textwidth]{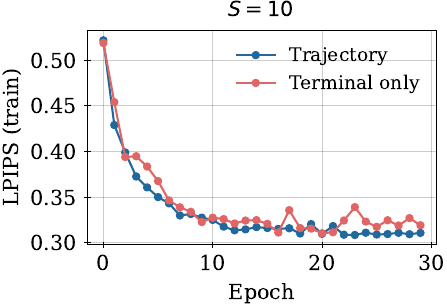}
    \includegraphics[width=0.32\textwidth]{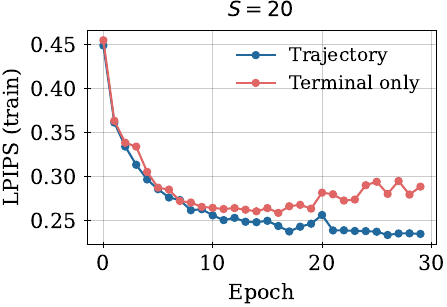}
    \includegraphics[width=0.32\textwidth]{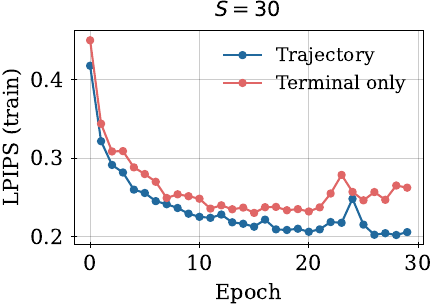}\\
    \includegraphics[width=0.32\textwidth]{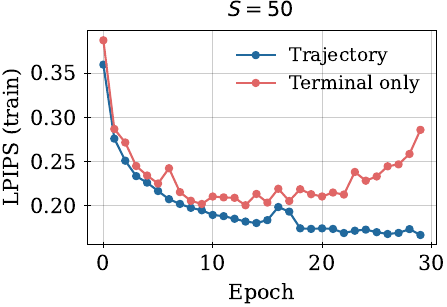}
    \includegraphics[width=0.32\textwidth]{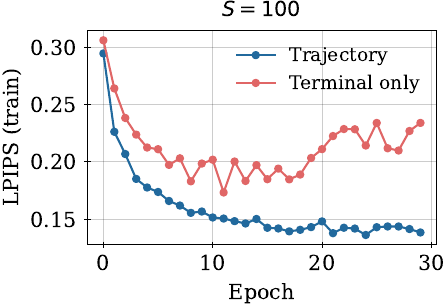}
    \caption{Training LPIPS vs.\ epoch, full vs.\ terminal-only constraint set, one panel per
    diffusion step budget $S$.}
    \label{fig:trajectory_objective_train_lpips}
\end{figure}

\begin{figure}[ht]
    \centering
    \includegraphics[width=0.55\textwidth]{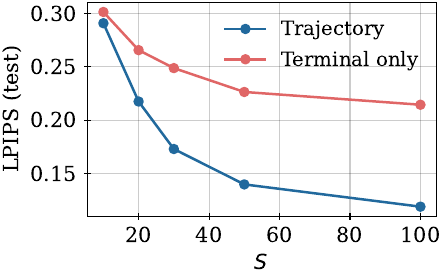}
    \caption{Held-out LPIPS vs.\ $S$, full vs.\ terminal-only constraint set.}
    \label{fig:trajectory_objective_eval_lpips}
\end{figure}

\paragraph{Observations.} Figure \ref{fig:trajectory_objective_train_lpips} shows the training LPIPS vs.\ epoch for the full (trajectory) and terminal-only objectives, for each $S$. We see that the training curve for the terminal-only objective is consistently worse than the full objective, and the gap widens with $S$. Moreover, the terminal-only objective also exhibits a reward-hacking elbow at which the terminal constraint keeps improving while the reconstruction quality degrades. However, note that the gap between the two objectives with respect to training LPIPS cannot be completely attributed to reward-hacking, since there is a gap even if we consider the best training LPIPS achieved by each objective. Figure \ref{fig:trajectory_objective_eval_lpips} shows the held-out LPIPS vs.\ $S$ for both objectives. Here it is even more visually clear that the gap between the two objectives widens with $S$. Moreover, we see a plateauing effect for the LPIPS with increasing $S$ for both runs, but the terminal-only curve is much flatter than the full objective curve, indicating that the terminal-only objective is less able to leverage additional diffusion steps to improve reconstruction quality.
\clearpage
\newpage
\section{Extension to sampling with renoising}
\label{app:time_travel}

\begin{figure}[ht]
    \centering
    \includegraphics[width=0.32\textwidth]{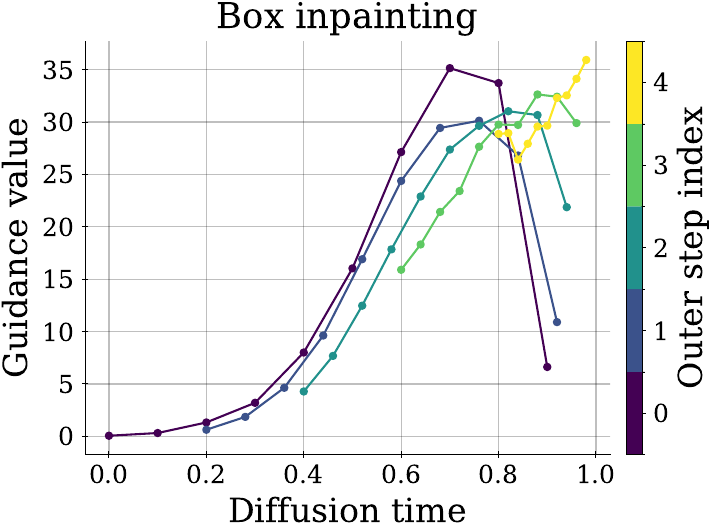}
    \hfill
    \includegraphics[width=0.32\textwidth]{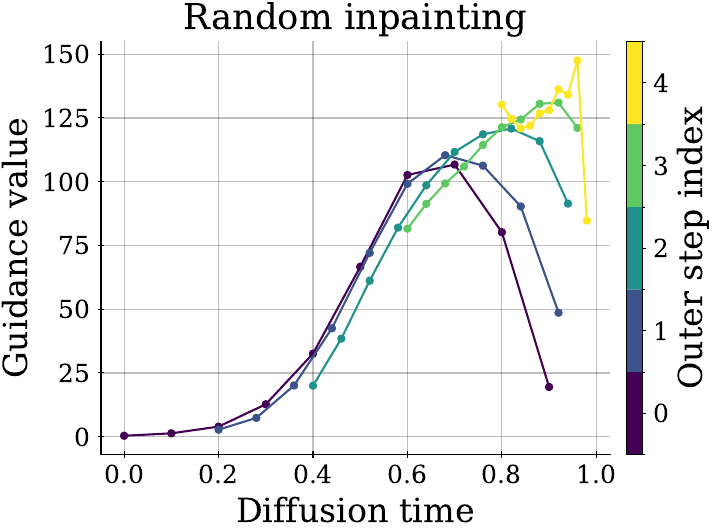}
    \hfill
    \includegraphics[width=0.32\textwidth]{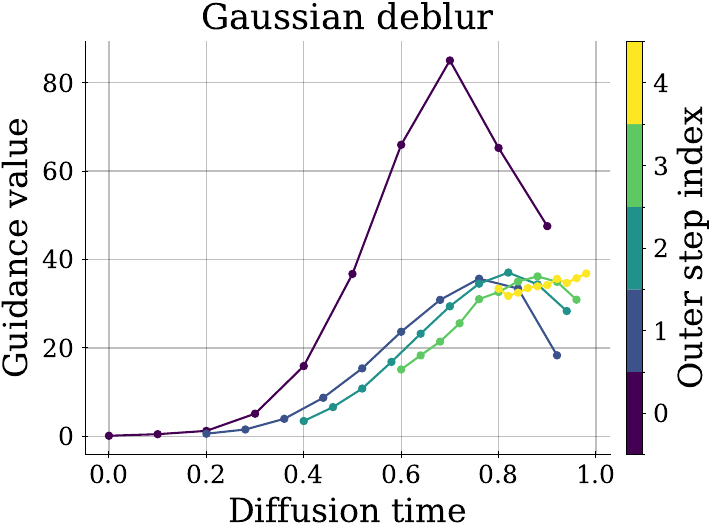}\\[1ex]
    \includegraphics[width=0.32\textwidth]{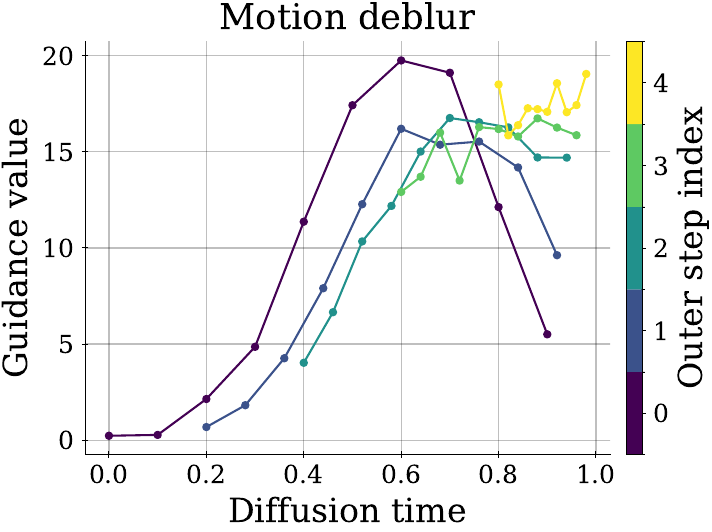}
    \hspace{0.02\textwidth}
    \includegraphics[width=0.32\textwidth]{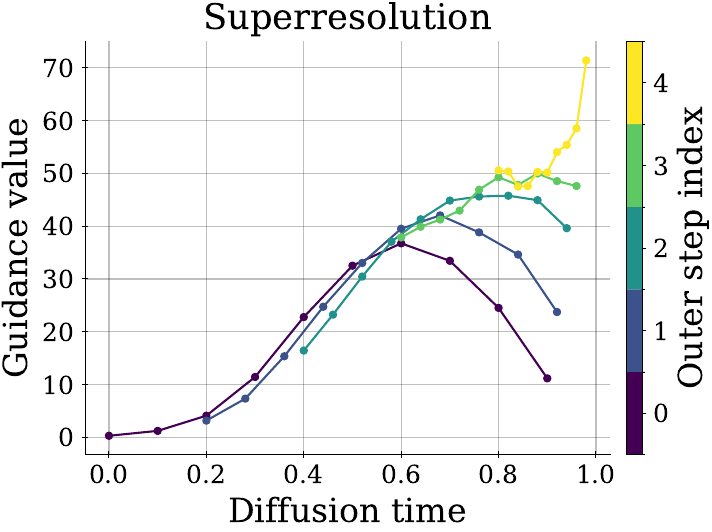}
    \caption{Learned LEEGS-renoise guidance schedules for the five FFHQ tasks ($S=50$, $K=5$) against normalized diffusion time ($0$: noisiest, $1$: clean); color indicates the outer step. Every outer step except the last has a hill shape.}
    \label{fig:time_travel_schedules}
\end{figure}

This section applies LEEGS to a sampler that alternates guided reverse diffusion with renoising, i.e.\ forward diffusion back to a noisier state. Renoising is used by, e.g., RePaint \citep{lugmayr2022repaintinpaintingusingdenoising}, DDNM \citep{wang2022zero}, which calls it time travel, and DAPS \citep{zhang2025improvingdiffusioninverseproblem}. We denote this variant LEEGS-renoise, and the standard DPS-style sampling used in the rest of the paper simply LEEGS. Using normalized diffusion time $\tau \in [0, 1]$ ($0$: noisiest, $1$: clean), the sampler splits the $S$ guided steps into $K$ outer steps of $S/K$ inner steps each. Outer step $j = 0, \ldots, K-1$ starts at $\tau = j/K$ and reverse-diffuses with guidance to the clean end $\tau = 1$ over $S/K$ uniformly spaced steps. The resulting sample is then renoised in without guidance, to the start $\tau = (j+1)/K$ of the next outer step, using the forward process.
The sampler therefore moves back and forth between the clean and noisy ends, with progressively later starting points after  each outer step. Thus, each outer step also covers a shorter interval with finer steps than the previous one. The final sample is the output of the last outer step. The learned schedule has one weight per guided step, i.e.\ $S$ weights in total, and the NFE budget is the same as for LEEGS with $S$ steps, since renoising requires no network evaluation. For example, in Figure~\ref{fig:time_travel_schedules} (box inpainting), the five outer steps start at $\tau = 0, 0.2, 0.4, 0.6, 0.8$, and each has its own $10$ weights. Training only changes how the trajectory is unrolled: only \textsc{ComputeGradient} of Algorithm~\ref{alg:leegs} is replaced, by the version in Algorithm~\ref{alg:leegs_time_travel}; \textsc{GuidedStep} itself is unchanged.

\begin{algorithm}[h]
    \captionof{algorithm}{LEEGS-renoise training (replaces \textsc{ComputeGradient} of Algorithm~\ref{alg:leegs})}
    \label{alg:leegs_time_travel}
    \begin{algorithmic}[1]
        \Function{ComputeGradient}{$W, c, \texttt{compute\_leegs\_approx}$}
            \State $x \sim \mathcal{N}(0, \mathbf{I})$
            \State $\ell \gets 0$; $\ i \gets 1$ \Comment{$i$ is the flat index into $W = (w_1, \ldots, w_S)$}
            \For{$j = 0, \ldots, K-1$} \Comment{outer step}
                \For{$s = 1, \ldots, S/K$} \Comment{inner step, from $\tau = j/K$ to $\tau = 1$}
                    \State $t \gets \texttt{time\_at}(j, s)$ \Comment{$s$-th point of linspace$(j/K,\, 1,\, S/K+1)$}
                    \State $(x, h_i) \gets \textsc{Checkpoint}\big(\textsc{GuidedStep},\ x,\ t,\ w_i,\ c,\ \texttt{compute\_leegs\_approx}\big)$
                    \State $\ell \gets \ell + h_i$; $\ i \gets i + 1$
                \EndFor
                \If{$j < K - 1$}
                    \State $\eps \sim \mathcal{N}(0, \mathbf{I})$
                    \State $x \gets \textsc{Renoise}\big(x, (j+1)/K , \eps \big) $ \Comment{renoise to the start of outer step $j+1$}
                \EndIf
            \EndFor
            \State $h_0 \gets \texttt{compute\_guidance\_function}(x, c)$ \Comment{terminal term $h(x_0, c)$}
            \State $\ell \gets \ell + h_0$
            \State \Return $\textsc{Backward}(\ell,\ W)$
        \EndFunction
    \end{algorithmic}
\end{algorithm}

\paragraph{Training and evaluation setup.} We use $K=5$ outer steps of $S/K=10$ inner steps each ($S=50$) on all five FFHQ noisy inverse tasks. The learning rates, epochs and weight decay are given in Table~\ref{tab:cfg_time_travel}; all other hyperparameters match the LEEGS runs of Table~\ref{tab:results_ffhq_noisy} (Appendix~\ref{app:exp_details}). Table \ref{tab:cfg_time_travel} shows the results of evaluating LEEGS-renoise against the first random run of LEEGS in Table~\ref{tab:results_ffhq_noisy}, i.e. they share the same training and test set of 950 images. The relative change is computed as $(\text{LEEGS-renoise} - \text{LEEGS}) / \text{LEEGS}$, so negative values indicate that LEEGS-renoise outperforms LEEGS.

\begin{table}[ht]
    \small
\centering
\begin{tabular}{l ccc}
\toprule
Subtask & Learning rate $\eta$ & Epochs & Weight decay \\
\midrule
Super-resolution ($4\times$) & 0.3 & 15 & 0.0 \\
Box inpainting & 0.2 & 30 & 0.0 \\
Random inpainting & 0.8 & 30 & 0.0 \\
Gaussian deblur & 0.5 & 25 & 0.0 \\
Motion deblur & 0.3 & 15 & 0.0 \\
\bottomrule
\end{tabular}
\vspace{2mm}
\caption{LEEGS-renoise training configuration, $S=50$, $K=5$.}
\label{tab:cfg_time_travel}
\end{table}

\begin{table}[ht]
\centering
\begin{tabular}{l c c c}
\hline
Task & LEEGS & LEEGS-renoise & Relative change \\
\hline
Box inpainting & 0.174 & \textbf{0.165} & $-5.2\%$ \\
Random inpainting & 0.196 & \textbf{0.181} & $-7.5\%$ \\
Gaussian deblur & \textbf{0.243} & 0.246 & $+1.3\%$ \\
Motion deblur & \textbf{0.212} & 0.227 & $+7.0\%$ \\
Superresolution & \textbf{0.222} & 0.224 & $+0.6\%$ \\
\hline
\end{tabular}
\vspace{2mm}
\caption{LPIPS on the FFHQ test set ($950$ images), $S=50$.}
\label{tab:time_travel_vs_e2e}
\end{table}

\paragraph{Results.} Table~\ref{tab:time_travel_vs_e2e} shows that LEEGS-renoise is comparable to LEEGS overall, and improves on both inpainting tasks ($-5.2\%$ for box, $-7.5\%$ for random inpainting).

\paragraph{Learned schedules.} Figure~\ref{fig:time_travel_schedules} shows the learned schedules. The magnitude varies somewhat between outer steps. Every outer step except the last has a hill shape, rising and then turning down towards its end. The last outer step instead increases roughly linearly, except for random inpainting, where it drops sharply at the final step. Similar to the observation for LEEGS (Appendix~\ref{app:cross_task}), the learned schedules for superresolution and random inpainting are also similar for LEEGS-renoise.  For both, the peak guidance grows with each outer step, whereas for the other tasks it stays roughly constant after the first outer step. Gaussian deblur is an outlier, with an anomalously large magnitude in the first outer step (peak $\approx 85$, vs.\ $\approx 37$ in later outer steps).

\paragraph{Conclusion.} This experiment shows that the main idea of this paper (learning time dependent guidance scales to minimize guidance functions by backpropagating through the sampling trajectory) is applicable to diverse scenarios beyond DPS-style sampling. Moreover, while DPS-style sampling may admit competitive simple schedules, such as the increasing linear schedule in Table~\ref{tab:results_ffhq_noisy}, it is much harder to tune or approximate a schedule by hand for a sampler with renoising, whose schedule is a sequence of per-outer-step segments. For LEEGS, the cost is the same in both cases, since it depends only on the number of NFEs in the sampling trajectory.

\clearpage
\section{Comparison at $S=10$ NFEs (FFHQ)}
\label{app:s10_comparison}

As a further stress test at a much smaller NFE budget, we repeat the noisy-inverse-problem
comparison of Table~\ref{tab:results_ffhq_noisy} at $S=10$ NFEs, on FFHQ. We compare against Const.,
DAPS, DSG, CDIM and Linear increase, using the same hyperparameter-selection protocol as
Appendix~\ref{app:exp_details} (grid search on $20$ validation images for Const./DSG/Linear
increase, and a discrete-configuration search for DAPS/CDIM), evaluated on the same $950$-image
FFHQ test set as Table~\ref{tab:results_ffhq_noisy}. Linear decrease is again omitted (it is
uniformly the worst baseline at every other NFE budget in this paper); we did not extend the
comparison to ImageNet or run the $1000$-NFE DPS reference at $S=10$. Unlike the main
experiments (three independent training runs per subtask), we train a single LEEGS run per
subtask for this $S=10$ extension. Table~\ref{tab:cfg_noisy_s10} gives the learning rate, number
of epochs and weight decay used.

\begin{table}[ht]
    \small
\centering
\begin{tabular}{l ccc}
\toprule
Subtask & Learning rate $\eta$ & Epochs & Weight decay \\
\midrule
Super-resolution ($4\times$) & 0.3 & 30 & 0.0 \\
Box inpainting & 0.2 & 20 & 0.0 \\
Random inpainting & 0.8 & 30 & 0.0 \\
Gaussian deblur & 0.3 & 20 & 0.0 \\
Motion deblur & 0.3 & 20 & 0.0 \\
\bottomrule
\end{tabular}
\vspace{2mm}
\caption{LEEGS training configuration for the $S=10$ FFHQ extension (Table \ref{tab:results_ffhq_noisy_s10}).}
\label{tab:cfg_noisy_s10}
\end{table}

\begin{table}[ht]
\centering
\begin{adjustbox}{max width=\linewidth}
\begin{tabular}{l l cc cc cc cc cc}
\hline
\vspace{1mm}
\multirow{2}{*}{} & \multirow{2}{*}{NFEs} & \multicolumn{2}{c}{SR (x4)} & \multicolumn{2}{c}{Inpaint (box)} & \multicolumn{2}{c}{Inpaint (random)} & \multicolumn{2}{c}{Deblur (gauss)} & \multicolumn{2}{c}{Deblur (motion)} \\ \vspace{1mm}
 &  &LPIPS &FID &LPIPS &FID &LPIPS &FID &LPIPS &FID &LPIPS &FID \\ \hline
Const. & 10 & 0.562 & 104.92 & 0.569 & 102.78 & 0.583 & 102.25 & 0.563 & 108.08 & 0.570 & 103.75 \\
DAPS & 10 & 0.331 & 60.55 & \best{0.207} & \scnd{41.14} & 0.619 & 234.89 & \scnd{0.323} & \scnd{57.76} & \best{0.294} & \best{58.96} \\
DSG & 10 & 0.588 & 133.25 & 0.616 & 127.00 & 0.615 & 140.01 & 0.530 & 131.50 & 0.535 & 127.15 \\
CDIM & $\approx$ 10 & \scnd{0.297} & \best{40.11} & \scnd{0.217} & \best{33.10} & \scnd{0.363} & \scnd{75.65} & 0.326 & \best{40.19} & 0.411 & 69.11 \\
Linear increase & 10 & 0.469 & 100.31 & 0.499 & 99.20 & 0.483 & 99.07 & 0.479 & 96.26 & 0.474 & 97.06 \\
\textbf{LEEGS (ours)} & 10 & \best{0.289} & \scnd{54.88} & 0.323 & \scnd{53.45} & \best{0.305} & \best{51.34} & \best{0.322} & 63.21 & \scnd{0.305} & \scnd{60.05} \\ \hline
\end{tabular}
\end{adjustbox}
\vspace{2mm}
\caption{Comparison for noisy image inverse tasks (FFHQ), at $S=10$ NFEs. \textbf{Green} (bold): best; \textbf{blue}: second best.}
\label{tab:results_ffhq_noisy_s10}
\end{table}

As at $S=50$ and $S=100$ (see the CDIM item in Appendix~\ref{app:exp_details} and
Table~\ref{tab:cdim_nfes}), CDIM's NFE count varies per sample; the value $T=4$ used here was
chosen as the closest match to $10$ NFEs among $T \in \{4,5,6\}$, and its actual average NFEs
per task range from $9.7$ to $16.6$ (mean $12.4$ across the five tasks).

Overall, LEEGS has the best performance in this comparison: it achieves the best result on four
of the ten columns (SR and random-inpainting LPIPS, random-inpainting FID, and Gaussian-deblur
LPIPS) and is best or second-best on eight of the ten. DAPS and CDIM are the strongest
competitors, each best on three columns, with the latter having an advantage with respect to NFEs. Meanwhile, DSG and
Const.\ perform very poorly, and the Linear increase baseline -- a
strong heuristic at higher NFE budgets -- also fails to be the best at any task at this very low budget.

\begin{figure}[ht]
    \centering
    \includegraphics[width=\textwidth]{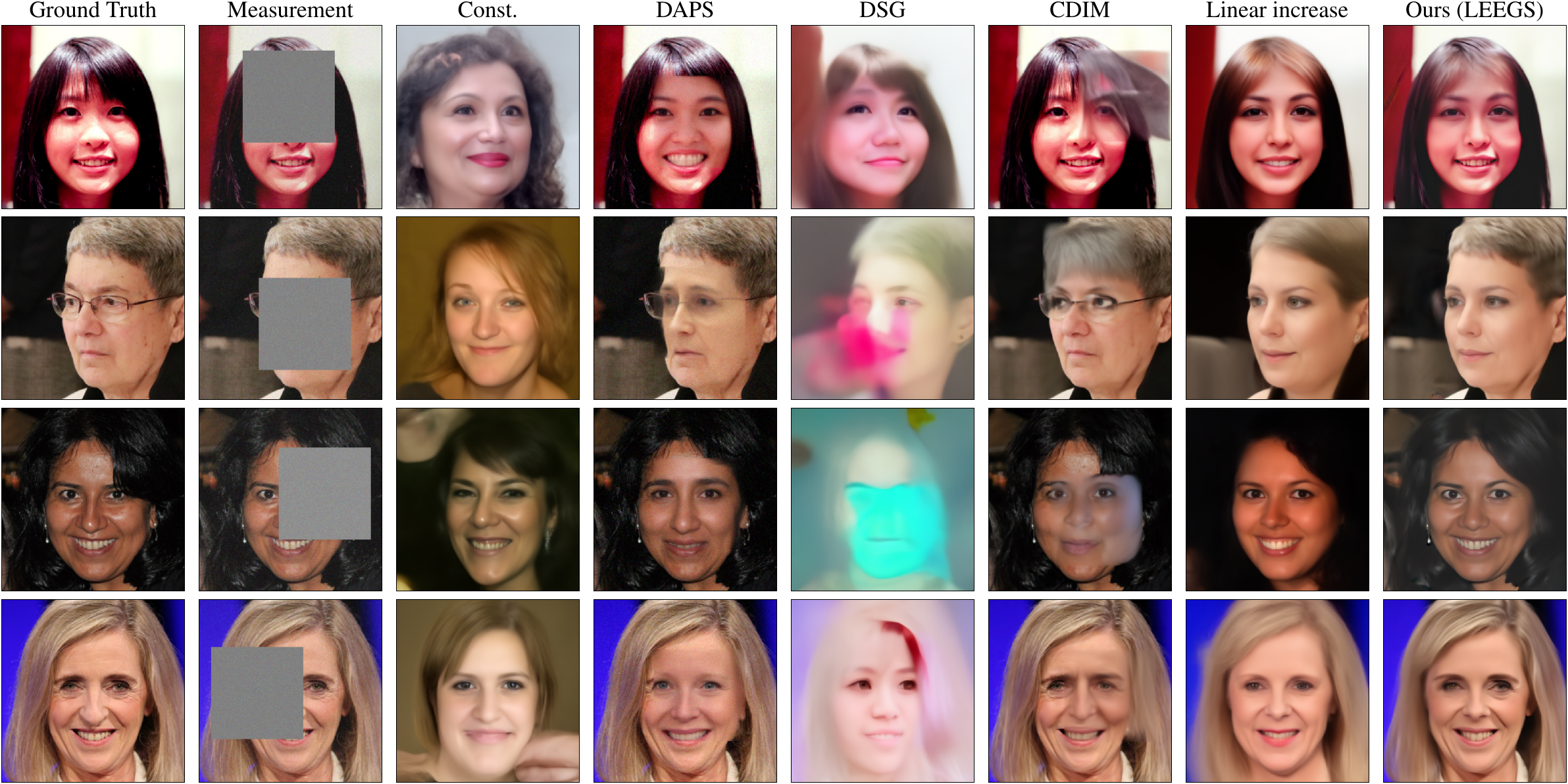}
    \caption{Box inpainting, FFHQ, $10$ NFEs.}
    \label{fig:baseline_panel_box_inpainting_ffhq_s10}
\end{figure}

\begin{figure}[ht]
    \centering
    \includegraphics[width=\textwidth]{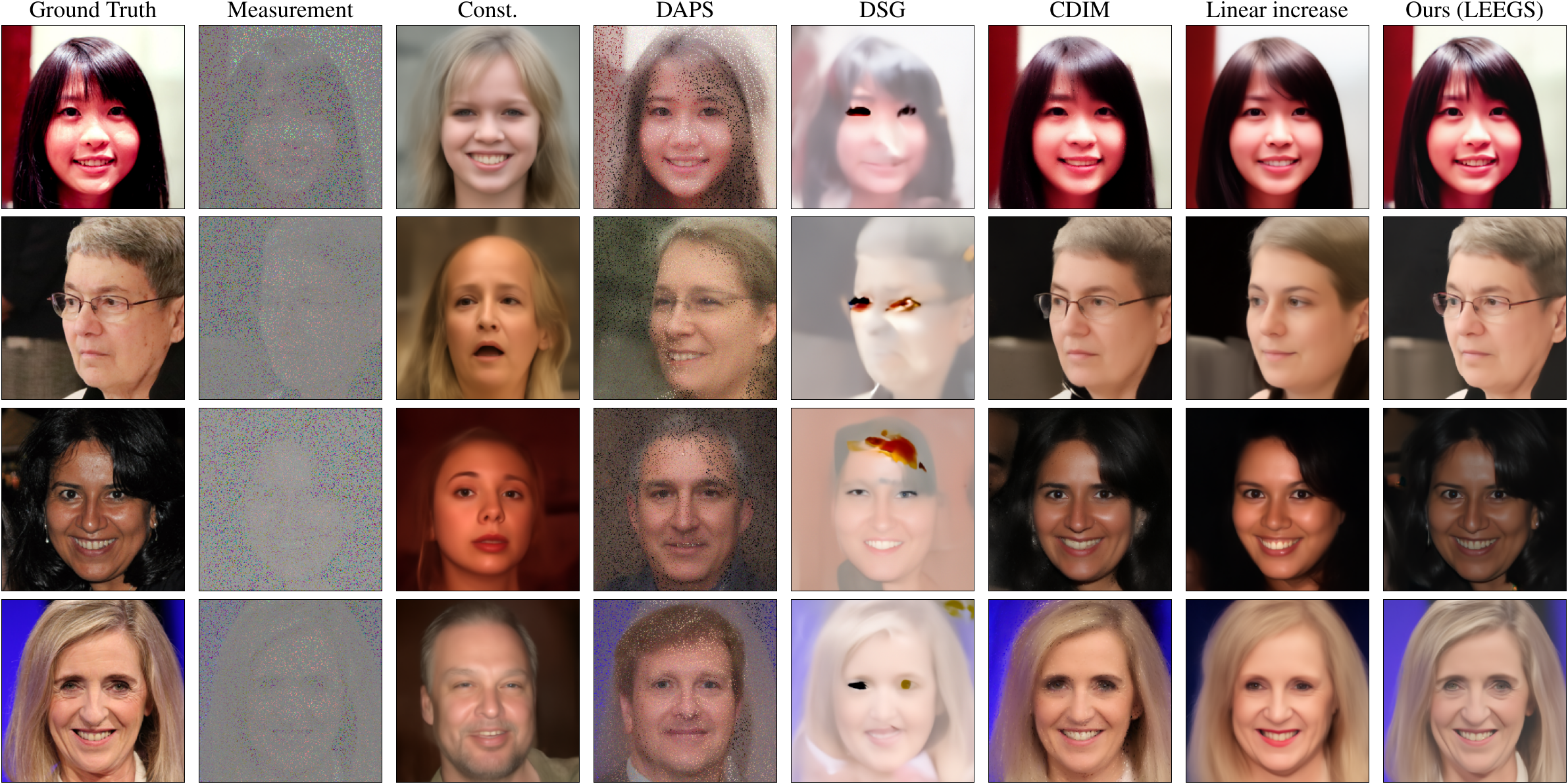}
    \caption{Random inpainting, FFHQ, $10$ NFEs.}
    \label{fig:baseline_panel_random_inpainting_ffhq_s10}
\end{figure}

\begin{figure}[ht]
    \centering
    \includegraphics[width=\textwidth]{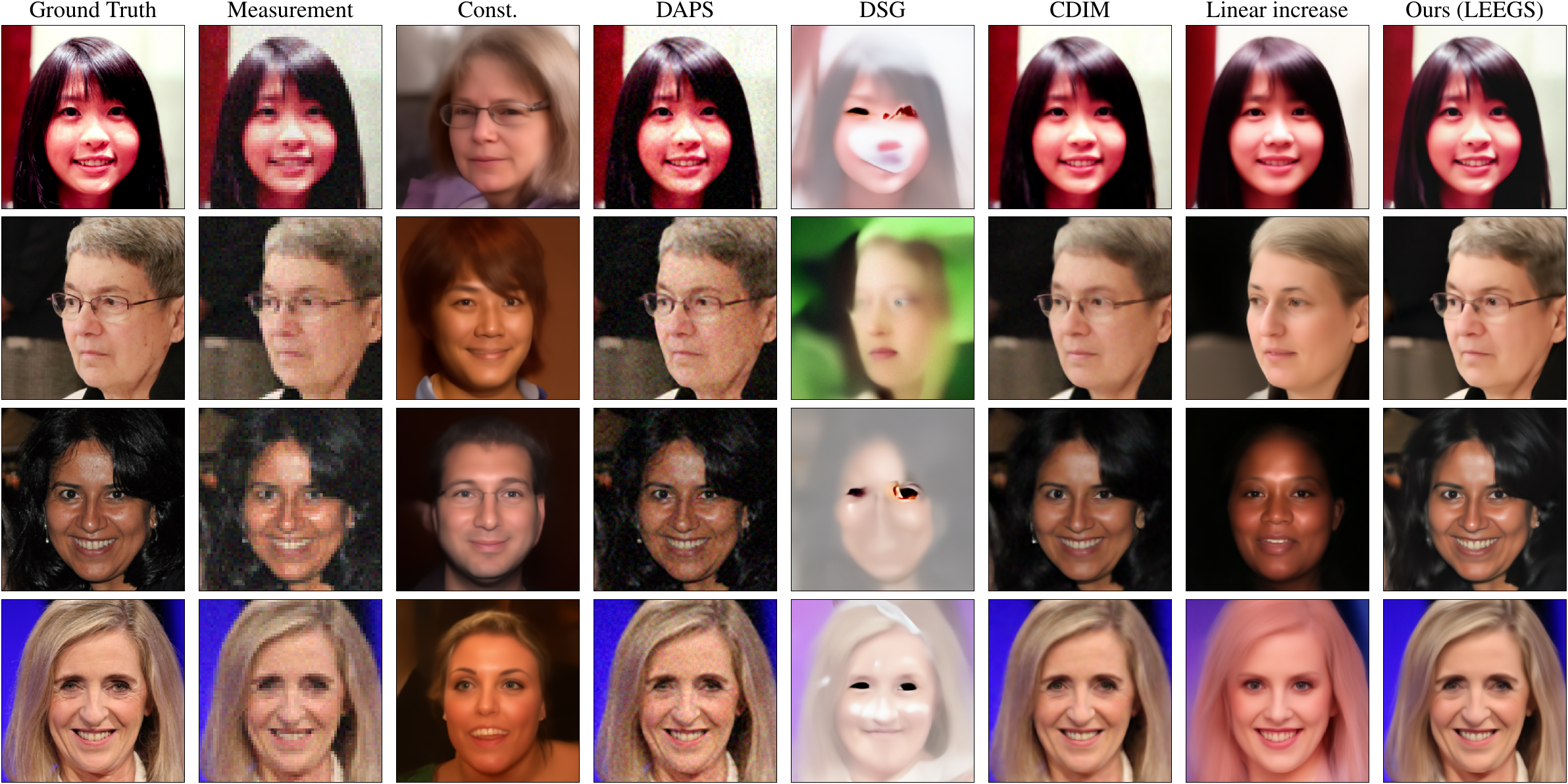}
    \caption{Superresolution ($\times 4$), FFHQ, $10$ NFEs.}
    \label{fig:baseline_panel_superresolution_ffhq_s10}
\end{figure}

\begin{figure}[ht]
    \centering
    \includegraphics[width=\textwidth]{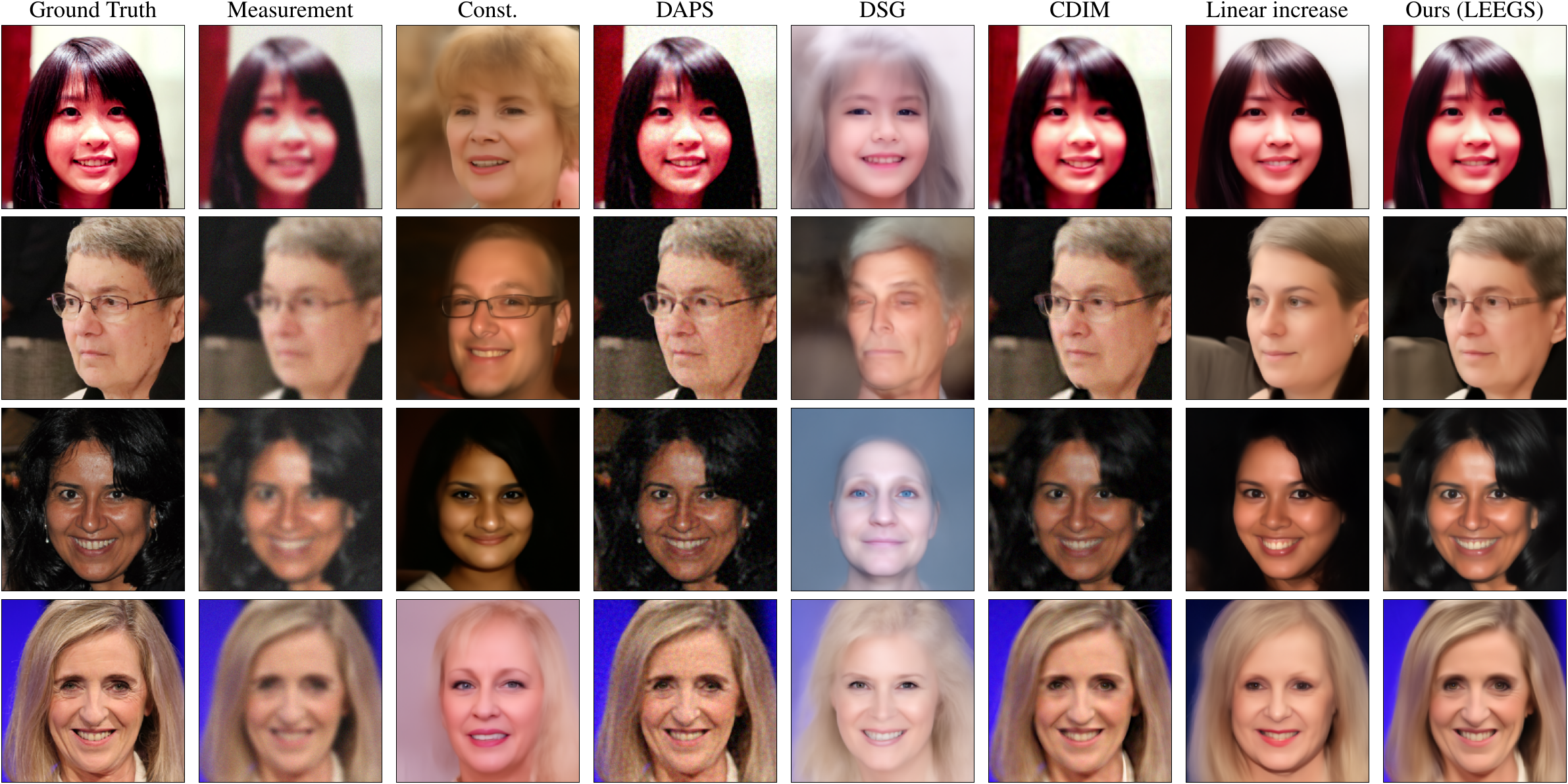}
    \caption{Gaussian deblur, FFHQ, $10$ NFEs.}
    \label{fig:baseline_panel_gaussian_deblur_ffhq_s10}
\end{figure}

\begin{figure}[ht]
    \centering
    \includegraphics[width=\textwidth]{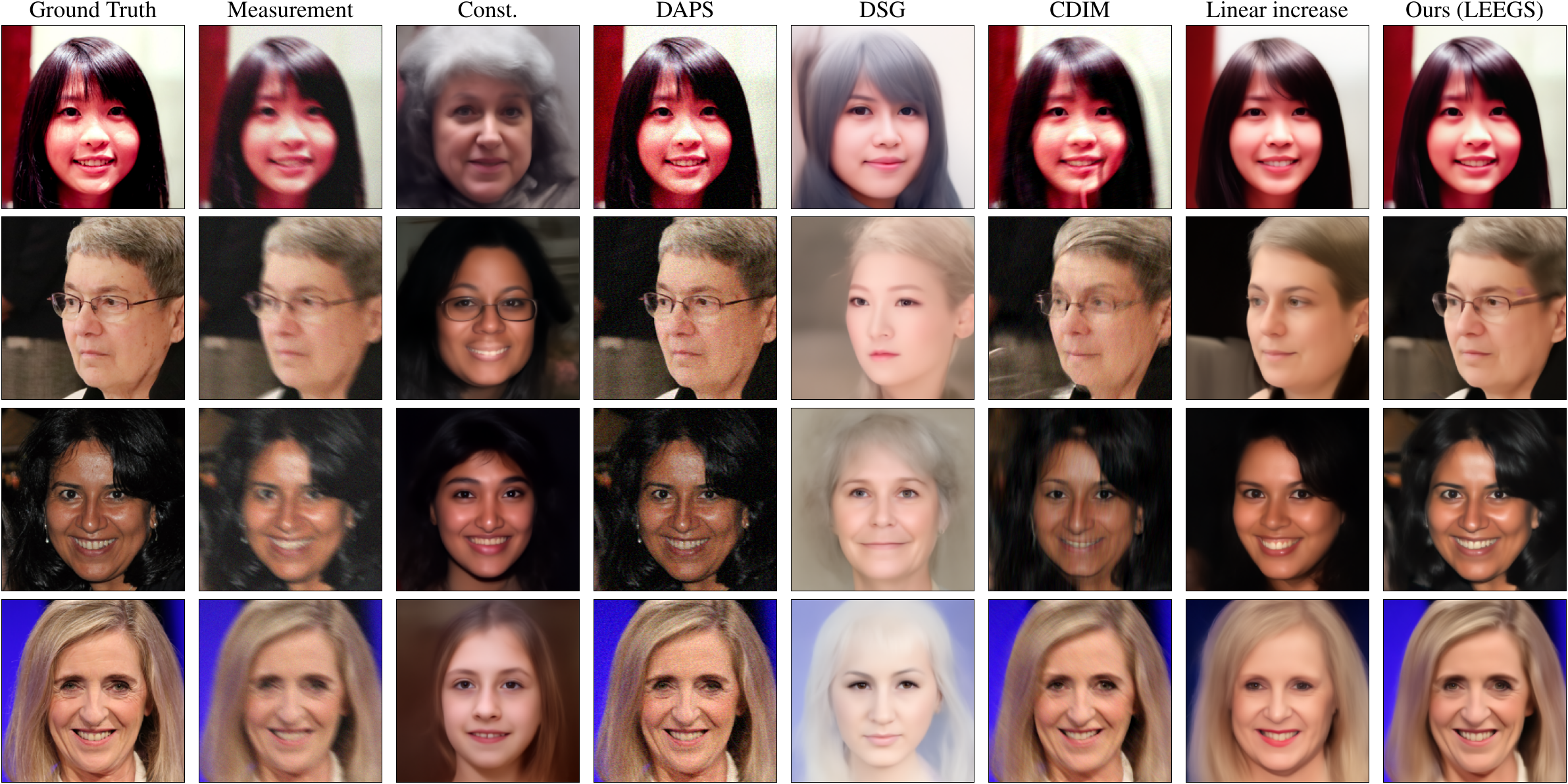}
    \caption{Motion deblur, FFHQ, $10$ NFEs.}
    \label{fig:baseline_panel_motion_deblur_ffhq_s10}
\end{figure}

\clearpage
\section{Additional metrics for main experiments}
\label{app:additional_metrics_main_tables}

This section gives the variance of LEEGS across training draws for the main results, and additional metrics for the noisy inverse and image inpainting problems.

\subsection{Variance of LEEGS across draws}
\label{app:leegs_std}

The LEEGS rows of Tables~\ref{tab:results_ffhq_noisy}, \ref{tab:results_imagenet_noisy}, \ref{tab:results_inpainting_celeba}, \ref{tab:results_faceid_celeba} and~\ref{tab:results_dpde} are means over $3$ independent draws of the training set. Tables~\ref{tab:results_ffhq_noisy_std}--\ref{tab:results_dpde_std} give the corresponding standard deviations. The baselines have no repeated draws and are omitted.

\begin{table}[!ht]
\centering
\begin{adjustbox}{max width=\linewidth}
\begin{tabular}{l cc cc cc cc cc}
\hline
\vspace{1mm}
\multirow{2}{*}{NFEs} & \multicolumn{2}{c}{SR (x4)} & \multicolumn{2}{c}{Inpaint (box)} & \multicolumn{2}{c}{Inpaint (random)} & \multicolumn{2}{c}{Deblur (gauss)} & \multicolumn{2}{c}{Deblur (motion)} \\ \vspace{1mm}
 &LPIPS &FID &LPIPS &FID &LPIPS &FID &LPIPS &FID &LPIPS &FID \\ \hline
50 & 0.0002 & 0.01 & 0.0034 & 0.22 & 0.0007 & 0.14 & 0.0001 & 0.16 & 0.0012 & 0.44 \\
100 & 0.0004 & 0.07 & 0.0047 & 0.33 & 0.0016 & 0.11 & 0.0039 & 0.15 & 0.0013 & 0.41 \\ \hline
\end{tabular}
\end{adjustbox}
\vspace{2mm}
\caption{Standard deviation across 3 draws for the LEEGS rows of Table \ref{tab:results_ffhq_noisy} (FFHQ).}
\label{tab:results_ffhq_noisy_std}
\end{table}

\begin{table}[!ht]
\centering
\begin{adjustbox}{max width=\linewidth}
\begin{tabular}{l cc cc cc cc cc}
\hline
\vspace{1mm}
\multirow{2}{*}{NFEs} & \multicolumn{2}{c}{SR (x4)} & \multicolumn{2}{c}{Inpaint (box)} & \multicolumn{2}{c}{Inpaint (random)} & \multicolumn{2}{c}{Deblur (gauss)} & \multicolumn{2}{c}{Deblur (motion)} \\ \vspace{1mm}
 &LPIPS &FID &LPIPS &FID &LPIPS &FID &LPIPS &FID &LPIPS &FID \\ \hline
50 & 0.0001 & 0.23 & 0.0026 & 1.45 & 0.0056 & 1.07 & 0.0035 & 1.23 & 0.0048 & 2.41 \\
100 & 0.0013 & 0.29 & 0.0092 & 4.29 & 0.0006 & 0.43 & 0.0039 & 0.88 & 0.0151 & 3.67 \\ \hline
\end{tabular}
\end{adjustbox}
\vspace{2mm}
\caption{Standard deviation across 3 draws for the LEEGS rows of Table \ref{tab:results_imagenet_noisy} (ImageNet).}
\label{tab:results_imagenet_noisy_std}
\end{table}

\begin{table}[!ht]
    \small
\centering
\begin{tabular}{l l cccc c}
\hline
\vspace{1mm}
Dataset & NFEs & Wide & Narrow & S. res. (2x) & Alternate Lines & Half \\ \hline
Imagenet & 200 & 0.0019 & 0.0040 & 0.0016 & 0.0018 & 0.0041 \\
CelebA-HQ & 200 & 0.0010 & 0.0001 & 0.0001 & 0.0005 & 0.0016 \\ \hline
\end{tabular}
\vspace{2mm}
\caption{Standard deviation across 3 draws for the LEEGS rows of Table \ref{tab:results_inpainting_celeba} (image inpainting).}
\label{tab:results_inpainting_celeba_std}
\end{table}

\begin{table}[!ht]
    \small
\centering
\begin{tabular}{l cc}
\hline
\vspace{1mm}
NFEs & FaceID Score & FID \\ \hline
50 & 0.0036 & 0.90 \\
100 & 0.0119 & 0.85 \\ \hline
\end{tabular}
\vspace{2mm}
\caption{Standard deviation across 3 draws for the LEEGS rows of Table \ref{tab:results_faceid_celeba} (Face-ID guidance).}
\label{tab:results_faceid_celeba_std}
\end{table}

\begin{table}[!ht]
    \small
\centering
\begin{tabular}{l cc cc cc}
\hline
\vspace{1mm}
\multirow{2}{*}{NFEs} & \multicolumn{2}{c}{Poisson} & \multicolumn{2}{c}{Helmholtz} & \multicolumn{2}{c}{Darcy} \\ \vspace{1mm}
 & Forward & Inverse & Forward & Inverse & Forward & Inverse \\ \hline
50x1 & 0.01\% & 0.06\% & 0.08\% & 0.25\% & 4.27\% & 1.54\% \\
100x1 & 0.01\% & 0.19\% & 0.06\% & 0.18\% & 1.45\% & 5.28\% \\ \hline
\end{tabular}
\vspace{2mm}
\caption{Standard deviation across 3 draws for the LEEGS rows of Table \ref{tab:results_dpde} (PDE problems).}
\label{tab:results_dpde_std}
\end{table}
\FloatBarrier

\subsection{Noisy inverse problems}
\label{app:additional_metrics_noisy_inverse}

For every row of Tables~\ref{tab:results_ffhq_noisy} and~\ref{tab:results_imagenet_noisy}, we report the variance of LPIPS across the test samples (Tables~\ref{tab:results_ffhq_noisy_lpipsvar} and~\ref{tab:results_imagenet_noisy_lpipsvar}), PSNR (Tables~\ref{tab:results_ffhq_noisy_psnr} and~\ref{tab:results_imagenet_noisy_psnr}) and SSIM (Tables~\ref{tab:results_ffhq_noisy_ssim} and~\ref{tab:results_imagenet_noisy_ssim}). PSNR (peak signal-to-noise ratio) measures the pixel-wise error between the reconstruction and the ground truth on a logarithmic scale, as $10 \log_{10}(\text{MAX}^2 / \text{MSE})$, where MAX is the maximum pixel value. SSIM (structural similarity index) compares local statistics (mean intensity, contrast and correlation) of the two images over sliding windows, and ranges up to $1$ for identical images. Higher is better for both. Unlike LPIPS, both are pixel-level metrics that tend to favor smooth reconstructions over sharp but slightly misaligned details. We compute both with \texttt{piq} on images normalized to $[0, 1]$, so that $\text{MAX} = 1$. The DPS reference row is omitted, since its results are taken from \citet{chung2023DPS}.

\begin{table}[H]
\centering
\begin{adjustbox}{max width=\linewidth}
\begin{tabular}{l l cc cc cc cc cc}
\hline
\vspace{1mm}
\multirow{2}{*}{} & \multirow{2}{*}{NFEs} & SR (x4) & Inpaint (box) & Inpaint (random) & Deblur (gauss) & Deblur (motion) \\ \hline
Const. & 50 & 0.0161 & 0.0166 & 0.0149 & 0.0185 & 0.0108 \\
DAPS & 50 & \best{0.0011} & 0.0014 & \scnd{0.0019} & \best{0.0011} & \best{0.0011} \\
DSG & 50 & 0.0021 & 0.0053 & 0.0037 & 0.0025 & 0.0018 \\
CDIM & 50 & 0.0018 & \best{0.0011} & 0.0024 & 0.0026 & 0.0037 \\
Linear increase & 50 & \scnd{0.0013} & 0.0015 & \best{0.0016} & 0.0021 & 0.0016 \\
Linear decrease & 50 & 0.0153 & 0.0146 & 0.0132 & 0.0152 & 0.0150 \\
\textbf{LEEGS (ours)} & 50 & 0.0014 & \scnd{0.0012} & \best{0.0016} & \scnd{0.0018} & \scnd{0.0014} \\
\hline
Const. & 100 & 0.0039 & 0.0078 & 0.0056 & 0.0064 & 0.0055 \\
DAPS & 100 & \best{0.0010} & 0.0014 & \scnd{0.0017} & \best{0.0011} & \best{0.0010} \\
DSG & 100 & 0.0022 & 0.0031 & 0.0020 & 0.0024 & 0.0033 \\
CDIM & 100 & 0.0016 & 0.0011 & 0.0028 & 0.0023 & 0.0034 \\
Linear increase & 100 & 0.0014 & \scnd{0.0009} & \best{0.0012} & 0.0020 & \scnd{0.0016} \\
Linear decrease & 100 & 0.0113 & 0.0165 & 0.0095 & 0.0082 & 0.0079 \\
\textbf{LEEGS (ours)} & 100 & \scnd{0.0013} & \best{0.0006} & \best{0.0012} & \scnd{0.0016} & \best{0.0010} \\
\hline
\end{tabular}
\end{adjustbox}
\vspace{2mm}
\caption{Variance of LPIPS across test samples ($\downarrow$) for the noisy inverse problems (FFHQ). \textbf{Green} (bold): best; \textbf{blue}: second best.}
\label{tab:results_ffhq_noisy_lpipsvar}
\end{table}

\begin{table}[H]
\centering
\begin{adjustbox}{max width=\linewidth}
\begin{tabular}{l l cc cc cc cc cc}
\hline
\vspace{1mm}
\multirow{2}{*}{} & \multirow{2}{*}{NFEs} & SR (x4) & Inpaint (box) & Inpaint (random) & Deblur (gauss) & Deblur (motion) \\ \hline
Const. & 50 & 0.0084 & 0.0220 & 0.0129 & 0.0121 & 0.0103 \\
DAPS & 50 & \best{0.0038} & \scnd{0.0032} & \best{0.0021} & \best{0.0044} & \best{0.0030} \\
DSG & 50 & 0.0147 & 0.0106 & 0.0144 & 0.0102 & \scnd{0.0056} \\
CDIM & 50 & 0.0099 & \best{0.0021} & \scnd{0.0062} & 0.0090 & 0.0098 \\
Linear increase & 50 & 0.0114 & 0.0080 & 0.0110 & 0.0110 & 0.0126 \\
Linear decrease & 50 & 0.0119 & 0.0108 & 0.0129 & 0.0098 & 0.0131 \\
\textbf{LEEGS (ours)} & 50 & \scnd{0.0060} & 0.0060 & 0.0077 & \scnd{0.0089} & 0.0090 \\
\hline
Const. & 100 & 0.0182 & 0.0081 & 0.0186 & 0.0143 & 0.0097 \\
DAPS & 100 & \best{0.0041} & 0.0031 & \best{0.0041} & \best{0.0049} & \scnd{0.0033} \\
DSG & 100 & 0.0069 & 0.0091 & 0.0078 & 0.0098 & \best{0.0024} \\
CDIM & 100 & 0.0097 & \best{0.0018} & 0.0059 & 0.0093 & 0.0098 \\
Linear increase & 100 & 0.0086 & 0.0059 & 0.0055 & 0.0100 & 0.0127 \\
Linear decrease & 100 & 0.0172 & 0.0190 & 0.0158 & 0.0143 & 0.0146 \\
\textbf{LEEGS (ours)} & 100 & \scnd{0.0059} & \scnd{0.0027} & \scnd{0.0046} & \scnd{0.0082} & 0.0092 \\
\hline
\end{tabular}
\end{adjustbox}
\vspace{2mm}
\caption{Variance of LPIPS across test samples ($\downarrow$) for the noisy inverse problems (ImageNet). \textbf{Green} (bold): best; \textbf{blue}: second best.}
\label{tab:results_imagenet_noisy_lpipsvar}
\end{table}

\begin{table}[H]
\centering
\begin{adjustbox}{max width=\linewidth}
\begin{tabular}{l l cc cc cc cc cc}
\hline
\vspace{1mm}
\multirow{2}{*}{} & \multirow{2}{*}{NFEs} & SR (x4) & Inpaint (box) & Inpaint (random) & Deblur (gauss) & Deblur (motion) \\ \hline
Const. & 50 & 19.62 & 16.25 & 19.53 & 18.59 & 17.15 \\
DAPS & 50 & \scnd{28.04} & \best{24.24} & 20.65 & \best{28.20} & \best{29.85} \\
DSG & 50 & 21.83 & 15.89 & 18.78 & 20.03 & 10.03 \\
CDIM & 50 & \best{28.32} & 22.47 & \scnd{26.50} & \scnd{26.79} & 25.91 \\
Linear increase & 50 & 27.40 & 22.27 & \best{26.86} & 25.66 & 27.43 \\
Linear decrease & 50 & 15.35 & 14.50 & 15.88 & 15.04 & 15.19 \\
\textbf{LEEGS (ours)} & 50 & 27.47 & \scnd{22.59} & 26.37 & 26.35 & \scnd{27.94} \\
\hline
Const. & 100 & 26.26 & 19.13 & 24.43 & 23.60 & 19.41 \\
DAPS & 100 & \best{28.58} & \best{24.66} & 23.63 & \best{28.65} & \best{30.58} \\
DSG & 100 & 21.19 & 16.38 & 21.78 & 19.99 & 12.67 \\
CDIM & 100 & \scnd{28.24} & 22.43 & 26.21 & \scnd{26.88} & 26.22 \\
Linear increase & 100 & 27.45 & 22.80 & \best{27.46} & 25.78 & 27.76 \\
Linear decrease & 100 & 20.11 & 16.55 & 19.94 & 19.51 & 19.75 \\
\textbf{LEEGS (ours)} & 100 & 27.43 & \scnd{23.62} & \scnd{27.17} & 26.54 & \scnd{28.76} \\
\hline
\end{tabular}
\end{adjustbox}
\vspace{2mm}
\caption{PSNR in dB ($\uparrow$) for the noisy inverse problems (FFHQ). \textbf{Green} (bold): best; \textbf{blue}: second best.}
\label{tab:results_ffhq_noisy_psnr}
\end{table}

\begin{table}[H]
\centering
\begin{adjustbox}{max width=\linewidth}
\begin{tabular}{l l cc cc cc cc cc}
\hline
\vspace{1mm}
\multirow{2}{*}{} & \multirow{2}{*}{NFEs} & SR (x4) & Inpaint (box) & Inpaint (random) & Deblur (gauss) & Deblur (motion) \\ \hline
Const. & 50 & 15.08 & 14.29 & 14.79 & 14.71 & 13.74 \\
DAPS & 50 & \scnd{24.40} & \scnd{18.35} & 17.99 & \best{24.44} & \best{27.06} \\
DSG & 50 & 18.85 & 13.27 & 15.77 & 20.18 & 11.64 \\
CDIM & 50 & 23.45 & \best{18.64} & \scnd{22.66} & 22.37 & 22.57 \\
Linear increase & 50 & 23.65 & 16.09 & 22.25 & 21.16 & 22.36 \\
Linear decrease & 50 & 12.91 & 11.64 & 12.56 & 13.12 & 12.46 \\
\textbf{LEEGS (ours)} & 50 & \best{24.45} & 17.14 & \best{23.62} & \scnd{22.44} & \scnd{24.47} \\
\hline
Const. & 100 & 20.41 & 15.29 & 19.56 & 18.45 & 16.13 \\
DAPS & 100 & \best{24.71} & \best{18.86} & 20.27 & \best{24.66} & \best{27.54} \\
DSG & 100 & 21.42 & 10.62 & 20.47 & 19.60 & 8.68 \\
CDIM & 100 & 23.66 & \scnd{18.59} & 22.43 & 22.66 & 23.01 \\
Linear increase & 100 & 24.13 & 16.28 & \scnd{24.26} & 21.72 & 23.00 \\
Linear decrease & 100 & 15.81 & 12.51 & 15.80 & 15.86 & 16.04 \\
\textbf{LEEGS (ours)} & 100 & \scnd{24.37} & 18.20 & \best{24.84} & \scnd{22.68} & \scnd{25.27} \\
\hline
\end{tabular}
\end{adjustbox}
\vspace{2mm}
\caption{PSNR in dB ($\uparrow$) for the noisy inverse problems (ImageNet). \textbf{Green} (bold): best; \textbf{blue}: second best.}
\label{tab:results_imagenet_noisy_psnr}
\end{table}

\begin{table}[H]
\centering
\begin{adjustbox}{max width=\linewidth}
\begin{tabular}{l l cc cc cc cc cc}
\hline
\vspace{1mm}
\multirow{2}{*}{} & \multirow{2}{*}{NFEs} & SR (x4) & Inpaint (box) & Inpaint (random) & Deblur (gauss) & Deblur (motion) \\ \hline
Const. & 50 & 0.614 & 0.561 & 0.621 & 0.568 & 0.512 \\
DAPS & 50 & 0.710 & 0.724 & 0.509 & 0.715 & 0.758 \\
DSG & 50 & 0.739 & 0.658 & 0.696 & 0.675 & 0.333 \\
CDIM & 50 & 0.798 & \best{0.839} & \scnd{0.799} & 0.733 & 0.731 \\
Linear increase & 50 & \scnd{0.808} & 0.829 & \best{0.844} & \scnd{0.758} & \scnd{0.820} \\
Linear decrease & 50 & 0.470 & 0.444 & 0.488 & 0.457 & 0.465 \\
\textbf{LEEGS (ours)} & 50 & \best{0.812} & \scnd{0.832} & \best{0.844} & \best{0.781} & \best{0.833} \\
\hline
Const. & 100 & 0.787 & 0.687 & 0.757 & 0.692 & 0.559 \\
DAPS & 100 & 0.753 & 0.738 & 0.633 & 0.752 & 0.804 \\
DSG & 100 & 0.727 & 0.695 & 0.745 & 0.685 & 0.449 \\
CDIM & 100 & 0.800 & \scnd{0.839} & 0.790 & 0.744 & 0.745 \\
Linear increase & 100 & \best{0.809} & 0.835 & \best{0.844} & \scnd{0.760} & \scnd{0.827} \\
Linear decrease & 100 & 0.600 & 0.537 & 0.599 & 0.567 & 0.579 \\
\textbf{LEEGS (ours)} & 100 & \scnd{0.806} & \best{0.840} & \scnd{0.838} & \best{0.784} & \best{0.849} \\
\hline
\end{tabular}
\end{adjustbox}
\vspace{2mm}
\caption{SSIM ($\uparrow$) for the noisy inverse problems (FFHQ). \textbf{Green} (bold): best; \textbf{blue}: second best.}
\label{tab:results_ffhq_noisy_ssim}
\end{table}

\begin{table}[H]
\centering
\begin{adjustbox}{max width=\linewidth}
\begin{tabular}{l l cc cc cc cc cc}
\hline
\vspace{1mm}
\multirow{2}{*}{} & \multirow{2}{*}{NFEs} & SR (x4) & Inpaint (box) & Inpaint (random) & Deblur (gauss) & Deblur (motion) \\ \hline
Const. & 50 & 0.341 & 0.567 & 0.385 & 0.354 & 0.324 \\
DAPS & 50 & 0.601 & 0.657 & 0.351 & \best{0.601} & \best{0.715} \\
DSG & 50 & 0.554 & 0.495 & 0.516 & 0.534 & 0.303 \\
CDIM & 50 & 0.625 & \best{0.771} & 0.661 & 0.575 & 0.608 \\
Linear increase & 50 & \scnd{0.650} & 0.653 & \scnd{0.678} & 0.538 & 0.598 \\
Linear decrease & 50 & 0.316 & 0.295 & 0.311 & 0.304 & 0.300 \\
\textbf{LEEGS (ours)} & 50 & \best{0.683} & \scnd{0.698} & \best{0.728} & \scnd{0.598} & \scnd{0.695} \\
\hline
Const. & 100 & 0.535 & 0.686 & 0.544 & 0.445 & 0.357 \\
DAPS & 100 & 0.635 & 0.669 & 0.465 & \best{0.630} & \best{0.752} \\
DSG & 100 & 0.628 & 0.470 & 0.613 & 0.547 & 0.207 \\
CDIM & 100 & 0.632 & \best{0.772} & 0.650 & 0.586 & 0.627 \\
Linear increase & 100 & \scnd{0.674} & 0.691 & \scnd{0.737} & 0.561 & 0.628 \\
Linear decrease & 100 & 0.391 & 0.368 & 0.399 & 0.373 & 0.385 \\
\textbf{LEEGS (ours)} & 100 & \best{0.681} & \scnd{0.744} & \best{0.745} & \scnd{0.610} & \scnd{0.729} \\
\hline
\end{tabular}
\end{adjustbox}
\vspace{2mm}
\caption{SSIM ($\uparrow$) for the noisy inverse problems (ImageNet). \textbf{Green} (bold): best; \textbf{blue}: second best.}
\label{tab:results_imagenet_noisy_ssim}
\end{table}
\FloatBarrier

\subsection{Image inpainting problems}
\label{app:additional_metrics_inpainting}

For the $200$-NFE rows of Table~\ref{tab:results_inpainting_celeba}, we report the variance of LPIPS across the test samples (Table~\ref{tab:results_inpainting_celeba_lpipsvar}), PSNR (Table~\ref{tab:results_inpainting_celeba_psnr}) and SSIM (Table~\ref{tab:results_inpainting_celeba_ssim}), computed as for the noisy inverse problems. The RePaint reference row at $\approx 2400$ NFEs is omitted, since its results are taken from \citet{lugmayr2022repaintinpaintingusingdenoising}.

\begin{table}[H]
    \small
\centering
\begin{tabular}{l | l cc cc cc cc cc}
\hline
\vspace{1mm}
Dataset & Method & NFEs & Wide & Narrow & S. res. (2x) & Alternate Lines & Half \\ \hline
\multirow[c]{2}{*}{Imagenet} & \textbf{LEEGS (ours)} & 200 & \best{0.0058} & \best{0.0045} & 0.0029 & \best{0.0020} & 0.0030 \\
& RePaint & 200 & 0.0093 & 0.0075 & \best{0.0021} & 0.0079 & \best{0.0014} \\ \hline 
\multirow[c]{2}{*}{CelebA-HQ} & \textbf{LEEGS (ours)} & 200 & \best{0.0022} & \best{0.0011} & \best{0.0003} & \best{0.0002} & \best{0.0013} \\
& RePaint & 200 & 0.0053 & 0.0045 & 0.0023 & \best{0.0002} & 0.0017 \\ \hline 
\end{tabular}
\vspace{2mm}
\caption{Variance of LPIPS across test samples ($\downarrow$) for image inpainting. \textbf{Green} (bold): best.}
\label{tab:results_inpainting_celeba_lpipsvar}
\end{table}

\begin{table}[H]
    \small
\centering
\begin{tabular}{l | l cc cc cc cc cc}
\hline
\vspace{1mm}
Dataset & Method & NFEs & Wide & Narrow & S. res. (2x) & Alternate Lines & Half \\ \hline
\multirow[c]{2}{*}{Imagenet} & \textbf{LEEGS (ours)} & 200 & \best{20.28} & \best{24.20} & \best{27.76} & \best{29.74} & \best{13.71} \\
& RePaint & 200 & 18.88 & 21.69 & 16.81 & 21.65 & 13.47 \\ \hline 
\multirow[c]{2}{*}{CelebA-HQ} & \textbf{LEEGS (ours)} & 200 & \best{24.99} & \best{29.00} & \best{32.76} & 35.00 & \best{16.41} \\
& RePaint & 200 & 21.40 & 24.73 & 29.24 & \best{36.46} & 13.88 \\ \hline 
\end{tabular}
\vspace{2mm}
\caption{PSNR in dB ($\uparrow$) for image inpainting. \textbf{Green} (bold): best.}
\label{tab:results_inpainting_celeba_psnr}
\end{table}

\begin{table}[H]
    \small
\centering
\begin{tabular}{l | l cc cc cc cc cc}
\hline
\vspace{1mm}
Dataset & Method & NFEs & Wide & Narrow & S. res. (2x) & Alternate Lines & Half \\ \hline
\multirow[c]{2}{*}{Imagenet} & \textbf{LEEGS (ours)} & 200 & 0.777 & \best{0.808} & \best{0.834} & \best{0.880} & \best{0.594} \\
& RePaint & 200 & \best{0.784} & 0.794 & 0.295 & 0.635 & 0.584 \\ \hline 
\multirow[c]{2}{*}{CelebA-HQ} & \textbf{LEEGS (ours)} & 200 & \best{0.879} & \best{0.917} & \best{0.941} & 0.964 & \best{0.720} \\
& RePaint & 200 & 0.847 & 0.868 & 0.845 & \best{0.966} & 0.670 \\ \hline 
\end{tabular}
\vspace{2mm}
\caption{SSIM ($\uparrow$) for image inpainting. \textbf{Green} (bold): best.}
\label{tab:results_inpainting_celeba_ssim}
\end{table}

\clearpage
\section{Additional learned guidance schedules}
\label{app:additional_schedules}

This section gives the guidance schedules learned by LEEGS for the main experiments, grouped by task. All schedules are plotted against normalized diffusion time ($0$: noisiest, $1$: clean), so that schedules trained at different NFE budgets are directly comparable. The training configurations are given in Appendix~\ref{app:exp_details}.

\subsection{Noisy inverse problems}
\label{app:schedules_noisy_inverse}

The schedules for FFHQ are given in Figure~\ref{fig:crosstask_ffhq} (Appendix~\ref{app:cross_task}).

\begin{figure}[ht]
    \centering
    \includegraphics[width=0.9\textwidth]{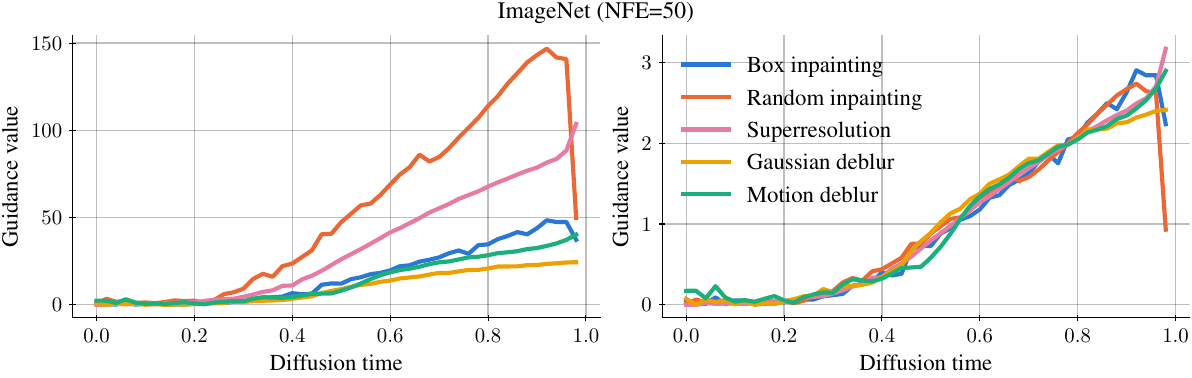}\\[1ex]
    \includegraphics[width=0.9\textwidth]{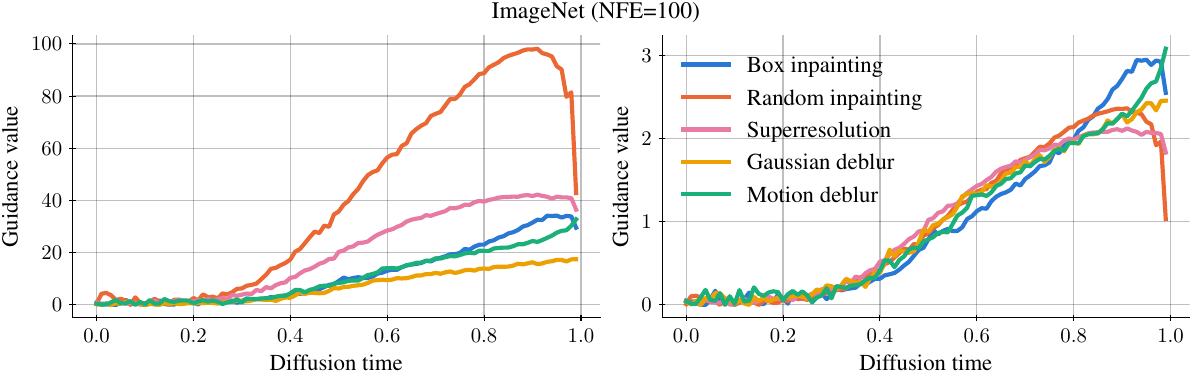}
    \caption{ImageNet. \emph{Left:} raw. \emph{Right:} mean-normalized. \emph{Top:} $50$ NFEs. \emph{Bottom:} $100$ NFEs.}
    \label{fig:crosstask_imagenet}
\end{figure}
\clearpage

\subsection{Image inpainting}
\label{app:schedules_inpainting}

\begin{figure}[ht]
    \centering
    \includegraphics[width=0.9\textwidth]{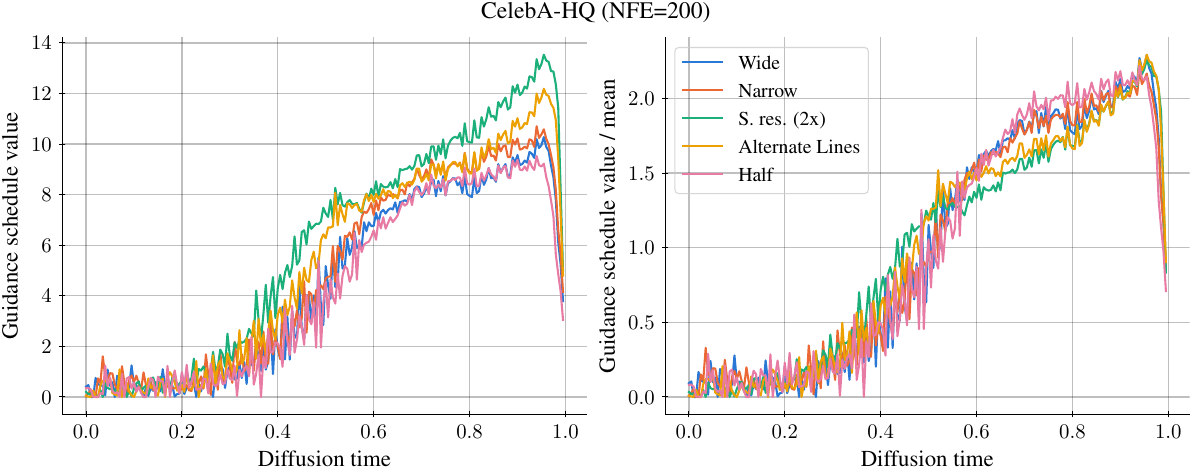}\\[1ex]
    \includegraphics[width=0.9\textwidth]{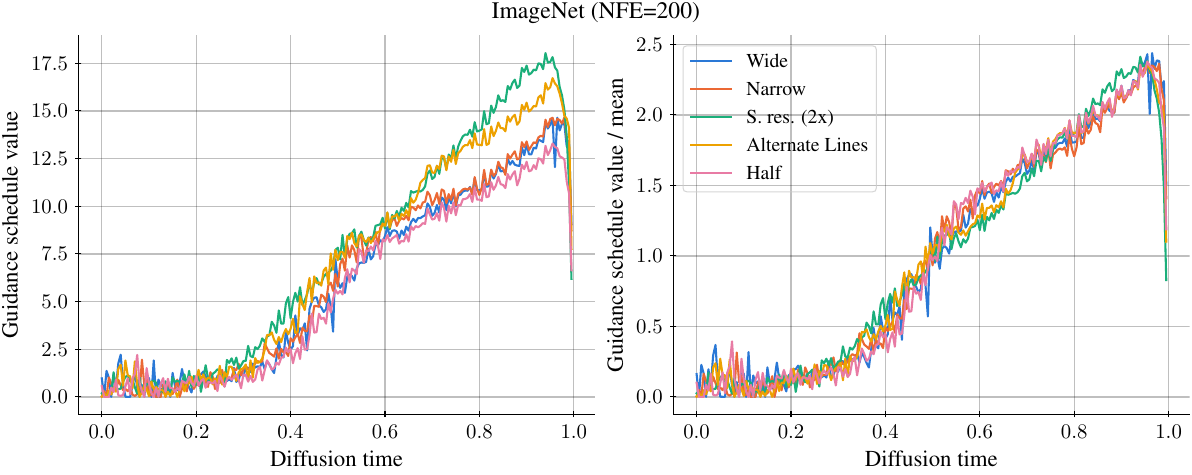}
    \caption{Image inpainting, $200$ NFEs. \emph{Left:} raw. \emph{Right:} mean-normalized. \emph{Top:} CelebA-HQ. \emph{Bottom:} ImageNet.}
    \label{fig:schedules_inpainting}
\end{figure}
\clearpage

\subsection{Face-ID guidance}
\label{app:schedules_faceid}

\begin{figure}[ht]
    \centering
    \includegraphics[width=0.9\textwidth]{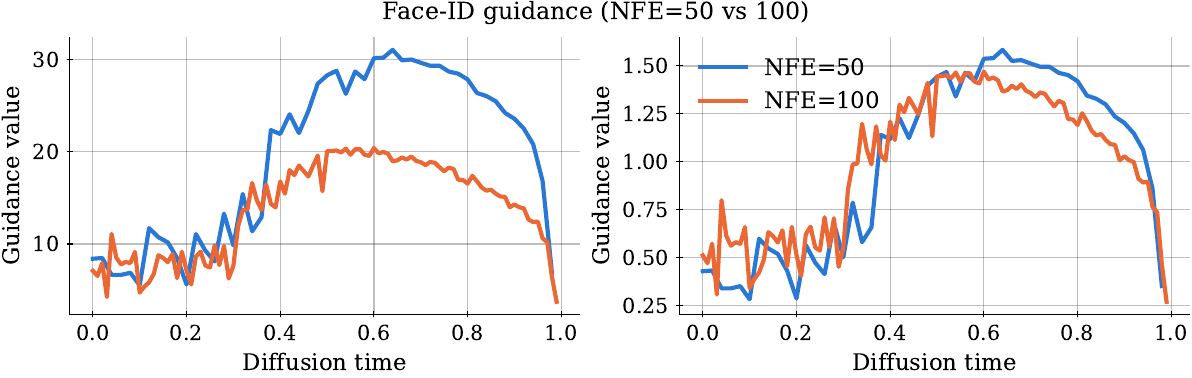}
    \caption{Face-ID guidance. \emph{Left:} raw. \emph{Right:} mean-normalized. Both NFE budgets overlaid.}
    \label{fig:crosstask_faceid}
\end{figure}
\clearpage

\subsection{PDE problems}
\label{app:schedules_pde}

\begin{figure}[ht]
    \centering
    \includegraphics[width=0.9\textwidth]{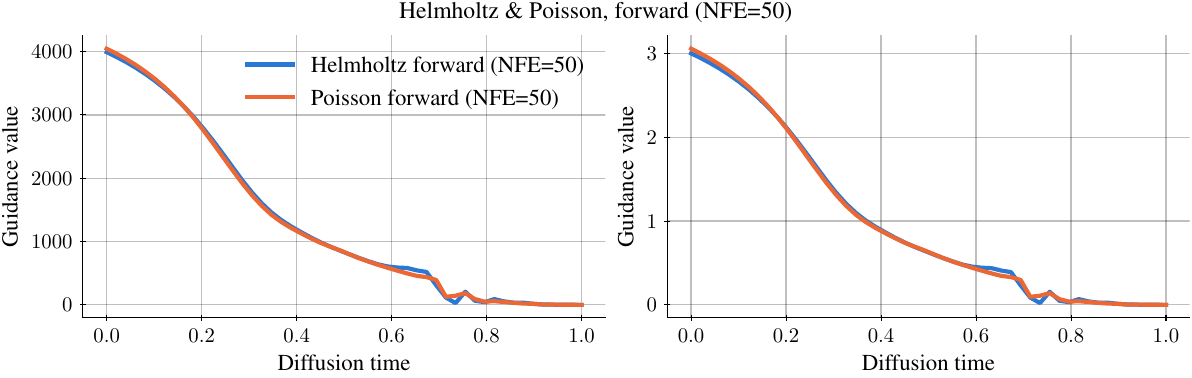}\\[1ex]
    \includegraphics[width=0.9\textwidth]{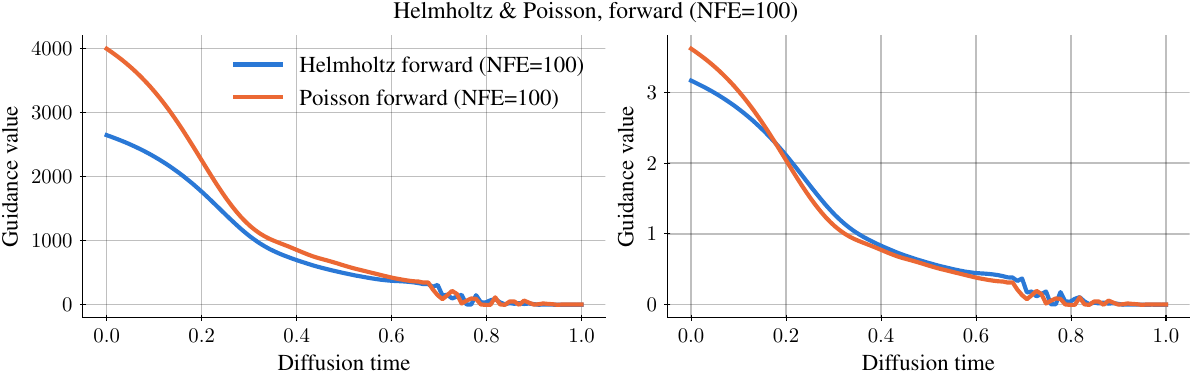}
    \caption{Helmholtz and Poisson, forward. \emph{Left:} raw. \emph{Right:}    mean-normalized. \emph{Top:} $50$ NFEs. \emph{Bottom:} $100$ NFEs.}
    \label{fig:crosstask_pde_forward}
\end{figure}

\begin{figure}[ht]
    \centering
    \includegraphics[width=0.9\textwidth]{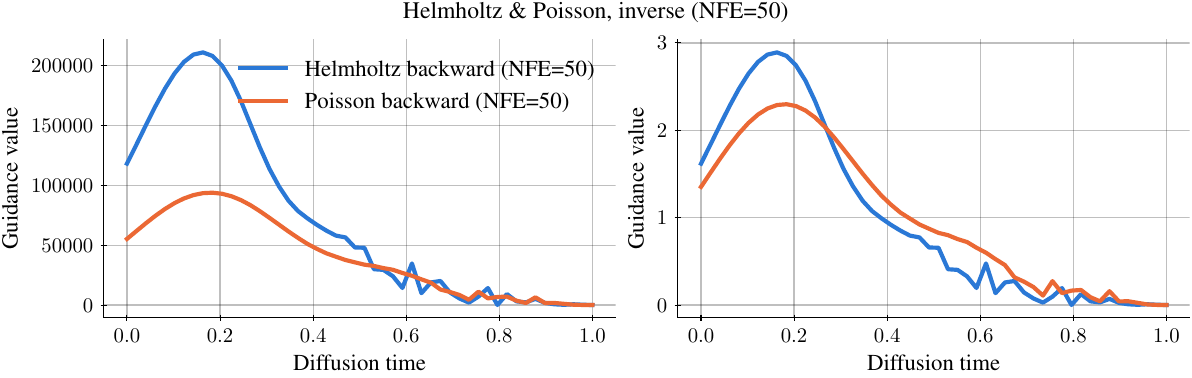}\\[1ex]
    \includegraphics[width=0.9\textwidth]{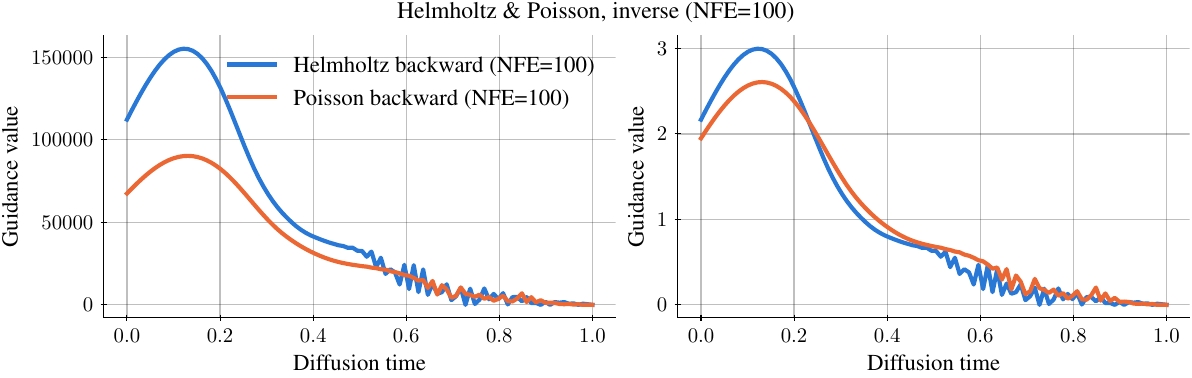}
    \caption{Helmholtz and Poisson, inverse. \emph{Left:} raw. \emph{Right:} mean-normalized. \emph{Top:} $50$ NFEs. \emph{Bottom:} $100$ NFEs.}
    \label{fig:crosstask_pde_inverse}
\end{figure}

\begin{figure}[ht]
    \centering
    \includegraphics[width=0.9\textwidth]{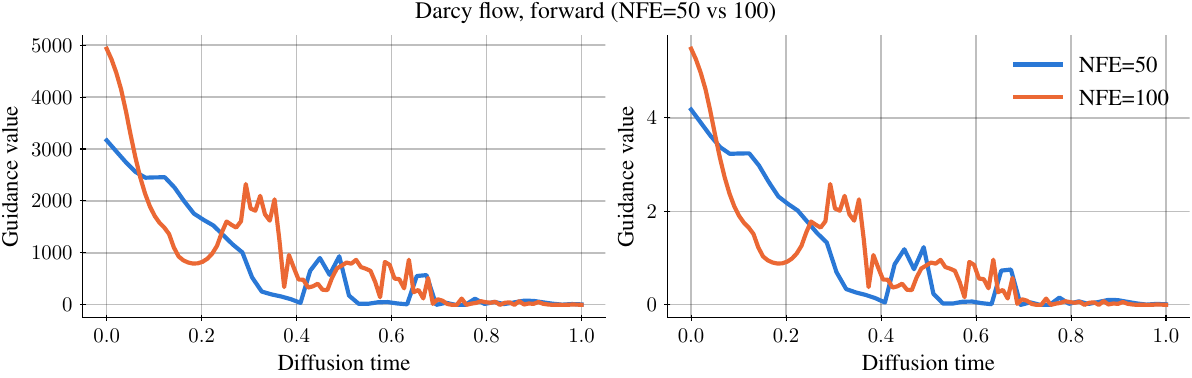}\\[1ex]
    \includegraphics[width=0.9\textwidth]{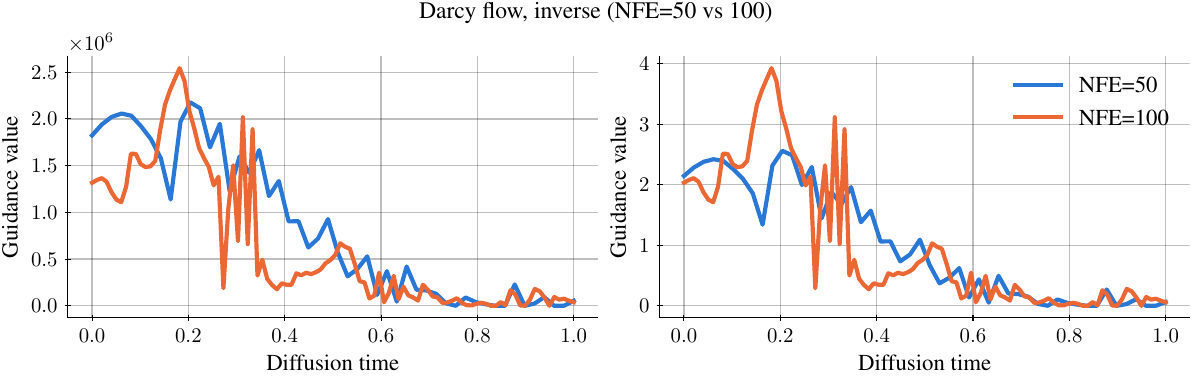}
    \caption{Darcy flow, $50$ and $100$ NFEs; raw and normalized within each panel. \emph{Top:} forward. \emph{Bottom:} inverse.}
    \label{fig:crosstask_darcy}
\end{figure}

\clearpage
\section{Hyperparameter searches for the baselines}
\label{app:additional_hyperparameter_searches}

This section gives the hyperparameter-search results for the baselines, grouped by task. The search procedures are described in Appendix~\ref{app:exp_details}.

\subsection{Noisy inverse problems}
\label{app:hyper_noisy_inverse}

The selection metric is LPIPS.

\paragraph{FFHQ, $10$ NFEs.} An additional search at $\mathrm{NFEs}=10$, FFHQ only, for the constant guidance, linear increase, DSG, and DAPS baselines, following the same procedures as above.

\begin{figure}[ht]
    \centering
    \includegraphics[width=0.9\textwidth]{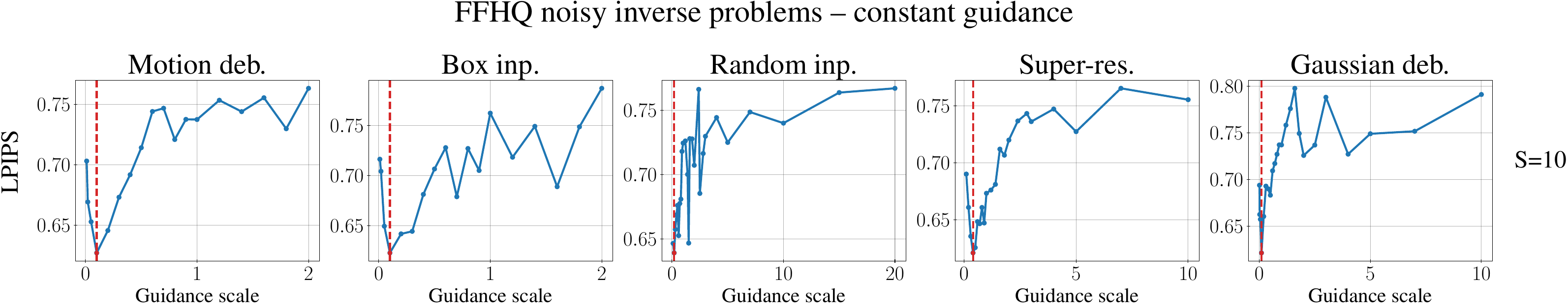}
    \caption{Constant guidance, FFHQ, $10$ NFEs: LPIPS vs.\ guidance scale; best value in red.}
    \label{fig:grid_search_dps_ffhq_s10}
\end{figure}

\begin{figure}[ht]
    \centering
    \includegraphics[width=0.9\textwidth]{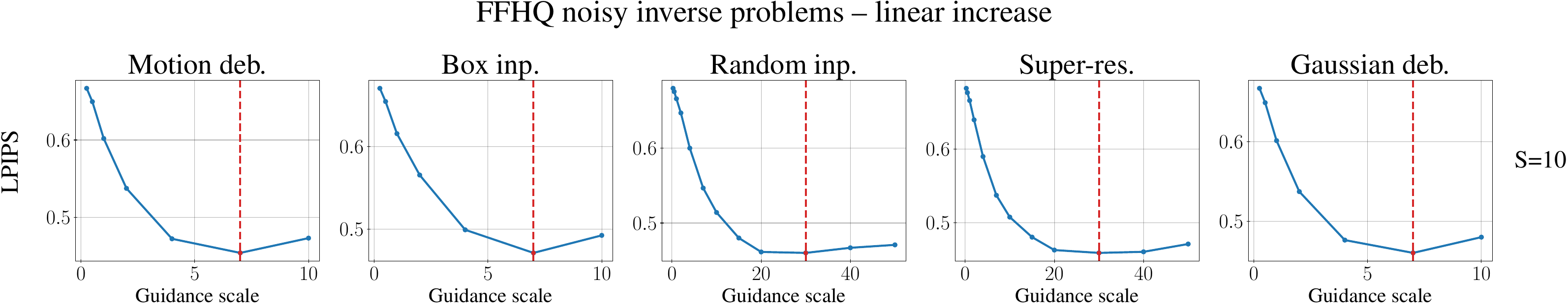}
    \caption{Linear increase, FFHQ, $10$ NFEs: LPIPS vs.\ guidance scale; best value in red.}
    \label{fig:grid_search_linear_increase_ffhq_s10}
\end{figure}

\begin{figure}[ht]
    \centering
    \includegraphics[width=0.9\textwidth]{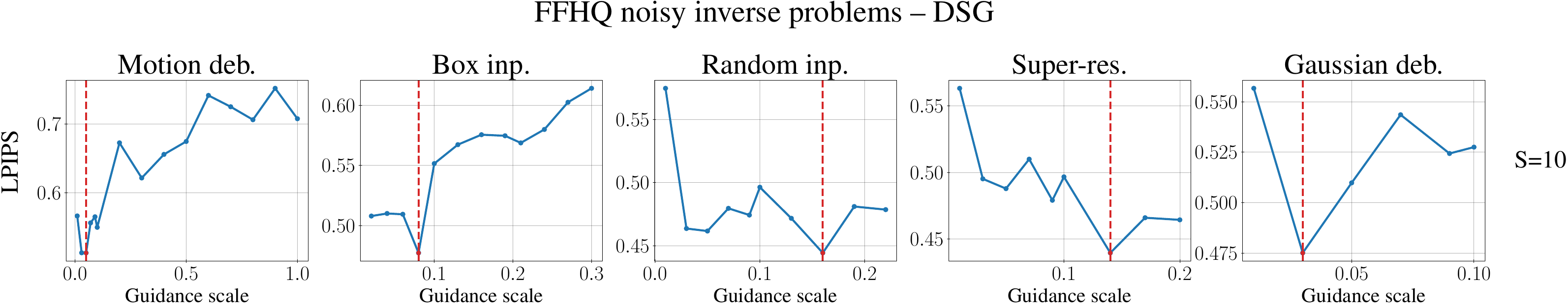}
    \caption{DSG, FFHQ, $10$ NFEs: LPIPS vs.\ guidance scale; best value in red.}
    \label{fig:grid_search_dsg_ffhq_s10}
\end{figure}

\begin{figure}[ht]
    \centering
    \includegraphics[width=0.9\textwidth]{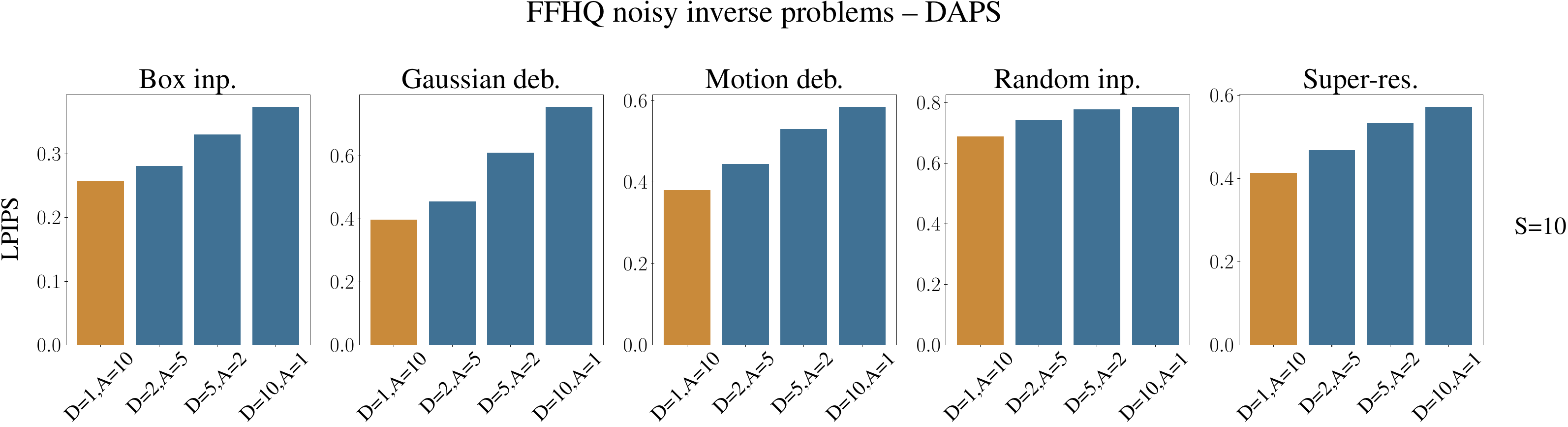}
    \caption{DAPS, FFHQ, $10$ NFEs: LPIPS per configuration $(D, A)$; best value in orange.}
    \label{fig:grid_search_daps_ffhq_s10}
\end{figure}
\clearpage

\subsection{Image inpainting}
\label{app:hyper_inpainting}

The selection metric is LPIPS.

\begin{figure}[ht]
    \centering
    \includegraphics[width=0.9\textwidth]{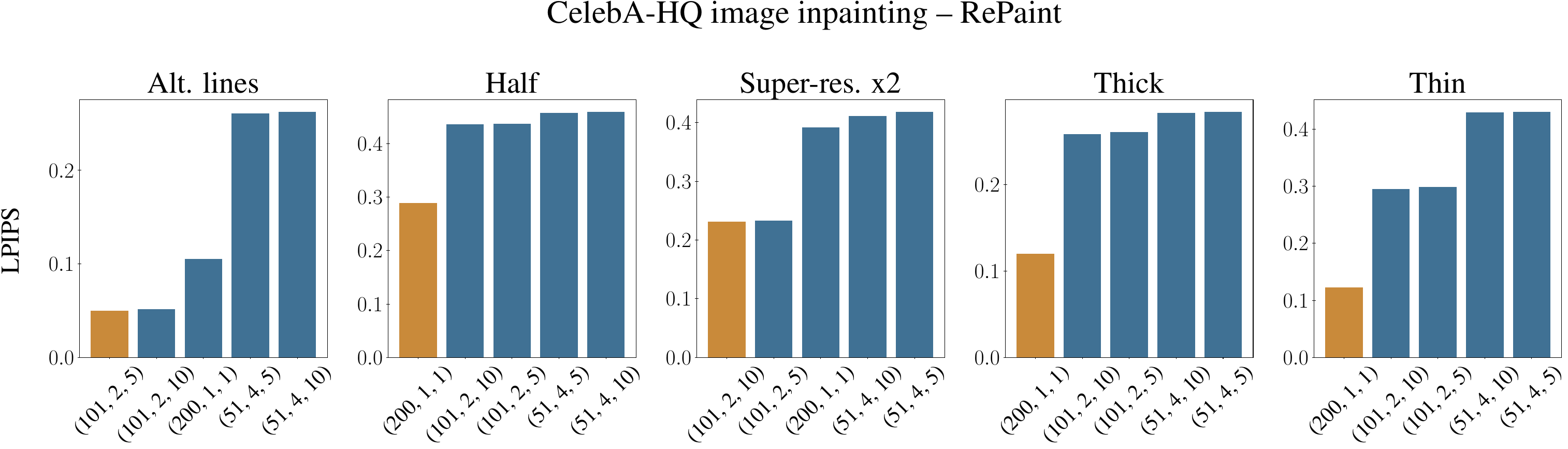}
    \caption{RePaint, CelebA-HQ: LPIPS per configuration $(S, N, J)$; best value in orange.}
    \label{fig:grid_search_celeba_inpaint}
\end{figure}

\begin{figure}[ht]
    \centering
    \includegraphics[width=0.9\textwidth]{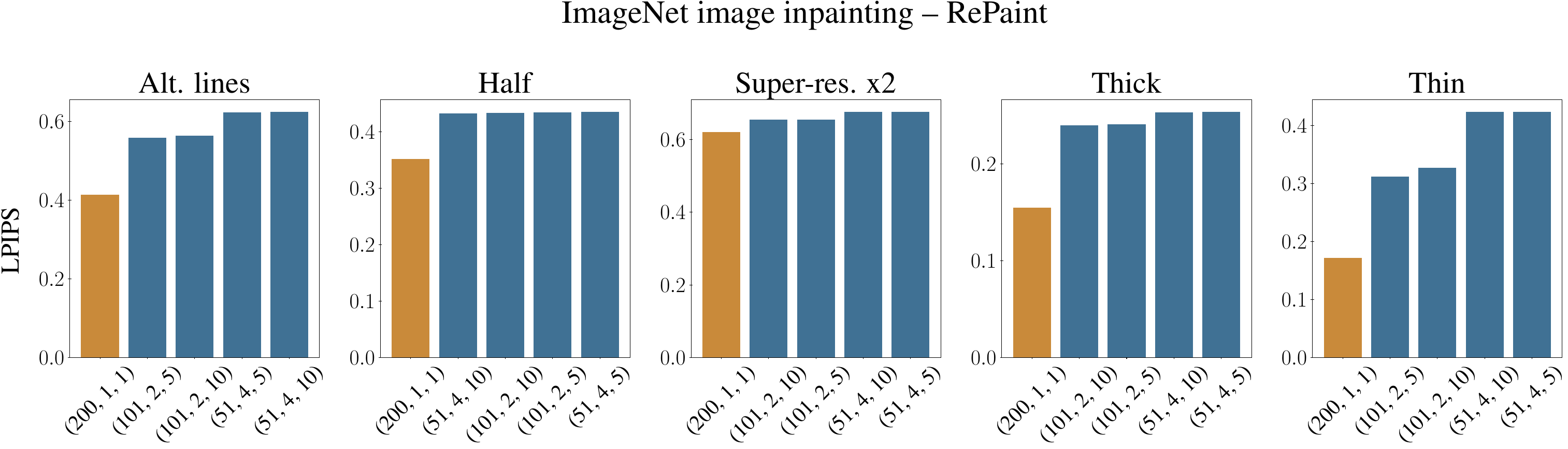}
    \caption{RePaint, ImageNet: LPIPS per configuration $(S, N, J)$; best value in orange.}
    \label{fig:grid_search_imagenet_inpaint}
\end{figure}
\clearpage

\subsection{Face-ID guidance}
\label{app:hyper_faceid}

The selection metric is the mean FaceID score.

\begin{figure}[ht]
    \centering
    \includegraphics[width=0.5\textwidth]{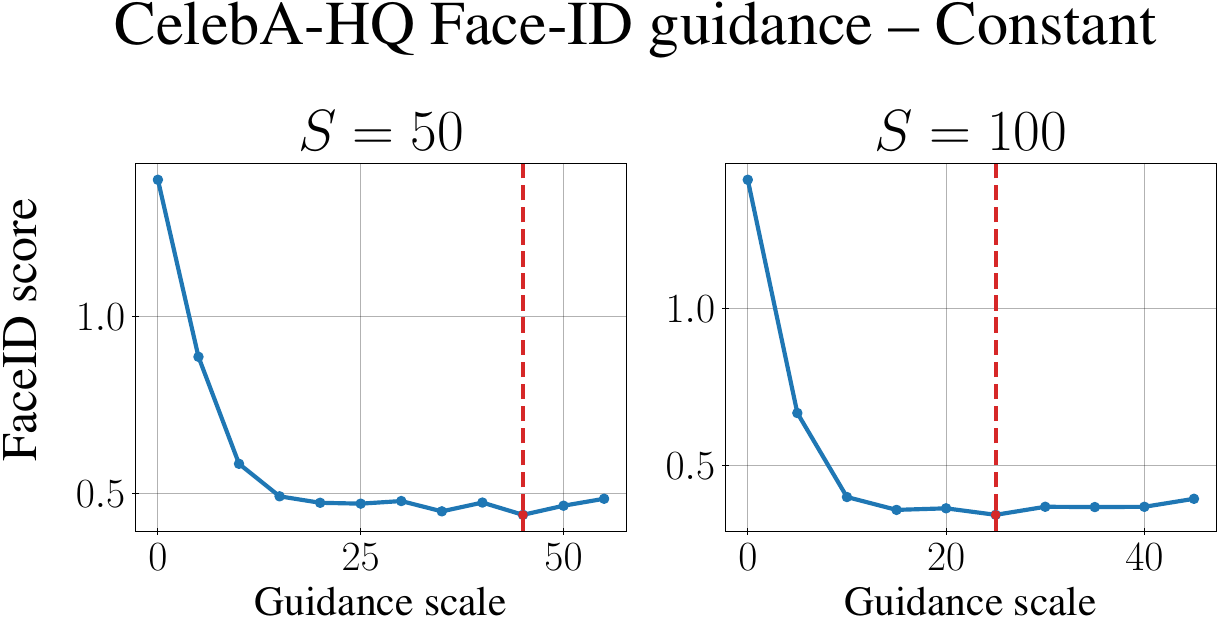}
    \caption{Constant guidance, Face-ID (CelebA-HQ): FaceID score vs.\ guidance scale; columns: $S=50,100$; best value in red.}
    \label{fig:grid_search_const_faceid}
\end{figure}

\begin{figure}[ht]
    \centering
    \includegraphics[width=0.5\textwidth]{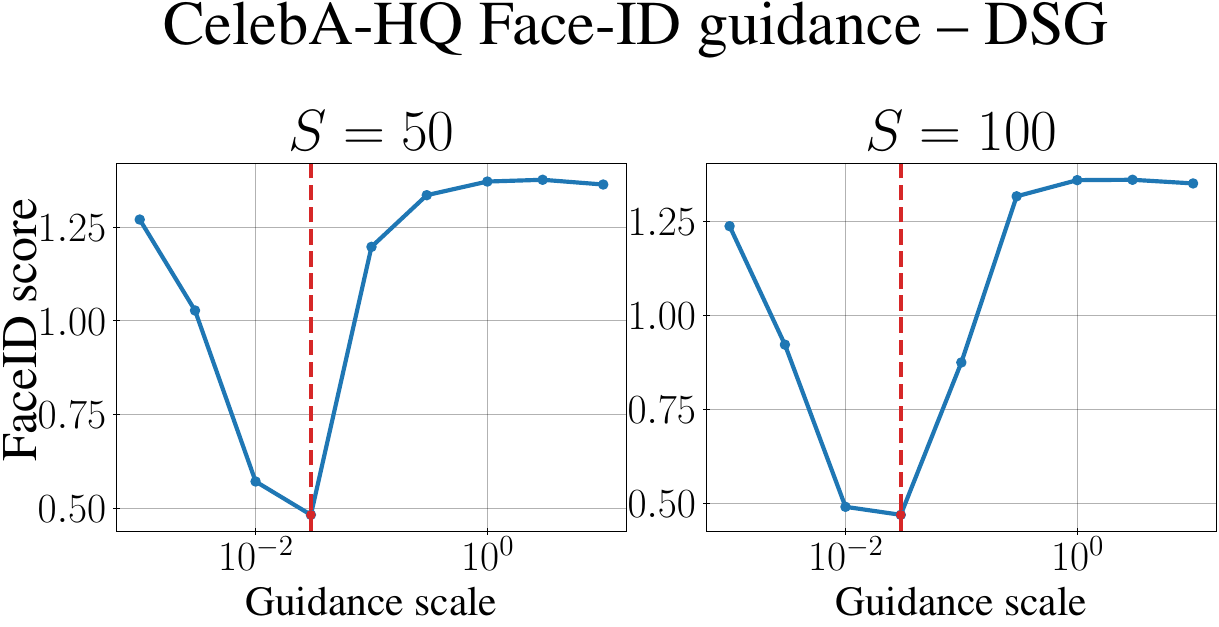}
    \caption{DSG, Face-ID (CelebA-HQ): FaceID score vs.\ guidance scale; columns: $S=50,100$; best value in red.}
    \label{fig:grid_search_dsg_faceid}
\end{figure}
\clearpage

\subsection{PDE problems}
\label{app:hyper_pde}

The selection metric is the relative error on $u$ for the forward problems, and the relative error on $a$ (Poisson, Helmholtz) or the error rate on $a$ (Darcy) for the inverse problems.

\begin{figure}[ht]
    \centering
    \includegraphics[width=0.9\textwidth]{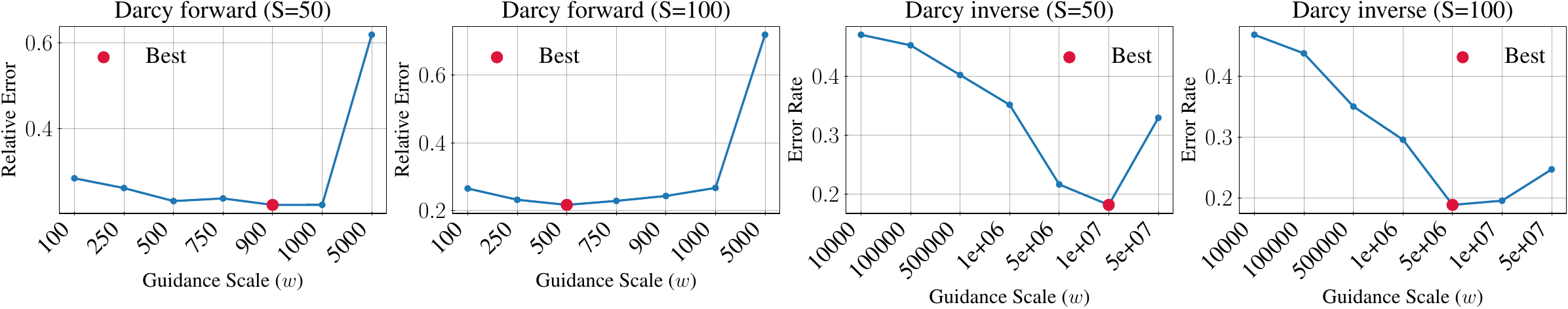}
    \caption{Linear increase, Darcy: error vs.\ guidance scale; columns: forward/inverse $\times$ \{$50$, $100$ NFEs\}; best value in red.}
    \label{fig:grid_search_dpde_linear_increase_darcy}
\end{figure}

\begin{figure}[ht]
    \centering
    \includegraphics[width=0.9\textwidth]{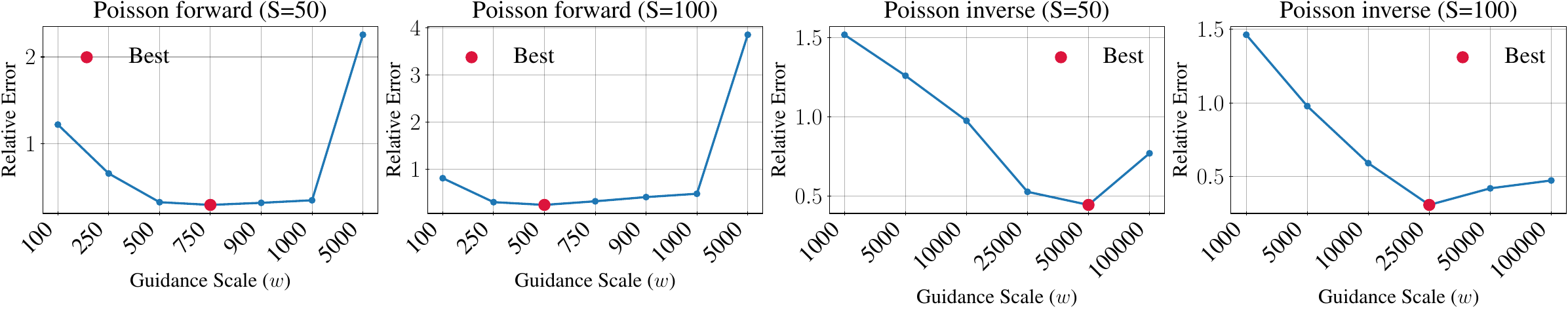}
    \caption{Linear increase, Poisson: as in Figure~\ref{fig:grid_search_dpde_linear_increase_darcy}.}
    \label{fig:grid_search_dpde_linear_increase_poisson}
\end{figure}

\begin{figure}[ht]
    \centering
    \includegraphics[width=0.9\textwidth]{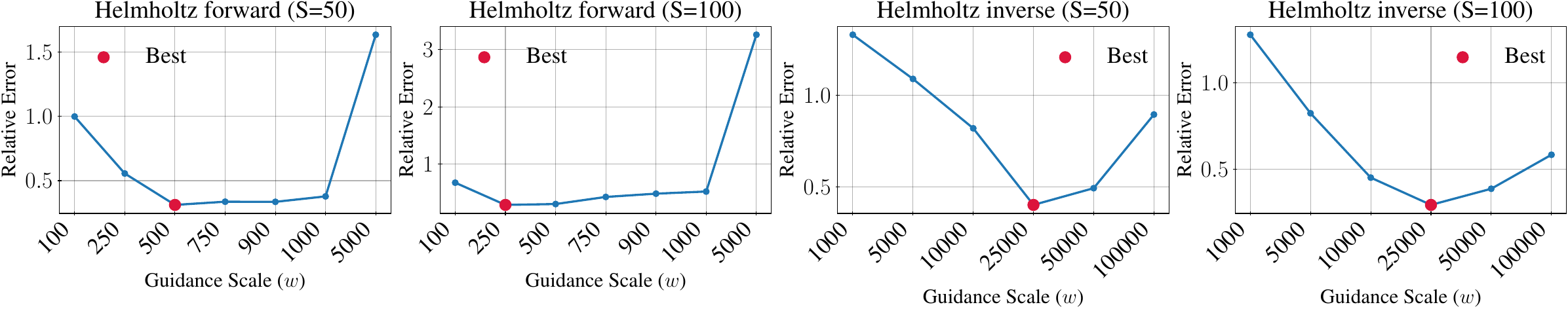}
    \caption{Linear increase, Helmholtz: as in Figure~\ref{fig:grid_search_dpde_linear_increase_darcy}.}
    \label{fig:grid_search_dpde_linear_increase_helmholtz}
\end{figure}

\begin{figure}[ht]
    \centering
    \includegraphics[width=0.9\textwidth]{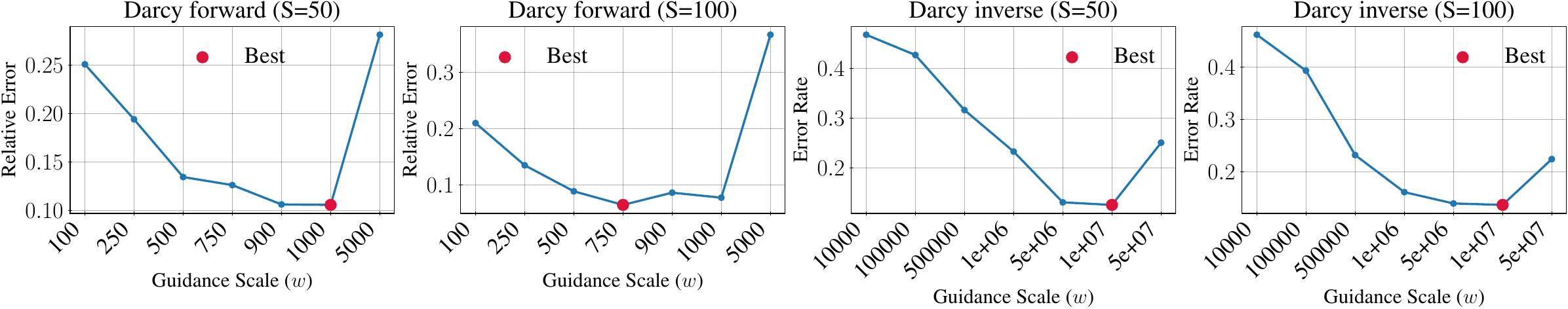}
    \caption{Linear decrease, Darcy: as in Figure~\ref{fig:grid_search_dpde_linear_increase_darcy}.}
    \label{fig:grid_search_dpde_linear_decrease_darcy}
\end{figure}

\begin{figure}[ht]
    \centering
    \includegraphics[width=0.9\textwidth]{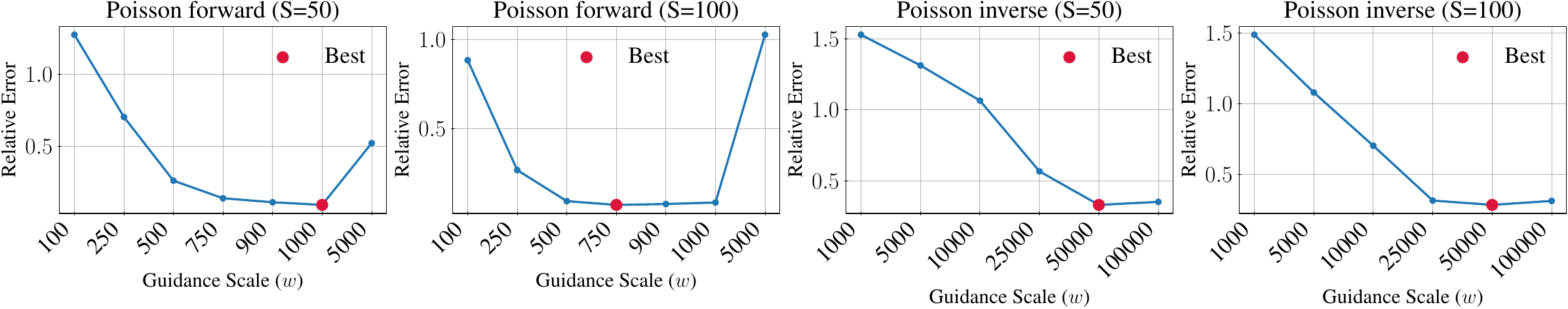}
    \caption{Linear decrease, Poisson: as in Figure~\ref{fig:grid_search_dpde_linear_increase_darcy}.}
    \label{fig:grid_search_dpde_linear_decrease_poisson}
\end{figure}

\begin{figure}[ht]
    \centering
    \includegraphics[width=0.9\textwidth]{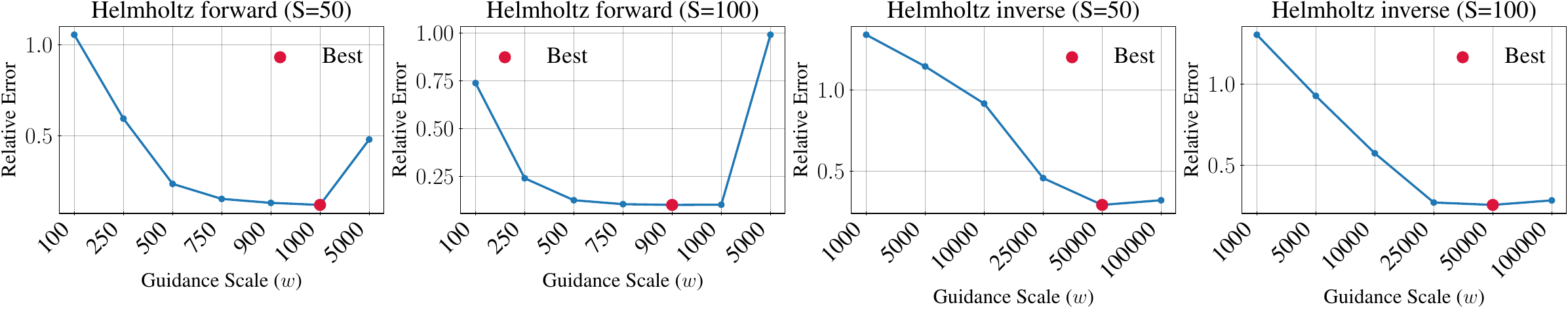}
    \caption{Linear decrease, Helmholtz: as in Figure~\ref{fig:grid_search_dpde_linear_increase_darcy}.}
    \label{fig:grid_search_dpde_linear_decrease_helmholtz}
\end{figure}

\clearpage
\section{Qualitative results}
\label{app:baseline_comparison_panels}
The qualitative results for the main experiments consist of large image panels and are provided as separate supplementary material at \url{https://drive.google.com/file/d/1XGPjo6rcNeuRd6c9JcHuaGJr3O1U0yF9/view?usp=sharing}.

\end{document}